\documentclass[11pt,a4paper,logo]{googledeepmind}

\setleftlogo[320pt]{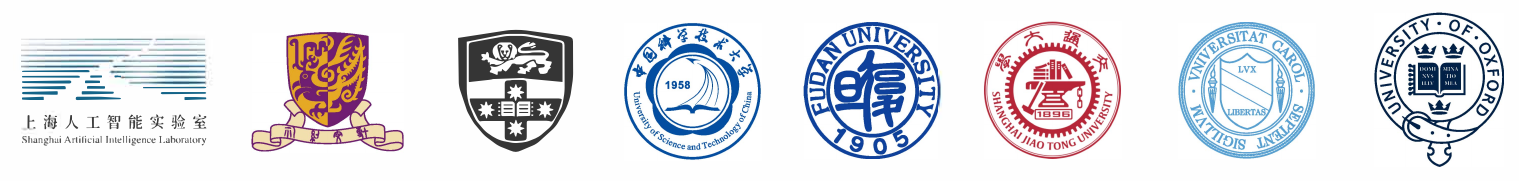}
\setrightlogo[0pt]{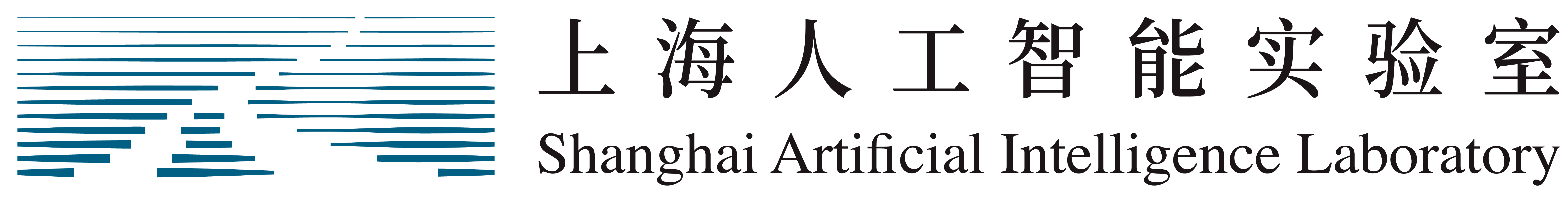}
\usepackage[T1]{fontenc}
\usepackage{pifont}
\usepackage[
    natbib=true,
    backend=biber,
    style=numeric, % Changed from 'alphabetic' to 'numeric' as per common technical report styles
    sorting=none % To keep references in order of appearance or as defined in .bib
]{biblatex}
\AtEveryBibitem{\clearfield{month}}
\AtEveryBibitem{\clearfield{day}}
\usepackage{csquotes}

\newcommand{\ProjectName}{SciReasoner}

\title{\ProjectName: Laying the Scientific Reasoning Ground Across Disciplines}

\correspondingauthor{
$\spadesuit$: Co-first Author \quad
$\diamondsuit$: Student Project Lead \quad
$\dag$: Core Contributor \quad
$\clubsuit$: Corresponding Author \\ 
Please send correspondence regarding this report to tangshixiang@pjlab.org.cn
}

\author[1,2 $\diamondsuit \spadesuit$]{Yizhou Wang}
\author[1,2 $\diamondsuit \spadesuit$]{Chen Tang}
\author[1,2 $\spadesuit$]{Han Deng}
\author[1,2 $\dag$]{Jiabei Xiao}
\author[1 $\dag$]{Jiaqi Liu}
\author[1,7 $\dag$]{Jianyu Wu}
\author[1,4 $\dag$]{Jun Yao}
\author[1,6 $\dag$]{Pengze Li}
\author[1,4 $\dag$]{Encheng Su}
\author[1,3$ \dag$]{Lintao Wang}
\author[1]{Guohang Zhuang}

\author[1,3]{Yuchen Ren}
\author[1,2]{Ben Fei}
\author[1]{Ming Hu}
\author[1]{Xin Chen}
\author[1]{Dongzhan Zhou}
\author[1]{Junjun He}
\author[2]{Xiangyu Yue}
\author[8]{Zhenfei Yin}
\author[1,2]{Jiamin Wu}
\author[1]{Qihao Zheng}
\author[1]{Yuhao Zhou}
\author[1]{Huihui Xu}
\author[1]{Chenglong Ma}
\author[1,2]{Yan Lu}
\author[1]{Wenlong Zhang}
\author[1]{Chunfeng Song}
\author[8]{Philip Torr}
\author[1,2 $\clubsuit$]{Shixiang Tang}
\author[1$\clubsuit$]{Xinzhu Ma}
\author[1,2]{Wanli Ouyang}
\author[1]{Lei Bai}

\affil[1]{Shanghai Artificial Intelligence Laboratory}
\affil[2]{The Chinese University of Hong Kong}
\affil[3]{The University of Sydney}
\affil[4]{University of Science and Technology of China}
\affil[5]{Fudan University}
\affil[6]{Shanghai Jiao Tong University}
\affil[7]{University of North Carolina at Chapel Hill}
\affil[8]{University of Oxford}

\usepackage{pdflscape}
\usepackage{textcomp}
\usepackage{rotating}

\usepackage{setspace}
\usepackage{microtype} 

\usepackage{soul} 

\usepackage{graphicx}
\usepackage{subcaption}
\usepackage{caption}

\usepackage{booktabs}

\sethlcolor{green!14}

\usepackage{array}
\usepackage{threeparttable}
\usepackage{multirow}

\usepackage{amsmath}
\usepackage{siunitx}

\usepackage{enumitem}
\usepackage{float}
\usepackage{seqsplit}
\usepackage{framed}

\usepackage{tikz}

\usepackage{ragged2e}
\usepackage[most]{tcolorbox}

\usepackage{makecell}
\usepackage{tabularray}    
\usepackage{longtable}
\usepackage{adjustbox}

\usepackage[symbol]{footmisc}
\newcolumntype{Y}{>{\RaggedRight\arraybackslash}X}
\newcommand{\cmark}{\ding{51}}
\newcommand{\xmark}{\ding{55}}

\newcommand{\Match}[1]{\textcolor{blue!70!black}{\textbf{#1}}}  
\newcommand{\PMatch}[1]{\textcolor{orange!85!black}{#1}} 
\newcommand{\BestPerf}[1]{\textbf{#1}}
\newcommand{\SecBestPerf}[1]{\underline{#1}}
\usepackage{hyperref}
\hypersetup{
    colorlinks=true,
    linkcolor=blue, % Changed for better visibility of links
    citecolor=blue,  % Changed for better visibility of citations
    filecolor=black,
    urlcolor=blue    % Changed for better visibility of URLs
}

\usepackage{tocloft}
\usepackage{etoolbox}
\usepackage{arydshln}
\makeatletter
\patchcmd{\@tocline}
    {\hfil}
    {\leaders\hbox{\hfil}\hfil}
    {}{}
\makeatother

\begin{abstract}
We present a scientific reasoning foundation model that aligns natural language with heterogeneous scientific representations. 
The model is pretrained on a 206B-token corpus spanning scientific text, pure sequences, and sequence–text pairs, then aligned via SFT on 40M instructions, annealed cold-start bootstrapping to elicit long-form chain-of-thought, and reinforcement learning with task-specific reward shaping, which instills deliberate scientific reasoning. It supports four capability families, covering up to 103 tasks across workflows: (i) faithful translation between text and scientific formats, (ii) text/knowledge extraction, (iii) property prediction, (iv) property classification, (v) unconditional and conditional sequence generation and design. Compared with specialist systems, our approach broadens instruction coverage, improves cross-domain generalization, and enhances fidelity. We detail data curation and training and show that cross-discipline learning strengthens transfer and downstream reliability. 
The model, instruct tuning datasets and the evaluation code are open-sourced at \url{https://huggingface.co/SciReason} and \url{https://github.com/open-sciencelab/SciReason}.
\end{abstract}

\begin{document}
\sloppy
\maketitle
\section*{Highlights}
\begin{itemize}

    \item \textbf{Adaptive Scientific Reasoning.} We obverse not all scientific tasks necessitate explicit reasoning—for instance, property prediction tasks often admit direct answers, whereas retrosynthesis and design tasks require elaborate reasoning. Our model can exploit the robustness of answer-based matching to generate reliable CoT supervision only where reasoning is essential, which explicitly differentiates between two task modes: “instant” tasks, where the original supervised fine-tuning (SFT) data is preserved, and “thinking” tasks, where the data is fully replaced with CoT-augmented counterparts. This principled separation enables the model to adaptively allocate reasoning capacity, producing coherent CoT rationales for complex scientific tasks while preserving efficiency and accuracy on simpler, non-reasoning tasks.
    
    \item \textbf{Task Grouped Rewarding.}
    Typical reward functions for reinforcement learning on typical LLMs lack explicit signals for encouraging and correcting scientific reasoning patterns across diverse thinking tasks, e.g., knowledge extraction, property prediction and classification, sequence generation and design. 
    To tackle these diverse scientific tasks, we propose to group various scientific rewards into three categories: distance-based reward functions for scientific predictions, matching-based reward functions for scientific retrieval and extraction, and scientific tooling-verified reward functions that are based on professional scientific tools. Such innovative grouping enables our model share the common knowledge among different tasks and simplifies the reinforcement learning framework.

    \item \textbf{Scientific Reward Softening.} Scientific tasks typically return True(1)/False(0), in which case we experimentally observe making the reinforcement learning difficult to converge.  In this paper, we develop a method to continuously scale reward signals. Specifically, we transform task-specific metrics, such as language-based scoring metrics (e.g., BERTScore) or quantitative measures (e.g., RMSE) for regression tasks, into a normalized $[0,1]$ range. This continuous reward signal effectively promotes stable training and significantly boosts the model's performance on complex scientific reasoning tasks.  

    \item \textbf{Comprehensive Scientific Capability. }
    Our model supports a broad scientific workflow, spanning five major families and encompassing up to 100 tasks: 
            \begin{itemize}
                \item \emph{Scientific Translation:}  Bidirectional conversion between human text and scientific representations for interoperability.
                \item \emph{Text \& Knowledge Extraction:} Literature-aware modeling that turns unstructured text into structured evidence for read–reason–design cycles. 
                 \item \emph{Property Prediction:} Accurate estimation of continuous scientific endpoints across chemistry, biology, and materials.
        \item \emph{Property Classification:} Robust categorization of molecular, biological, and material properties into discrete classes.
                \item \emph{Sequence Generation \& Design:} Constraint-aware, controllable generation that reuses predictive structure for efficient design loops.
            \end{itemize}
    
            We obtain \textbf{state-of-the-art performance on 54 tasks} and \textbf{rank among top-2 on 101 tasks}, demonstrating that a single backbone integrating natural language with multi‐representation scientific data can enable cross‐domain generalization beyond fixed task menus and reduce the fragmentation seen in specialist pipelines. 
            We highlight the superiority of our model in Tab.~\ref{tab:comparison_naturelm_ours_notabx_itemize_highlight} 

\begin{table}[t]
\centering
\scriptsize
\setlength{\tabcolsep}{2.5pt}
\renewcommand{\arraystretch}{1.16}
\begin{tabular}{p{2.0cm} p{3.6cm} p{4.0cm} p{3.0cm} p{3.0cm}}
\toprule
\textbf{Aspect} & \makecell[c]{\textbf{\ProjectName}} & \makecell[c]{\textbf{NatureLM}} & \makecell[c]{\textbf{Bio-T5/T5+}} & \makecell[c]{\textbf{UniGenX}} \\
\midrule
\makecell[l]{Scientific \\Reasoning} & \makebox[\linewidth][c]{\cmark} & \makebox[\linewidth][c]{\xmark} & \makebox[\linewidth][c]{\xmark} & \makebox[\linewidth][c]{\xmark} \\
\midrule
\multicolumn{5}{l}{\footnotesize\emph{Scientific Task Coverage}}\\
\cdashline{1-5}[0.6pt/2pt]
\makecell[l]{Scientific\\Translation} & \makecell[c]{10} & \makecell[c]{3} & \makecell[c]{2} & \makecell[c]{0} \\
\cdashline{1-5}[0.6pt/2pt]
\makecell[l]{Text/knowledge\\Extraction} & \makecell[c]{6} & \makecell[c]{0} & \makecell[c]{0} & \makecell[c]{0} \\
\cdashline{1-5}[0.6pt/2pt]
\makecell[l]{Property\\Prediction} & \makecell[c]{22} & \makecell[c]{0} & \makecell[c]{1} & \makecell[c]{4} \\
\cdashline{1-5}[0.6pt/2pt]
\makecell[l]{Property\\Classification} & \makecell[c]{49} & \makecell[c]{10} & \makecell[c]{13} & \makecell[c]{0} \\
\cdashline{1-5}[0.6pt/2pt]
\makecell[l]{Generation and \\ Design} & \makecell[c]{16} & \makecell[c]{17} & \makecell[c]{5} & \makecell[c]{18} \\
\cdashline{1-5}[0.6pt/2pt]
\makecell[l]{Total} & \makecell[c]{103} & \makecell[c]{30} & \makecell[c]{21} & \makecell[c]{22} \\
\midrule
\makecell[l]{Pretraining\\corpus} & \textbf{206B diverse tokens}: 

\begin{itemize}[leftmargin=*,nosep]
\item Scientific text
\item Pure scientific sequences
\item \textbf{Sequence–text interleaved pairs}
\item  \textbf{Sequence–sequence interleaved pairs}
\end{itemize}
 & 
    \textbf{140B sequence-centric tokens}: 
    \begin{itemize}[leftmargin=*,nosep]
\item \textbf{90\%} sequence-based data
\item \textbf{10\%} general text
\end{itemize}
 & 
\textbf{80B biochemical tokens}: 
     \begin{itemize}[leftmargin=*,nosep]
     \item Completely composed of proteins and small molecules data
\end{itemize}
 & 
     \textbf{84.5M structural samples}: 
    \begin{itemize}[leftmargin=*,nosep]
\item \textbf{99\%} protein-structure and materials data 
\item \textbf{1\%} small molecule data 
\end{itemize}
 \\
\cdashline{1-5}[0.6pt/2pt]
SFT corpus & 
   \makecell[c]{\textbf{40M} QA samples}
 & 
   \makecell[c]{\textbf{5M} QA samples}
 & \makecell[c]{\textbf{2.8M} QA samples} & \makecell[c]{\fontsize{7.0pt}{8.0pt}\selectfont \textbf{24M} task-related samples} \\
\cdashline{1-5}[0.6pt/2pt]
\makecell[l]{Reasoning\\corpus}  & 
    \textbf{570K} answer-aligned long-form scientific CoT data and \textbf{72K} difficulty-filtered RL data
& \makecell[c]{\textit{NA}} & \makecell[c]{\textit{NA}} & \makecell[c]{\textit{NA}} \\
\bottomrule
\end{tabular}
\caption{Comparison emphasizing \textbf{reasoning ability} and \textbf{coverage breadth}. Our model explicitly elicits scientific reasoning and supports \textbf{five} capability families with up to \textbf{103} tasks; While other methods either only emphasize cross-domain sequence-based discovery or support limited disciplines, they do not pay attention to scientific reasoning  and end-to-end workflow coverage. }
\label{tab:comparison_naturelm_ours_notabx_itemize_highlight}
\end{table} 
\end{itemize}

\clearpage

\tableofcontents
\enlargethispage{1cm} 
\thispagestyle{empty}

\section{Introduction}
Large foundation models (LFMs) have demonstrated tremendous potential in modeling natural language~\cite{wei2021finetuned,brown2020language,team2023internlm}, vision~\cite{oquab2023dinov2,shao2021intern,tang2023humanbench,ci2023unihcp,wang2025hulk}, audio~\cite{liu2024mumu,defossez2024moshi,fang2409llama,borsos2023audiolm} and multimodal~\cite{alayrac2022flamingo,wang2022ofa,cho2021unifying,hu2024ophclip} tasks. Recent works such as GPT-5~\cite{openai_introducing_gpt5}, Qwen-3~\cite{yang2025qwen3}, LLaMA-4~\cite{meta_llama4_multimodal_intelligence}, DeepSeek-v3~\cite{liu2024deepseek}, and Gemini-2.5~\cite{google_gemini25} have achieved remarkable performance across diverse domains, benefiting from advances in transformer architectures~\cite{han2022survey,dosovitskiy2020image,vaswani2017attention}, large-scale pretraining~\cite{kaplan2020scaling}, and instruction tuning~\cite{peng2023instruction,liu2023visual,dai2023instructblip}. 
While these models excel at general-purpose reasoning~\cite{wei2022chain,yao2023tree,besta2024graph}, summarization~\cite{zhang2024benchmarking,tang2023evaluating,zhang2024comprehensive}, and dialogue~\cite{hosseini2020simple,wang2023survey,gong2023multimodal}, their adaptation to scientific disciplines such as chemistry, biology, healthcare, and materials science remains an emerging frontie~\cite{hu2025survey,wei2025ai,bai2025intern,li2024gmai,ye2024gmai,hu2024omnimedvqa,ma2025meditok}. 
Scientific problems often involve diverse, domain-specific representations~\cite{Abramson2024,stokes2020deep,merchant2023scaling}, ranging from molecular representation (e.g., SMILES~\cite{weininger1988smiles}, SELFIES~\cite{krenn2020self}) and protein sequences (e.g., FASTA~\cite{pearson1988improved}) to genomic data, which differ fundamentally from natural language in structure and semantics. 
Bridging this gap requires foundation models capable of integrating symbolic, numerical, and sequence-based knowledge with natural language understanding, enabling applications such as molecular property prediction~\cite{walters2020applications,qiao2025self,li2025kolmogorov,zheng2025large}, biomolecule engineering~\cite{ulmer1983protein,kazlauskas2009finding,baek2021accurate,li2025harnessing,shaban2025foundation}, and the interpretation of complex experimental data~\cite{hadid2024geoscience,lin2024large,zhou2025scientists,he2025radarqa,wang2025omniearth}. 

Existing works on scientific large models can be generally classified into two parts: 
\begin{itemize}
    \item \textbf{Specialist Large Models. } 
    Bio-T5 families~\cite{pei2023biot5,pei2024biot5+}, ChemLLM~\cite{zhang2024chemllm}, ChemVLM~\cite{li2025chemvlm}, ChemMLLM~\cite{tan2025chemmllm}, Chem3DLLM~\cite{jiang2025chem3dllm}, ProLLama~\cite{lv2025prollama}, BioReason~\cite{fallahpour2025bioreason} adopt pretrained large language models (e.g., T5~\cite{raffel2020exploring}, LLaMA-2~\cite{touvron2023llama}, Qwen-3~\cite{yang2025qwen3}, InternLM~\cite{team2023internlm}) to solve the discipline-specific tasks~\cite{li2025ophora,chen2025slidechat,su2025gmai}. They typically collect \textit{curated, domain-specific corpora~\cite{jain2013commentary,berman2000protein,pruitt2012reference} and task-aligned instruction data~\cite{fang2023mol}, e.g., sequence–property pairs; protein/DNA/RNA sequences with structural or functional annotations; reaction and synthesis records; and multi-form molecular representations (e.g., SMILES/IUPAC/SELFIES)}.
    \item \textbf{Multi-discipline Large Models. } 
    NatureLM~\cite{xia2025naturelm} and UniGenX~\cite{zhang2025unigenx} adopt a pretrain-then-finetune scheme to learn a shared model for both life science and materials science. Galactica~\cite{taylor2022galactica} is designed to support scientific knowledge mining from massive academic literatures. 
    Biology-Instructions~\cite{he2024biology} collects a large-scale dataset to comprehend multi-omics biology. 
\end{itemize}

\begin{figure}[t]
    \centering
    \includegraphics[width=1.0\linewidth]{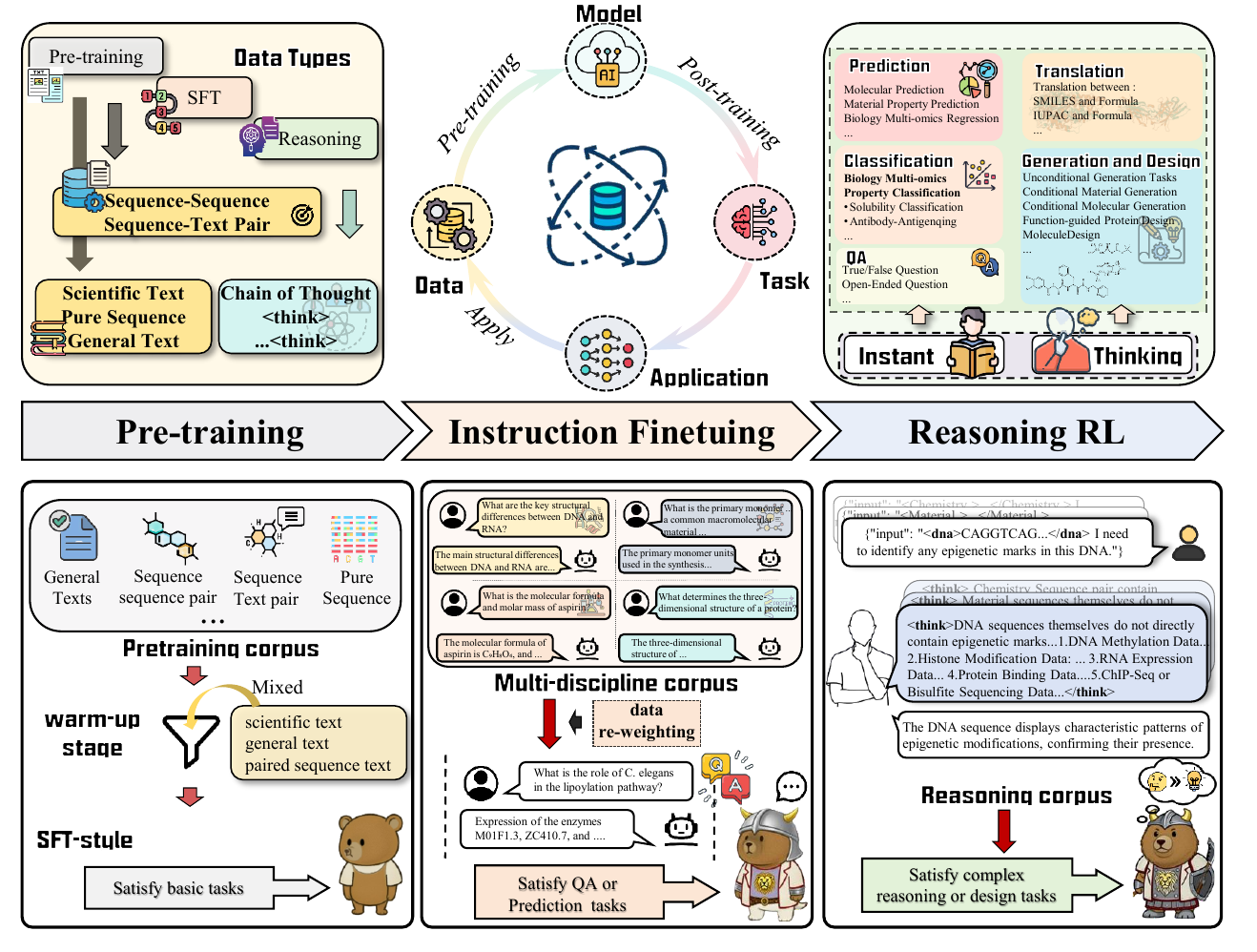}
    \caption{We present a unified scientific LLM framework organized into three stages (Pre-training, SFT, and Reasoning). In the pre-training stage, a mixed ``warm-up'' corpus aligns the base model with scientific syntax and sequences, combining scientific and general text with pure sequences and paired sequence-text / sequence-sequence data spanning DNA/RNA, proteins, small molecules, and materials. The SFT stage aggregates more than 100 curated scientific sub-tasks across these domains to adapt the model to actionable research objectives such as property prediction and sequence-text transformations. The reasoning stage introduces a rationale-augmented corpus and supports two usage modes—Instant (fast answers) and Thinking (step-by-step deliberation)—to solve complex problems. 
    % The pipeline follows the data-model-task-application lifecycle, enabling natural-language interaction and optional task-specific fine-tuning or reinforcement learning for downstream applications. 
    }
    \label{fig:framework}
\end{figure}

Although these methods and other homologous techniques~\cite{liang2023rethinking,jiang2025chem3dllm,ye2024genomics,tang2025unistd,li2025chemvlm,liang2025scwgbs,qiao2024model,dusplit} demonstrate the large models can successfully handle scientific tasks, \textit{key gaps remain in cross-domain generalization, grounding to physical and experimental constraints, and systematic coverage of heterogeneous scientific modalities}. 
For specialist models, their narrow scope and customized pipelines limit transfer outside the target subfield, fragment the model ecosystem, and, more importantly, require retraining to support unseen tasks or modalities. For multi-discipline large models~\cite{xia2025naturelm,zhang2025unigenx}, they typically focus on limited tasks (e.g., NatureLM supports only 30 tasks) and therefore fail to cover the long tail of scientific problems or generalize compositionally beyond a pre-specified task list. 
Furthermore, they still produce shallow pattern-based predictions rather than explicit multi-step reasoning traces that mirror how scientists derive conclusions from data and theory. 
We therefore aim to \textit{close these gaps by unifying multi-representation scientific data with natural language in a single backbone, scaling instruction coverage to a substantially broader and more diverse task mixture, and grounding reasoning via long-form thinking process to ensure fidelity and verifiability}.  

In this work, we propose the first scientific reasoning large language model that couples multi-representation pretraining with instruction-driven alignment and reasoning-inducing post-training. 
Our model ingests natural language alongside scientific sequences (DNA/RNA/protein), molecular strings (e.g., SMILES/IUPAC/SELFIES) and material representation through task-aware tokenization mapped into a shared backbone. 
Our training process involves pretraining, SFT-based post-training and RL-based post-training. 
Pretraining first aligns heterogeneous corpus (e.g., sequence-structure-text), while instruction tuning abstracts diverse tasks into a small set of consistent input–output schemas that encourage compositional generalization. 
Finally, a \emph{reasoning-inducing} phase equips the model with deliberate and \emph{verifiable} problem solving: we bootstrap long-form rationales via an annealed cold-start procedure, and finish with reinforcement learning using physics- and task-aware rewards (factuality, stoichiometry consistency, constraint satisfaction), yielding reliable chain-of-thought solutions. 
Our method supports a broad range of the following tasks. We organize capabilities into five families that align with the scientific workflow and enable cross-domain generalization beyond prior systems (e.g., limited task menus as in NatureLM), as follows:
\begin{itemize}
\item \textbf{Scientific Translation.} Ensures interoperability between human-readable text and machine-oriented scientific representations, enabling faithful bidirectional conversion that preserves semantics and physical meaning. Compared with earlier systems confined to a few format mappings, this category scales to many representations and thus supports data augmentation, tool chaining, and transparent human–AI collaboration. 
\item \textbf{Text and Knowledge Extraction.} 
Focuses on natural language understanding and reasoning tasks that operate directly on diverse scientific text. This evaluates the model's ability to identify scientific entities, extract structured knowledge, and answer domain-specific questions from literature. By isolating linguistic comprehension from sequence modeling, it probes whether the model can capture the semantics of scientific discourse, resolve complex entity relations, and deliver accurate answers across diverse contexts. 
\item \textbf{Scientific Property Prediction.} Provides a unified interface for estimating continuous endpoints (e.g., biochemical, materials), producing continuous signals that can be used for screening and prioritization. Unlike prior models tied to a small set of benchmarks, our formulation flexibly accommodates heterogeneous continuous targets across disciplines, enhancing transfer and compositional generalization.  
\item \textbf{Scientific Property Classification.} Handles categorical endpoints (binary, multi-class, and hierarchical) such as solubility, stability, functional annotations, or phase labels. 
It delivers high-fidelity and domain-general capabilities over categorical properties, facilitating practical applications such as candidate triage, property screening, and informed design decisions across scientific domains.
\item \textbf{Scientific Sequence Generation and Design.} Closes the loop from inference to actionable design under multi-objective and constraint-aware settings (e.g., synthesizability, stability, safety). This family goes beyond template matching by supporting controllable, cross-domain generation that reuses learned structure from prediction tasks, yielding better sample efficiency than siloed, single-discipline models. 
\end{itemize}

The overall illustration of this work is shwon in Fig.~\ref{fig:framework}. In summary, we introduce a unified scientific LLM that couples scaled multi-representation pretraining with instruction-aligned I/O schemas and a \emph{verifiable} reasoning stage. Task-aware tokenization aligns heterogeneous sequences and notations; a solve–check protocol with retrieval and domain tools, followed by reinforcement learning with physics- and task-aware rewards, enforces unit/stoichiometry and constraint consistency. The resulting single backbone supports prediction, design, translation, and literature-grounded extraction within a \emph{read–reason–design} loop, delivering strong cross-domain generalization across chemistry, proteins, genomics, and materials; the following sections detail data, training, and evaluation.

\section{Pre-Training}

\subsection{Pretraining Data Types}

To ensure downstream results on scientific sequence generation, property prediction, and sequence‐manipulation tasks, we design our pretraining strategy around four complementary types:

\begin{enumerate}
  \item \textbf{Scientific Text.} We harvest long and short prose from PubMed~\cite{pubmed}, PubChem~\cite{pubchem}, textbooks, and materials databases (e.g.,\ crystal‐structure annotations), as well as solution threads from StackExchange for a specific discipline. This corpus grounds the model's understanding of domain‐specific reasoning and terminology needed for scientific Q\&A and context‐aware property inference. 

  \item \textbf{Pure Sequence.} To capture the contextual dependencies of biomolecular and chemical sequences, we collect:
    \begin{itemize}
      \item \emph{DNA/RNA}: Serving as the fundamental blueprint of genetic information, pure nucleotide sequences allow the model to directly learn base-level patterns and long-range dependencies without being influenced by external structural or functional annotations. 
    \item \emph{Protein}: As the direct products of gene expression, amino acid sequences encode the full spectrum of structural and functional diversity in proteins, enabling the model to capture residue-level motifs and evolutionary conservation signals. 
      \item \emph{Small Molecules}: We collect Simplified Molecular Input Line Entry System (SMILES) strings from PubChem. Furthermore, 
      to ensure that the model can learn various representations of molecules beyond SMILES, we also adopt International Union of Pure and Applied Chemistry (IUPAC) and Self-Referencing Embedded Strings (SELFIES) names to achieve enhanced molecule representations. For SMILES sequences, we warp them with the \verb|<SMILES>…</SMILES>| tags. 
    \end{itemize}

  \item \textbf{Sequence–Sequence and Sequence–Text Pairs.}  Beyond standard sequence collections, we enrich the training data with cross‐discipline examples drawn from life sciences (e.g.\ protein structure), materials chemistry, and scientific literature. We hope this hybrid strategy preserves broad language fluency while instilling deep domain‐specific proficiency. To simulate the diversity of cross‐disciplinary queries, we develop: 
    \begin{itemize}
      \item \emph{Intra‐discipline pairs:} e.g.\ multi‐protein homology judgment based on sequence patterns. \textit{Such data enhances the model's ability to discern subtle evolutionary and structural relationships within a single domain, improving performance in specialized sequence comparison and classification.} 
      \item \emph{Cross‐discipline pairs:} e.g.\ prediction of small‐molecule binding to a protein segment. \textit{By bridging cross-disciplinary representations, these pairs could improve the capacity to integrate different knowledge for tasks such as drug discovery and biomolecular engineering.}
      \item \emph{Multi‐representation of the same entity:} concurrent IUPAC, SELFIES, and SMILES representations of a molecule. \textit{This could promote translation and reasoning between complementary notations, enabling robust understanding and cross‐format interoperability.} 
      \item \emph{Text–data pairs:} combined prompts and labels for molecular, nucleic, or protein properties. \textit{Pairing natural language descriptions with scientific sequences enables rich bidirectional interaction, empowering downstream natural language understanding (e.g., protein or molecule interpretation) and language‐driven generation of scientific sequences. }
    \end{itemize}

    \item \textbf{General Text.} In order to maintain the language capabilities of the model itself, we adopt C4 dataset as our general text corpus. Compared with the existing methods~\cite{xia2025naturelm}, the pre-training data they use contains only about 10\% of natural text, and almost entirely relies on pure sequence data; therefore, they lack coherent conversation capabilities and cross-domain reasoning abilities.
\end{enumerate}

\begin{figure}
    \centering
    \includegraphics[width=1.0\linewidth]{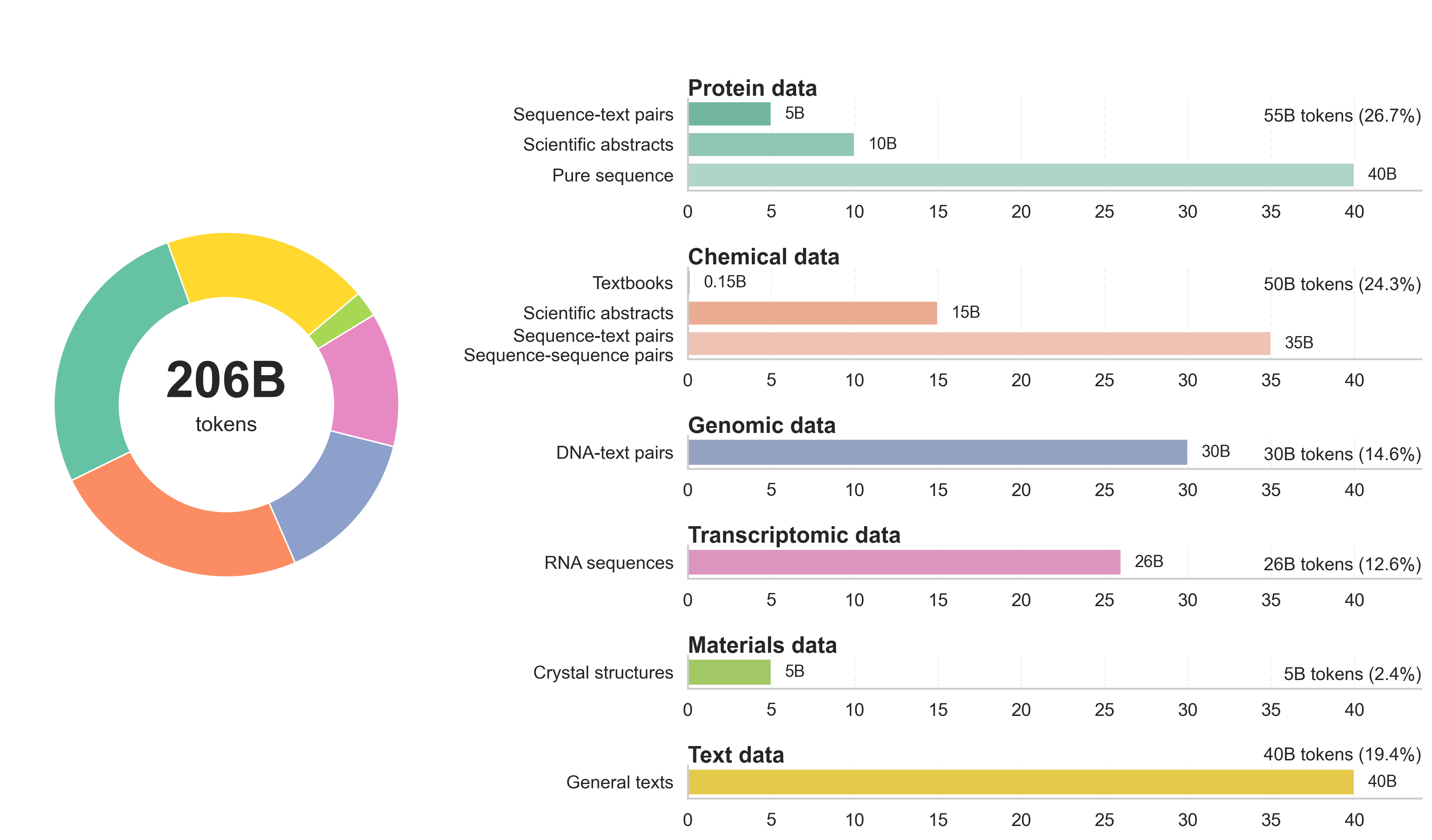}
    \caption{Composition of the 206B-token scientific pretraining corpus across five scientific domains: protein data, chemical data, genomic data, transcriptomic data  and materials data.}
    \label{fig:pt_data_distribution}
\end{figure}

\subsection*{Data Preparation Details}

\paragraph{DNA.} We retrieve about 200 billion bases from NCBI~\cite{ncbi_homepage}, sampling at an average of 1,000 bp fragments per organism to produce 30 billion \verb|<dna></dna>|-tagged tokens and is represented by the FASTA format. Essential metadata (e.g., organism name and genomic coordinates) is included to anchor the model's interpretation of each sequence's context. 

\paragraph{RNA.} We collect about 20 billion RNA tokens from RNAcentral~\cite{rnacentral_homepage}, reflecting diverse transcriptomic contexts. Then, we warp the collected RNA sequence (FASTA format) with the \verb|<rna></rna>| tag. The basic information of the RNA sequence (e.g., Family) is also retrieved and is used to guide the model’s understanding of sequence context. 

\paragraph{Protein.} We collect protein sequences from UniRef50 and UniRef90 (mammalia) and warp the raw sequence with \verb|<protein></protein>| tag, this yields about 40 billion protein tokens. 
We also pair sequence fragments with concise natural‐language descriptions, and sample pure text abstracts from PubMed for additional context. 

\paragraph{Small Molecules.} 
To ensure reliable representation and alignment of small molecules, we curate diverse textual resources, construct multiple molecular encodings, and implement rigorous token validation. 
\begin{itemize}
  \item \emph{Text:} abstracts from PubChem and ``long‐form'' chemistry Q\&A from StackExchange. 
  \item \emph{Multi-Representations:} To ensure comprehensive molecular characterization, we utilize multiple chemical representations. These are obtained either directly from established databases (e.g., SMILES-to-IUPAC from PubChem) or generated via specialized toolkits (e.g., SMILES-to-SELFIES using the SELFIES library~\cite{krenn2020self}). 
  \item \emph{Tagging:} We apply a multistage filter: regex based on RDKit~\cite{rdkit} / SELFIES~\cite{krenn2020self} toolkits, small LLM validation and manual review to eliminate ambiguous tokens (e.g., ``In'', ``Is'') and ensure high‐fidelity tagging. 
\end{itemize}

\paragraph{Materials.} 
We collect raw material data from publicly available material databases, including Material Projects~\cite{materials_project}, GNoME~\cite{merchant2023scaling}, SNUMAT~\cite{snumat}, etc. 
The fields of raw data include the crystal structure files (Crystallographic Information File, CIFs), chemical composition, space-group and crystal-system information, material type (e.g., bulk or slab), thermodynamic metrics (formation energy, energy above hull), electronic and mechanical properties (band gap, density, elastic moduli), magnetic properties, per-site coordination details, bond-length descriptions, etc. 

To convert the raw data into the pretraining corpus, we generate task-specific templates by prompting an LLM to convert structural annotations into natural‐language descriptions, then score and select the best templates for large‐scale synthetic data generation. 

Fig.~\ref{fig:pt_data_distribution} summarizes the composition of our training corpus across five scientific domains and their internal subdivisions. Specifically, protein sequences account for the largest share (55 billion tokens, 33.1\%), followed by chemistry data (50 billion tokens, 30.1\%), DNA genome fragments (30 billion tokens, 18.1\%), RNA transcripts (26 billion tokens, 15.7\%), and materials‐science texts (5 billion tokens, 3.0\%). 

\subsection{Training Protocol}

\begin{figure}
    \centering
    \includegraphics[width=0.6\linewidth]{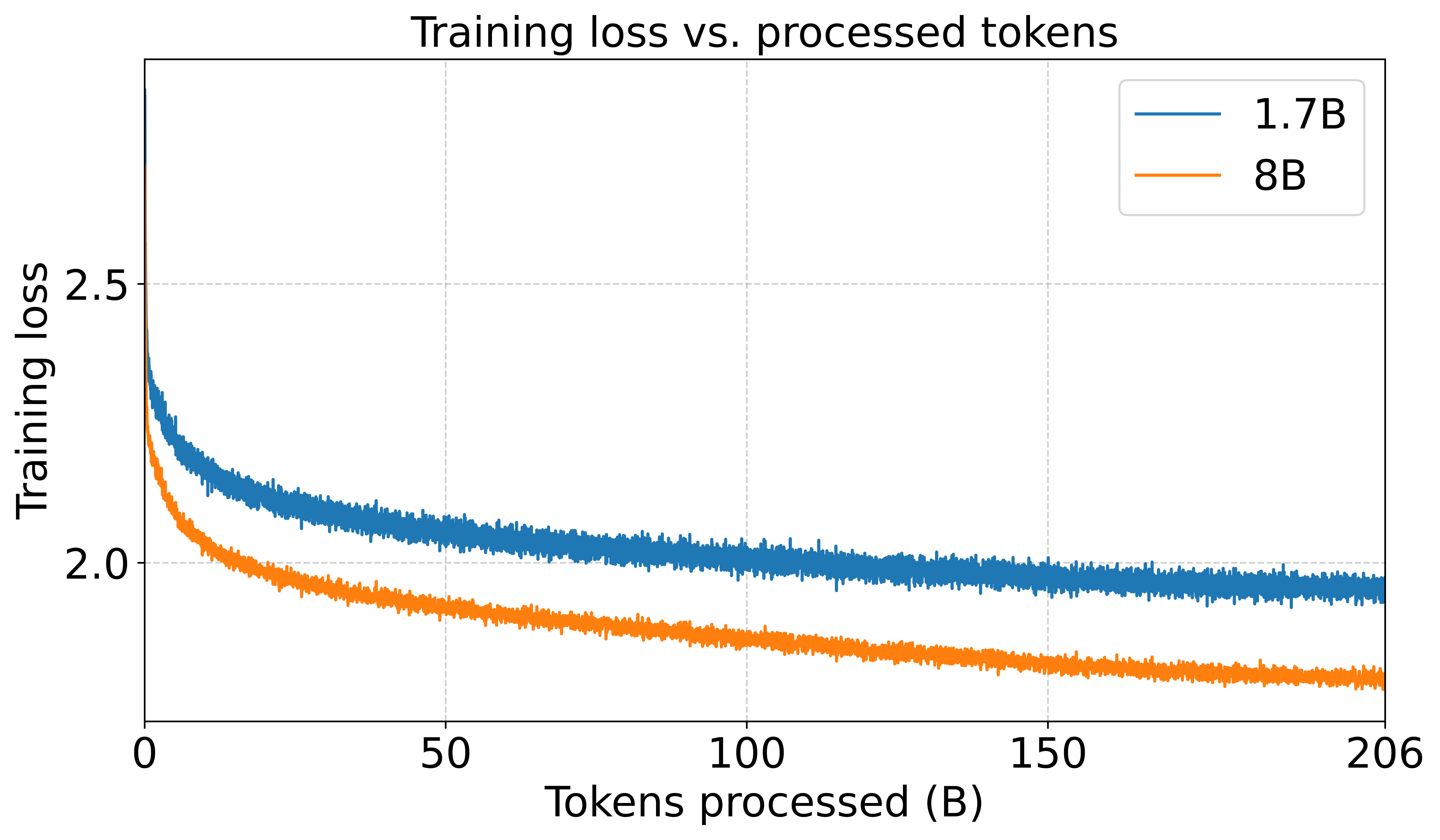}
    \caption{Training loss over train tokens for the 1.7B and 8B models. The model does not exhibit any saturation phenomenon during the training process. }
    \label{fig:pt_8b_loss}
\end{figure}

We train the model for one epoch using bfloat16 mixed-precision. The base models are Qwen3~\cite{yang2025qwen3} 1.7B and 8B, respectively. 
Training batches consisted of 16 and 32 sequences per device for 1.7B model and 8B model, respectively. 
We set the initial learning rate to $2\times10{-4}$ and accumulated gradients over two steps, applying a linear warm-up over the first 1\% of total steps. We especially arrange the data for the following training stages: 

\begin{itemize}
    \item At the warm-up stage, we adopt a mixed form of scientific text, general text and paired sequence text data. This could help the model rapidly adapt to both domain-specific and general linguistic patterns before the main training phase. 
    \item At the end of the training, we include SFT-style training data to refine the model's capability to follow complex, domain-specific instructions and produce task-aligned outputs. This stage serves to align the model’s generative behavior with SFT-like reasoning and formatting standards, thereby enhancing its utility in real-world scientific applications. 
\end{itemize}

For the training system, we use 128 A800 GPUs for 1.7B model and 256 A800 GPUs for 8B model. The training loss curves are shown in Fig.~\ref{fig:pt_8b_loss}.

\section{Post-Training}
\subsection{Supervised Fine-tuning}
\subsubsection{Data}

In the supervised fine-tuning (SFT) stage, we integrated about 100 scientific sub-tasks spanning multiple disciplines, including chemistry (Mol-Instructions~\cite{fang2023mol}, ChemData~\cite{zhang2024chemllm}, USPTO50k~\cite{schneider2016s}, SMolInstruct~\cite{yu2024llasmol}, etc.), DNA, RNA, proteins (Biology-Instructions~\cite{he2024biology}, PEER~\cite{xu2022peer}, OPI~\cite{xiao2024opi}, GUE~\cite{zhou2023dnabert}, BEACON~\cite{ren2024beacon}, etc.), materials (LLM4Mat~\cite{rubungo2025llm4mat}, Material Projects~\cite{materials_project}) and general instruction texts from OIG~\cite{oig_laion}. 
Moreover, cross-sequence interaction tasks and natural language–based scientific question answering are categorized within the cross-domain corpus. 
To warp the scientific sequences with discipline-related tags (i.e., \verb|<SMILES>…</SMILES>|, \verb|<dna>…</dna>|, \verb|<rna>…</rna>|, \verb|<protein>…</protein>|, etc.), the data preparation pipeline involved: 
\begin{itemize}
    \item Extracting scientific sequences from chemistry, DNA, RNA, protein, and macromolecule datasets via rule-based scripts tailored to each dataset format.
    \item Conducting secondary annotation using domain-specific checkpoints from pre-trained single-discipline models.
    \item Filtering and adding scientific tags with the assistance of small LLMs.
    \item Converting all data into a unified dialogue format for SFT training. 
\end{itemize}

We show the data distribution of supervised finetuning stage in Tab.~\ref{tab:sft_data_distribution}.

\begin{table}[t]
\centering
\begin{tabular}{l r r}
\toprule
Category & Training samples (M) & Proportion \\
\midrule
Materials     & 8.278 & 20.79\% \\
General Text & 10.873 & 27.30\% \\
DNA\&RNA      & 5.476 & 13.75\% \\
Protein       &  2.473 &  6.21\% \\
Molecule      & 12.556 & 31.53\% \\
Cross Domain      & 0.168 & 0.42\% \\
\midrule
\textbf{Total} & \textbf{39.824} & \textbf{100.00\%} \\
\bottomrule
\end{tabular}
% \caption{Distribution of SFT training samples across domains (millions and proportions); In total of 63.35 million samples.}
\caption{Distribution of supervised fine-tuning (SFT) training samples across five major scientific domains and additional general text data, reported in millions and proportions. Each domain undergoes tailored preprocessing, domain-specific annotation, and tagging (e.g., \texttt{<SMILES>...</SMILES>}, \texttt{<dna>...</dna>}, \texttt{<protein>...</protein>}) before being converted into a unified dialogue format for SFT. The corpus comprises 63.35 million samples.}

\label{tab:sft_data_distribution}
\end{table}

\subsubsection{Training Protocol}
We train the model for five epochs using 16-bit bfloat16 precision. A per-device batch size of 32 is used, with gradient accumulation over two steps to achieve a larger effective batch size. The initial learning rate is set to $5\times10^{-5}$, and we apply a linear warm-up over the first 5\% of total steps. 
Inputs are tokenized lazily with right-side truncation to a maximum sequence length of 8192 tokens, and the Liger-kernel~\cite{hsu2024liger} is enabled for efficiency. The training process is accelerated using DeepSpeed ZeRO Stage 2~\cite{rasley2020deepspeed} and FlashAttention implementation~\cite{dao2023flashattention}. 

Based on the SFT dataset, we train four single-discipline models in DNA/RNA, protein, materials science, and chemical molecules. However, during multi-disciplinary joint training, we observe substantial performance degradation in certain sub-tasks due to data imbalance after dataset merging. To address this, we adopt two strategies: (1) we conduct large-scale pre-training followed by augmentation with additional SFT data—with materials science receiving further data expansion—and (2) data re-weighting, which adjusts task-level proportions to mitigate the impact of low-quality instructions or underrepresented tasks. This re-weighting approach effectively alleviates performance decline in mixed-dataset training.

\subsection{Reasoning-based Reinforcement Learning}

\subsubsection{Data}
\paragraph{Cold Start}
\noindent Before reinforcement learning (RL), we construct a correct-only chain-of-thought (CoT) corpus as a cold-start prior. For each sub-task, we use \emph{DeepSeek-R1-Distill-Qwen-32B} to generate \texttt{CoT, answer} pairs and target \(T{=}500\) correctly answered items under a hard evaluation budget of \(C{=}50{,}000\) prompts. We first draw \(6{,}000\) random prompts, evaluate correctness, and retain the CoT attached to correct answers. If after \(N\) evaluated items the tally is \(S<T\), we estimate the additional sample size via the empirical accuracy \(\hat p{=}S/N\) as \(n_{\text{add}}=\left\lceil (T-S)/\max(\hat p,\varepsilon)\right\rceil\) (with small \(\varepsilon{>}0\) for numerical stability), then repeat sampling–generation–adjudication until either \(S\ge T\) or \(N\ge C\); if the cap is reached, we keep all accumulated correct examples. Correctness is adjudicated by direct comparison of the generator’s final answer to SFT ground truth for auto-gradable tasks (exact match or task-specific tolerance), and by an external \emph{GPT-3.5-turbo} judge for semantic equivalence on free-form tasks (e.g., protein descriptions). This adaptive, budgeted, correct-only harvesting mitigates rationale noise and yields a reliable initialization for subsequent RL.

\paragraph{Reinforcement Learning} 
We curate reinforcement learning (RL) data at the level of each sub-task using an empirical solve-rate filter that targets medium-difficulty instances. 

Let \(D^{\text{train}}_{k}\) denote the k-th sub-task's training set. For every example \(x \in D^{\text{train}}_{k}\):
\begin{enumerate}
    \item \textbf{Stochastic rollouts.} We generate \(N=8\) independent rollouts from \(M_0\) on \(x\) using stochastic decoding with \texttt{do\_sample=True}. To balance output diversity against format fidelity, we adopt an \emph{adaptive temperature} \(T\) tuned once per sub-task prior to collection: we initialize at \(T=0.9\) (the DAPO default setting) and \emph{increase} \(T\) if the eight-way samples show insufficient diversity (e.g., low uniqueness of normalized responses or highly repetitive content); conversely, if outputs become overly diffuse and induce frequent format non-compliance or invalid parses, we \emph{decrease} \(T\). During the final eight rollouts for each \(x\), we hold \(T\) fixed and vary only the random seed, keeping all other decoding parameters constant. 
    \item \textbf{Empirical solve rate.} For rollout \(i\), compute a correctness indicator \(f_i(x)\in\{0,1\}\) using the sub-task’s canonical metric (e.g., exact match or a thresholded score), and aggregate
    \begin{equation}
        \hat p(x) \;=\; \frac{1}{8}\sum_{i=1}^{8} f_i(x).
    \end{equation}
    \item \textbf{Mid-difficulty filter.} Retain examples with partial success:
    \begin{equation}
        S \;=\; \left\{\,x \in D^{\text{train}}_{k} \;\middle|\; 0.125 < \hat p(x) < 0.875 \,\right\}.
    \end{equation}
    This removes items that are nearly always solved (\(\hat p \approx 1\)) or nearly always failed (\(\hat p \approx 0\)), which typically yield weaker gradients for policy improvement.
    \item \textbf{Sampling.} Uniformly sample without replacement \(K=1000\) examples from \(S\). If \(|S|<1000\), include all of \(S\) and \emph{top up} to \(K=1000\) by uniformly sampling from \(D_{\text{train}}\setminus S\).
\end{enumerate}

The resulting 1,000-example set serves as the RL training pool for that sub-task.

Focusing RL on neither-trivial-nor-impossible instances yields stronger and more stable improvement: items with \(\hat p \in (0.125,0.875)\) (i) expose the policy to cases where exploration can flip outcomes, (ii) avoid vanishing advantages on saturated positives, and (iii) reduce uninformative gradients from uniformly failed cases. 

\subsubsection{Training Protocol}
\paragraph{Cold Start}
Before reinforcement learning, we apply an \emph{annealed cold-start} (ACS) adaptation to a the model after SFT stage to (i) endow it with the ability to produce long, coherent chains of thought (CoT) on tasks that genuinely require stepwise reasoning, while (ii) preserving concise, direct-answer behavior on tasks that do not. 
We therefore partition the SFT task suite into \emph{thinking} and \emph{instant} categories. For thinking tasks, we \textbf{replace} the original SFT samples with their CoT-augmented counterparts (teacher-forced rationales followed by the final answer). For instant tasks, we retain the original direct-answer SFT data unchanged. Moreover, we have observed that \textbf{task-level replacement} outperforms within-task \emph{mixing} of CoT and non-CoT targets. Mixing presents inconsistent sequence lengths, styles, and termination signals for near-identical instructions, weakening the model’s calibration about when to elaborate. Replacement yields a cleaner conditional distribution, i.e., inputs from thinking tasks consistently map to CoT-style trajectories, thereby improving both the incidence and quality of generated rationales without confusing non-reasoning behavior.

To minimize behavioral drift from SFT while shaping rationale production where appropriate, the model is trained with a \emph{reduced learning rate} and a short schedule relative to SFT (i.e., small-step updates). This ``low-temperature'' adjustment biases the model toward emitting structured reasoning \emph{only} when the input distribution indicates a reasoning task.

\paragraph{Reinforcement Learning.}

\begin{figure}[t]
\centering
\includegraphics[width=0.8\linewidth]{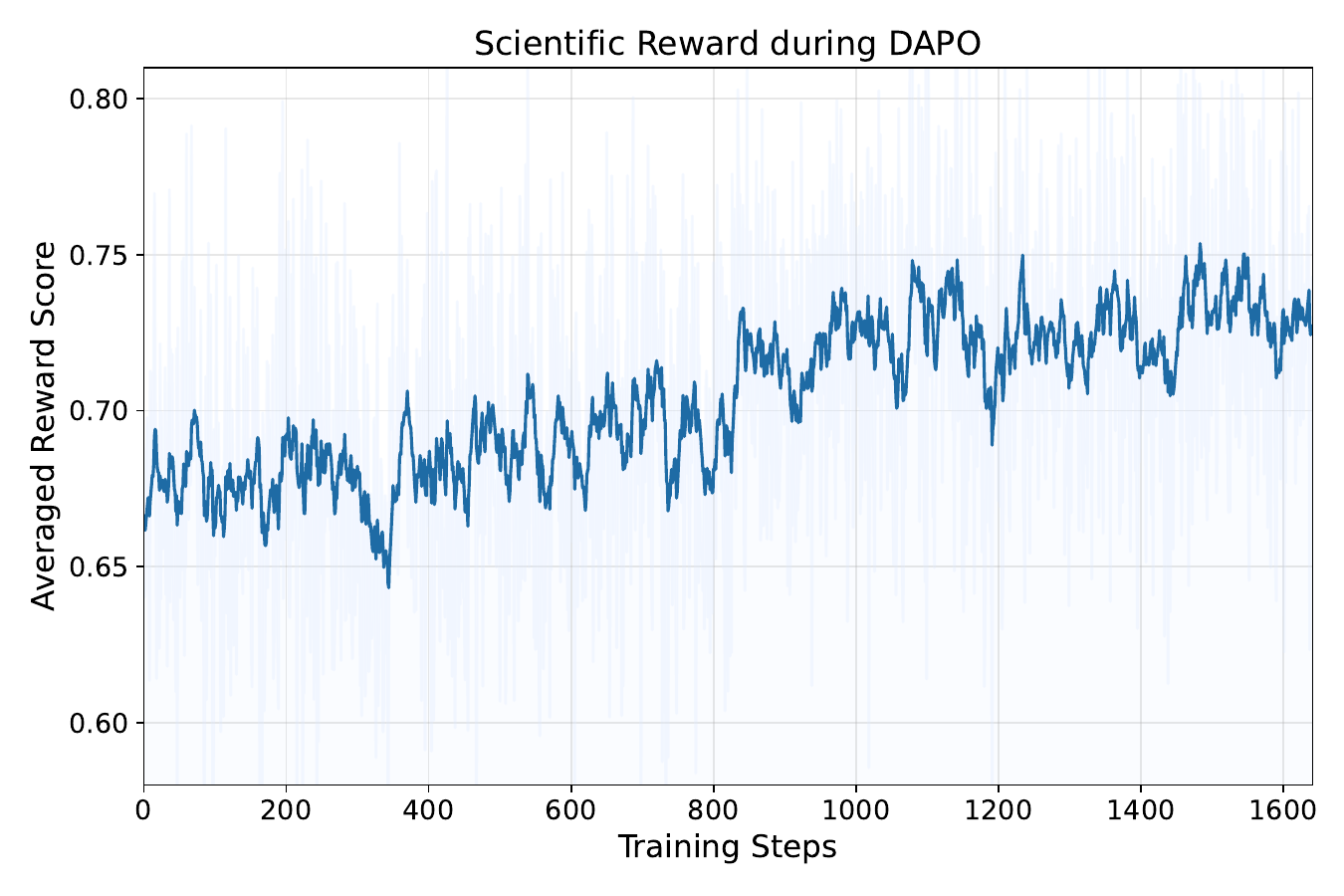}
\caption{\small Scientific reward during DAPO training progress. }
\label{fig:rl_reward_figs}
\end{figure}

We adopt DAPO~\cite{yu2025dapo} as the RL algorithm to optimize the policy on heterogeneous scientific tasks. For each prompt we draw \(G\) candidates from the old policy \(\pi_{\theta_{\text{old}}}\), compute group-standardized advantages:
\begin{equation}
    r_{i,t}(\theta)=\frac{\pi_{\theta}(o_{i,t} \mid q, o_{i,<t})}{\pi_{\theta_{\text{old}}}(o_{i,t} \mid q,o_{i,<t})},\quad
    \hat{A}_{i,t} = \frac{R_i - \text{mean}(\{R_i\}_{i=1}^G)}{\text{std}(\{R_i\}_{i=1}^G)}.
\label{eq:advantage_calculation}
\end{equation}

and maximize the PPO-style clipped objective:
\begin{equation}
\begin{aligned}
\mathcal{J}_{\text{DAPO}}(\theta) =\quad& \mathbb{E}_{(q,a)\sim \mathcal{D}, \{o_i\}_{i=1}^G\sim \pi_{\theta_\text{old}}(\cdot\mid q)}\\&
\Bigg[\frac{1}{\sum_{i=1}^{G}|o_i|}\sum_{i=1}^{G}\sum_{t=1}^{|o_i|} 
\min \Big( r_{i,t}(\theta) \hat{A}_{i,t},  
\ \text{clip} \Big( r_{i,t}(\theta), 1 - {\varepsilon_{\text{low}}}, 1 + {\varepsilon_{\text{high}}} \Big) \hat{A}_{i,t} \Big) \Bigg]
\\
\text{s.t.}\quad& 0< \Big|\{o_i\mid\texttt{is\_equivalent}(a,o_i)\}\Big|< G,
\end{aligned}
\label{eq:dapoloss}
\end{equation}
with the asymmetric \emph{Clip-Higher} schedule to preserve exploration, together with \emph{Dynamic Sampling} that filters degenerate groups with identical rewards and a token-level policy-gradient reduction. 
Unlike the original rule-based binary reward, many of our tasks (free-form semantic judgment, information extraction, and numeric regression) do not admit a single exact-match criterion. 
We therefore replace the binary signal with a \emph{reward softening} scheme that maps the grouped metrics to a common \([0,1]\) scale: 
\begin{equation}
\label{eq:task_aware_soft_reward}
R_i^{\text{soft}}
\;=\;
g\!\big(m_i(q,o_i,\mathcal{A}(q))\big)
\end{equation}
where \(g(\cdot)\) is a monotone calibration ensuring comparable reward magnitudes across tasks and \(m_i\) is a \emph{group-specific} quality metric. 
For grouping scheme, we employ distance-based rewards for scientific prediction tasks, matching-based rewards for retrieval and extraction tasks, and tool-verified rewards leveraging professional scientific software. 

We show the training dynamics of scientific reward in Fig~\ref{fig:rl_reward_figs}. 

\section{Evaluation and Application}

\subsection{Task Overview}

\paragraph{Translation.} 
The tasks in this group involve converting information between different scientific formats or representations, for example translating between IUPAC names and SMILES, generating descriptive text from molecular structures, or performing the inverse transformation. These capabilities are essential to ensure interoperability within the scientific data ecosystem and to enable effective human-machine communication.

\paragraph{Text and Knowledge Extraction/Question Answering.} 
Applying natural language processing to scientific literature, these tasks include chemical entity recognition, interaction extraction, and various formats of question answering (multiple-choice, open-ended, true/false). They integrate AI methods with literature analysis to facilitate rapid extraction of knowledge from scientific texts. 

\paragraph{Property Prediction and Classification.} 
These tasks span a wide range of predictive challenges, from molecular properties (e.g., solubility, toxicity) to materials characteristics (e.g., MP classification, properties derived from JARVIS-DFT), DNA/RNA predictions (e.g., epigenetic markers, transcription factor binding in human), and protein functions (e.g., catalytic activity, stability). Their scope—encompassing chemistry, materials, nucleic acids, and proteins—highlights the universal demand for accurate predictive models in scientific discovery. 

\paragraph{Generation and Design.} 
This category focuses on the unconditional generation tasks and the conditional generation tasks. It includes unconditional generation of molecules, proteins, RNA, and materials, as well as targeted design challenges such as rational antibody design (RAbD) and programmable RNA switches. Progress in this domain is critical for accelerating the discovery of new compounds and biomolecules. 

\subsection{Baseline}
For general-purpose closed-source models, we compare against \textbf{Gemini-2.5-pro}\footnote{Gemini-2.5-pro version: 2025-06-17} and \textbf{GPT-o3}\footnote{GPT-o3 version: 2025-04-16. Results are provided in Appendix Table.~\ref{tab:appendix_scientific_translation} to Table~\ref{tab:appendix_generation_design}.}\footnote{Since some test tasks involve extremely large-scale test sets, to avoid substantial testing costs, for closed-source models, we randomly sampled 1,000 samples from tasks with a test set size exceeding 1,000 for testing. Meanwhile, we also present the performance of our models on these 1,000 samples for comparison.}; for general-purpose open-source models, we use \textbf{GPT-oss-120B} and \textbf{DeepSeek-r1}\footnote{DeepSeek-r1 version: 2025-05-28. }. As specialist baselines, we consider two categories: (i) non-language models, represented by Transformer encoder-based architectures (denoted as \emph{non-LLM Experts}); and (ii) language models, represented by the best-performing specialist large language model on this task (denoted as \emph{Specialist LLMs}). 
Please refer to the Appendix for details. 

\subsection{Scientific Translation Tasks} 

\setlength{\extrarowheight}{1pt}
\begin{table}[t]
\centering
\scriptsize
\resizebox{\linewidth}{!}{

\begin{tabular}{l c c c c c c c} 
\toprule
Task & Metric & \makecell[c]{Gemini\\2.5-pro} & \makecell[c]{GPT\\oss} & \makecell[c]{Non-LLM \\ Experts} & \makecell[c]{Specialist\\LLMs} & \ProjectName-1.7B & \ProjectName-8B \\
\midrule
\addlinespace[0.8em]
\multicolumn{8}{l}{\raisebox{1.2ex}[0pt][0pt]{\bfseries Molecular Representation Translation}}\\[-0.8ex]
\quad SMILES to IUPAC  & \makecell[c]{Top1 \\Split Match} {\normalsize $\uparrow$} & 4.70\%\dag & 0.17\% & \textemdash{} & 29.00\% & \SecBestPerf{49.52\%}|50.20\%\dag & \BestPerf{56.63\%}|57.10\%\dag\\
\quad SMILES to Formula & \makecell[c]{Top1 \\Element Match} {\normalsize $\uparrow$} & 55.20\%\dag & 19.88\% & \textemdash{} & 93.20\% & \SecBestPerf{95.92\%}|87.1\%\dag & \BestPerf{96.39\%}|96.00\%\dag \\
\quad IUPAC to SMILES & \makecell[c]{Top1 \\Split Match} {\normalsize $\uparrow$} & 38.20\%\dag & 7.58\% & \textemdash{} & 70.10\% & \SecBestPerf{78.85\%}|86.50\%\dag & \BestPerf{84.40\%}|90.70\%\dag \\
\quad IUPAC to Formula & \makecell[c]{Top1 \\Element Match} {\normalsize $\uparrow$} & 88.50\%\dag & 33.24\% & \textemdash{} & 87.90\% & \SecBestPerf{91.81\%}|88.10\%\dag & \BestPerf{92.65\%}|93.00\%\dag \\
\midrule
\addlinespace[0.8em]
\multicolumn{8}{l}{\raisebox{1.2ex}[0pt][0pt]{\bfseries Natural Language Translation of Molecule}}\\[-0.8ex]
\quad Molecular Description            & ROUGE-L $\uparrow$ & 0.23 & 0.05 & \SecBestPerf{0.75} & 0.29 & \SecBestPerf{0.75} & \BestPerf{0.78} \\
\quad Molecular Captioning             & MENTOR $\uparrow$  & 0.42\dag & 0.15 & \textemdash{} & 0.45 & \SecBestPerf{0.59}|0.59\dag & \BestPerf{0.60}|0.58\dag \\
\midrule
\addlinespace[0.8em]
\multicolumn{8}{l}{\raisebox{1.2ex}[0pt][0pt]{\bfseries Protein Functional Translation}}\\[-0.8ex]
\quad CASPSimilarSeq                   & ROUGE-L$\uparrow$  & 0.01 & 0.02 & \textemdash{} & 0.74 & \SecBestPerf{0.83} & \BestPerf{0.85}  \\
\quad IDFilterSeq                      & ROUGE-L$\uparrow$  & 0.01 & 0.03 & \textemdash{} & 0.70 & \SecBestPerf{0.77} & \BestPerf{0.82} \\
\quad UniProtSeq                       & ROUGE-L$\uparrow$  & 0.01 & 0.03 & \textemdash{} & 0.71 & \SecBestPerf{0.84} & \BestPerf{0.88}  \\
\quad Mol-Instructions                 & ROUGE-L$\uparrow$  & 0.21\dag & 0.07 & 0.68 & 0.44 & \SecBestPerf{0.83}|0.82\dag & \BestPerf{0.99}|0.98\dag  \\
\bottomrule
\end{tabular}
}
\caption{Evaluation results of scientific translation tasks. 
\BestPerf{Bold} indicates the best performance, and \SecBestPerf{underline} indicates the second best. The names of expert models and specialist LLMs for each task are provided in Appendix Table.~\ref{tab:appendix_scientific_translation}.
\dag~denotes that only randomly sampled 1000 test samples were utilized for evaluation in the corresponding task. This measure is adopted because extremely large-scale test sets exist for certain tasks, which would lead to substantial testing costs. 
}
\label{tab:scientific_translation}
\end{table}

Scientific translation evaluates whether a model preserves scientific semantics while mapping across modalities—symbolic encodings and formulae, natural-language text, and biological sequences. The goal is not to invent new facts but to produce an equivalent rendering that remains invertible and respects domain constraints (e.g., stoichiometry, valence, sequence–function coherence). This suite therefore probes compositional reasoning, ontology grounding, and robustness to notation/paraphrase as a modality-agnostic measure of semantic fidelity. The results are shown in Tab.~\ref{tab:scientific_translation}.

\paragraph{Molecular Representation Translation. }
Following SMolInstruct~\cite{yu2024llasmol}, we evaluate the model's ability to convert faithfully among common chemical representations, i.e., SMILES strings, IUPAC names, and empirical formulae. The tasks cover SMILES$\leftrightarrow$\,IUPAC, SMILES$\leftrightarrow$\,Formula, and IUPAC$\leftrightarrow$\,Formula, with performance judged by exact-match accuracy and structure-level equivalence after canonicalization. 

\begin{itemize}
    \item \textbf{SMILES to IUPAC Translation. } This task takes a molecule's SMILES string as input and converts it into its corresponding IUPAC name. This is a fundamental challenge in cheminformatics, as it requires the model to interpret a simplified linear notation of a chemical structure and convert it into the standardized naming convention used by chemists.
    \item \textbf{SMILES to Formula Translation. } This task takes a molecule's SMILES string as input and converts it into its corresponding molecular formula. The model receives a SMILES string as input and generates the complete molecular formula, which specifies the number of atoms for each element in the molecule. This task is crucial for computational chemistry, as it bridges the gap between a structural representation and the molecule's elemental composition.
    \item \textbf{IUPAC to SMILES Translation. } This task converts a chemical name into its corresponding SMILES string. The model takes a molecule's IUPAC name as input and generates the SMILES string, which is a concise text-based representation of the molecule's chemical structure. This task is fundamental for many cheminformatics applications, as it allows for the conversion of human-readable names into a format that computers can process.
    \item \textbf{IUPAC to Formula Translation. } This task takes a molecule's IUPAC name as input and converts it into its corresponding molecular formula. The model's output is a list of all elements and the number of atoms for each element in the molecule. This task is crucial for chemical informatics, as it links a molecule's systematic name to its elemental composition.
\end{itemize}

\textbf{Evaluation Metric.} We evaluate performance using two metrics: \textit{Top-1 Split Match} and \textit{Top-1 Element Match}.  

\begin{itemize}
    \item \textbf{Split Match.} This metric assesses correctness by parsing the predicted formula $P$ and the ground-truth formula $G$ into structured components according to a predefined splitting rule. Let $\text{judge\_string\_split\_match}(P, G)$ denote a Boolean function that returns true if $P$ and $G$ are equivalent under this split-based comparison. The metric is computed as  
    \[
        M_{\mathrm{split}} = \frac{1}{N} \sum_{i=1}^{N} \mathbb{I}\!\left(\exists j \;\; \text{judge\_string\_split\_match}(P_i, G_{i,j}) \right),
    \]
    where $\mathbb{I}(\cdot)$ is the indicator function, $P_i$ the $i$-th prediction, and $\{G_{i,j}\}$ the corresponding set of references.

    \item \textbf{Element Match.} This metric provides a more permissive comparison by disregarding element order and only checking whether the predicted and reference formulas contain identical counts of each element. For example, $\mathrm{C}5\mathrm{H}8$ and $\mathrm{H}8\mathrm{C}5$ are considered equivalent. Define $C(F)$ as the function that maps a formula $F$ to its multiset of element counts. The metric is given by  
    \[
        M_{\mathrm{element}} = \frac{1}{N} \sum_{i=1}^{N} \mathbb{I}\!\left(\exists j \;\; C(P_i) = C(G_{i,j}) \right).
    \]
\end{itemize}

\begin{figure}[t]
\begin{tcolorbox}[title={Molecular Captioning Case}]
\textbf{Input}\\
\texttt{Please provide a brief introduction to this molecule. <SMILES> CCCCCC/C=C\textbackslash CCCCCCCC(=O)OC[C@H](COP(=O)([O-])OCC[N+](C)(C)C)OC(C)=O </SMILES>}

\medskip
\renewcommand{\arraystretch}{1.08}
\begin{tabularx}{\linewidth}{@{} l Y @{}}
\toprule
\textbf{Reference} &
The molecule is a 1,2-diacyl-sn-glycero-3-phosphocholine in which the acyl groups at positions 1 and 2 are specified as (9Z)-hexadecenoyl and acetyl respectively. It has a role as a mouse metabolite. It derives from a palmitoleic acid. \\
\midrule
\textbf{Our model} &
The molecule is a \Match{1,2-diacyl-sn-glycero-3-phosphocholine} in which the 1- and 2-acyl groups are \Match{palmitoleoyl} and \Match{acetyl} respectively. It has a role as a \Match{mouse metabolite}. It is \PMatch{functionally related to a palmitoleic acid}. \\
\midrule
\textbf{GPT-5} &
This molecule is an unsaturated analog of platelet-activating factor (PAF), consisting of a glycerol backbone esterified at the sn-1 position with \Match{hexadec-9-enoic acid}, \Match{acetylated at the sn-2 position}, and linked at the sn-3 position to a \PMatch{phosphocholine group}. The presence of a cis double bond in the long acyl chain introduces unsaturation, which may influence its membrane dynamics and biological activity. As a member of the PAF family, it is potentially involved in inflammatory signaling and immune modulation. \\
\bottomrule
\end{tabularx}
\end{tcolorbox}
\caption{Comparison of molecular captioning outputs.}\label{fig:molecular_captioning_case}
\end{figure}

\paragraph{Natural Language Translation of Molecule.} 
To bridge the gap between chemical representations and human-readable knowledge, we evaluate tasks that translate molecular structures into natural language descriptions. 

\begin{itemize}
    \item \textbf{Molecular Description. } This is a generative task designed to describe a molecule based on its chemical structure. The model takes a molecule's SMILES string as input. Its goal is to generate a natural language description that includes the molecule's properties, functions, and a summary of its effects. This task is crucial for making complex chemical information accessible and understandable to a broader audience. This task comes from Mol-Instructions~\cite{fang2023mol}.
    \item \textbf{Molecular Captioning. } Similar to the Molecular Description task, this task requires the model to output the SMILES string's explanation, using data from SMolInstruct~\cite{yu2024llasmol}. 
\end{itemize}

\textbf{Evaluation Metric.} We use ROUGE-L and Metric for Evaluation of Translation with Explicit ORdering (MENTOR) as metrics. 

As shown in Fig.~\ref{fig:molecular_captioning_case}, our model matches the reference on every core fact while avoiding unsupported inferences. It correctly identifies the lipid class as a 1,2-diacyl-sn-glycero-3-phosphocholine, specifies the exact acyl substituents (palmitoleoyl at sn-1 and acetyl at sn-2), reproduces the biological role (mouse metabolite), and links it to palmitoleic acid—all with normalized terminology and no embellishment. 

\begin{figure}[t]
    \centering
    \includegraphics[width=1.0\linewidth]{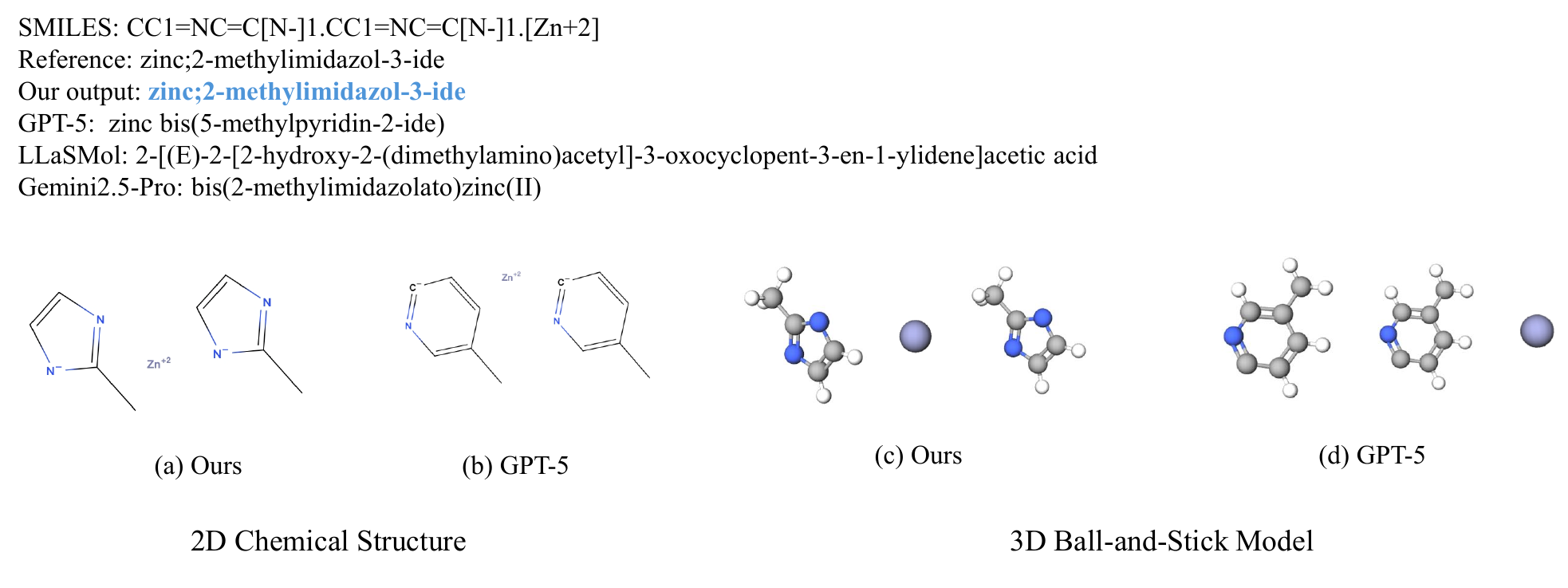}
   \caption{\textbf{Case study of ``ZIF-8'' in Metal Organic Framework
(MOF)}.}
    \label{fig:mol_translation_na}
\end{figure}

\begin{figure}
    \centering
    \includegraphics[width=1.0\linewidth]{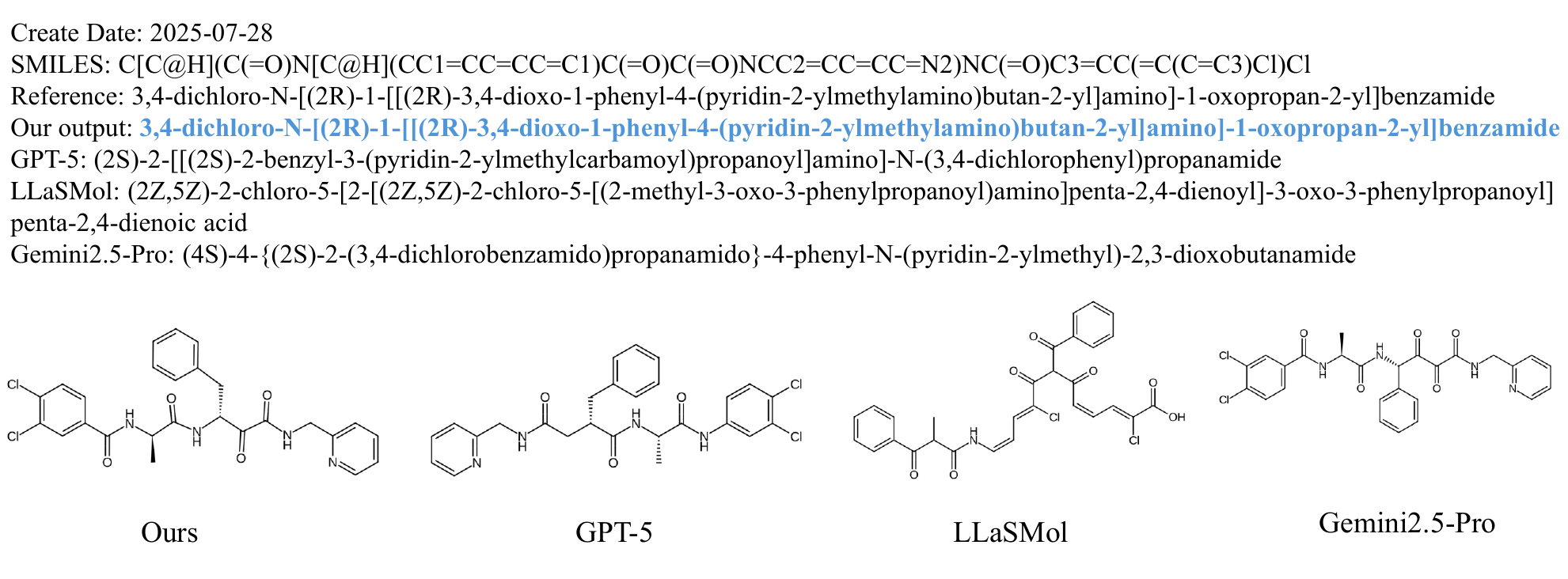}
   \caption{\textbf{Case study of SMILES\texorpdfstring{$\to$}{→}IUPAC from latest PubChem data.}
  Given the input SMILES (top), our model generates the exact reference IUPAC name (\textit{shown in blue}), preserving the 3,4-dichloro-benzamide core, the pyridin-2-ylmethylamino 1,3-dicarbonyl motif, and the (2R)/(2R) stereochemistry; the reconstructed structure (bottom left) is therefore congruent with the input. In contrast, competing LLMs (GPT-5, LLaSMol, Gemini-2.5-Pro) yield chemically inconsistent names that alter substituents, lose halogens, change chain length, or flip chirality, leading to mismatched reconstructions.}
    \label{fig:mol_translation}
\end{figure}

As shown in Fig.~\ref{fig:mol_translation_na}, we visualize the generated 2D chemical structure and 3D ball-and-stick model for CC1=NC=C[N-]1.CC1=NC=C[N-]1.[Zn+2], which is called ``ZIF-8‘' in Metal Organic Framework (MOF). Since MOF is a coordination compound, there is a ``Coordinate bond'' in MOF. 
Our method generates the IUPAC name of ZIF-8 successfully, while GPT-5 could not. Moreover, LLaSMol and Gemini2.5-Pro generate IUPAC names without meaningful 2D chemical structures.
In addition to ``Coordinate bond'', we also evaluate our model for generating molecules with only ``covalent bonds''.
As depicted in Fig.~\ref{fig:mol_translation}, our method is able to successfully translate SMILES to IUPAC name.
However, the IUPAC name generated by GPT-5 yields the opposite chiral isomer, and the IUPAC names generated by LLaSMol and Gemini2.5-Pro are not consistent with the SMILES.

\paragraph{Protein Functional Translation.}
For the given protein sequence, we prompt the model to describe the function of the protein in a clear and coherent manner within the context of the natural language domain. It thus tests both the model's reasoning when inferring functional roles from sequence data and its proficiency at producing accurate, fluent text. 
These tasks include four different data for testing, \textbf{(i)} CASPSimilarSeq: the sequences that are similar to the sequences released by CASP14. \textbf{(ii)} IDFilterSeq: this version uses ID-filtered sequences to reduce redundancy. \textbf{(iii)} UniProtSeq: this version's protein sequences are from the UniProt database. \textbf{(iv)} Mol-ins: The protein sequence of this version is derived from the Mol-Instructions~\cite{fang2023mol} benchmark. Additionally, this test set requires the model to also output subcellular localization information. We therefore consider the above four data sources are four sub-tasks of Protein Functional Translation. 

\paragraph{Summary of Evaluation Results.} 
Our models consistently outperform baselines. The \ProjectName-8B model achieves near-ceiling accuracy in formula recovery (>93\%), markedly improves challenging SMILES-IUPAC translations, and delivers the highest scores in molecular captioning (MENTOR 0.61) and protein functional translation (ROUGE-L up to 0.98). 
These results highlight strong semantic fidelity and scalability, establishing a new frontier for scientific translation tasks.

\subsection{Scientific Knowledge Extraction and Question Answering}

\setlength{\extrarowheight}{1pt}
\begin{table}[t]
\centering
\scriptsize
\resizebox{\linewidth}{!}{
\begin{tabular}{l c c c c c c c}
\toprule
Task & Metric & \makecell[c]{Gemini\\2.5-pro} & \makecell[c]{GPT\\oss} & \makecell[c]{Non-LLM\\Experts}& \makecell[c]{Specialist\\LLMs} & \makecell[c]{\ProjectName\\1.7B} & \makecell[c]{\ProjectName\\8B} \\
\midrule
\addlinespace[0.8em]
\multicolumn{8}{l}{\raisebox{1.2ex}[0pt][0pt]{\bfseries Scientific Knowledge Extraction and QA}}\\[-0.8ex]
\quad Chemical Entity Recognition & F1 $\uparrow$ & 0.71 & 0.55 & \textemdash{} & 0.75 & \SecBestPerf{0.81} & \BestPerf{0.92} \\
\quad Chemical–Protein Interaction Extraction & F1 $\uparrow$ & 0.12 & 0.05 & \textemdash{} & 0.22 & \SecBestPerf{0.46} & \BestPerf{0.83} \\
\quad Chemical–Disease Interaction Extraction & F1 $\uparrow$ & 0.35 & 0.23 & \textemdash{} & 0.40 & \SecBestPerf{0.66} & \BestPerf{0.97} \\
\quad General Multiple Choice Question & ACC $\uparrow$ & \SecBestPerf{0.94} & 0.87 & \textemdash{}  & 0.65 & {0.92} & \BestPerf{0.99} \\
\quad True/False Question & ACC $\uparrow$ & 0.62 & \SecBestPerf{0.67} & \textemdash{} & 0.55 & {0.64} & \BestPerf{0.68} \\
\quad Open-Ended Question & BertScore $\uparrow$ & \SecBestPerf{0.84} & 0.80 & \textemdash{} & \SecBestPerf{0.84} & 0.83 & \BestPerf{0.87} \\
\bottomrule
\end{tabular}
}
\caption{Evaluation results of scientific knowledge extraction and QA tasks. 
\BestPerf{Bold} indicates the best performance, and \SecBestPerf{underline} indicates the second best. The names of expert models and specialist LLMs for each task are provided in Appendix Table.~\ref{tab:appendix_scientific_knowledge_extraction}.}
\label{tab:scientific_knowledge_extraction}
\end{table}

Unlike our sequence-based tasks that require models to process structured biological inputs such as DNA, RNA, protein sequences, or SMILES strings, the tasks grouped under Scientific Knowledge Extraction and QA focus purely on natural language understanding and reasoning. They evaluate our model's ability to extract scientific knowledge, recognize scientific entities, and answer domain-specific questions directly from unstructured text. Furthermore, this isolates linguistic comprehension from sequence modeling, allowing us to probe whether models can capture the semantics of scientific literature, resolve complex entity relations, and deliver accurate answers across diverse contexts. The results are listed in Tab.~\ref{tab:scientific_knowledge_extraction}.

\paragraph{Chemical Entity Recognition.} This task entails detecting textual mentions of chemical substances from the inputs and assigning them to specific entities (e.g., chemical compounds). This poses challenges for models since the heterogeneity of chemical nomenclature. Accordingly, the performance of this task provides a sensitive probe of an LLM's domain competence in biomedical text understanding. 

\paragraph{Chemical-Protein Interaction Extraction} This task evaluates language models on chemical–protein interaction (CPI) annotation: given biomedical text snippets, the model must (i) detect mentions of chemical entities and protein/gene targets, (ii) link candidate pairs, and (iii) assign an interaction label that captures mechanism and polarity (e.g., binding). The task requires grounding each prediction to supporting evidence spans and correctly handling context , thereby probing biochemical and pharmacological knowledge under realistic nomenclature variability. 

\paragraph{Chemical-Disease Interaction Extraction} This task probes extraction of chemical-induced disease (CID) relations from biomedical literature.   For the input, the model must (i) detect mentions of chemicals and diseases, (ii) link candidate pairs, and (iii) decide whether the passage asserts a causal, adverse relation in which exposure to the chemical induces or exacerbates the disease, citing the supporting span. Accurate CID extraction enables downstream applications in toxicology, pharmacovigilance, and disease-mechanism mapping.

\paragraph{General Multiple Choice Question} The task is to select the correct option for each question, probing factual knowledge, conceptual understanding, and applied reasoning relevant to biology, chemistry, and other relevant fields. 

\paragraph{True/False Question} Following the practice in Mol-Instructions~\cite{fang2023mol}, this task requires the language models to answer research questions with positive/negative/uncertain results, without being given the references from PubMed.

\paragraph{Open-Ended Question} Open-ended questions require free-form responses to input content related to biomedicine, without following a fixed format. Open-ended QA tests genuine biomedical understanding, demands clear self-contained explanations, mirrors real use. 

\paragraph{Summary of Evaluation Results} 
Scaling from \textbf{\ProjectName-1.7B} to \textbf{\ProjectName-8B} yields consistent gains across scientific knowledge extraction and QA. The \textbf{\ProjectName-8B} model surpasses prior Specialist LLMs on every tasks. The most pronounced improvements appear on relation extraction (chemical–protein and chemical–disease), reflecting stronger entity linking, synonym/acronym disambiguation, and evidence grounding; chemical entity recognition also benefits, indicating sharper domain-specific span detection. In QA, multiple-choice performance approaches a ceiling, open-ended questions and true/false questions remain stable yet competitive with strong general-purpose models. Overall, scaling primarily boosts entity-dense, relation-heavy reading while preserving high performance on general scientific QA.

\begin{table}[t]
\centering
\scriptsize
\resizebox{\linewidth}{!}{
\begin{tabular}{l c c c c c c c} 
\toprule
Task & Metric & \makecell[c]{Gemini\\2.5-pro} & \makecell[c]{GPT\\oss} & \makecell[c]{Non-LLM \\ Experts} & \makecell[c]{Specialist\\LLMs} & \makecell[c]{\ProjectName\\1.7B} & \makecell[c]{\ProjectName\\8B} \\
\midrule
\addlinespace[0.8em]
\multicolumn{8}{l}{\raisebox{1.2ex}[0pt][0pt]{\bfseries Molecular Property Regression}}\\[-0.8ex]
\quad ESOL      & RMSE $\downarrow$        & 1.28 & 3.45  & \textemdash{}  & \BestPerf{1.04}    & 1.21     & \SecBestPerf{1.08}   \\
\quad LIPO      & RMSE  $\downarrow$       & 1.21 & 1.65  & \textemdash{}  & 1.01               & \SecBestPerf{0.94}   & \BestPerf{0.82}   \\
\quad Physicochemical Prediction & MAE  $\downarrow$     & 0.027 & 1.329   & \textemdash{}  & 0.013             & \SecBestPerf{0.003}   & \BestPerf{0.002}   \\
\midrule
\addlinespace[0.8em]
% 任务分类标题跨列数从9→8
\multicolumn{8}{l}{\raisebox{1.2ex}[0pt][0pt]{\bfseries Material Property Regression}}\\[-0.8ex]
% 多行合并Metric列：删除原GPT-o3列的数值（如原“1.32\dag & 1.22\dag & 0.95”改为“1.32\dag & 0.95”）
\quad MP$_\text{Regression}$     
  & \multirow{10}{*}{$\frac{\mathbf{MAD}}{\mathbf{MAE}}$$\uparrow$}
  & 1.32\dag & 0.95 & \BestPerf{5.32} & 0.76  & 4.98|4.67\dag   & \SecBestPerf{5.12}|5.21\dag   \\
\quad SNUMAT$_\text{Regression}$ &                                   
  & 1.07\dag & 1.00 & 1.83 & 1.16  & \BestPerf{2.18}|2.18\dag   & \SecBestPerf{1.96}|2.22\dag   \\
\quad JARVISDFT                  &                                   
  & 1.09\dag & 0.88 & 4.10 & 0.70  & \SecBestPerf{5.11}|5.28\dag   & \BestPerf{5.21}|5.59\dag   \\
\quad JARVISQETB                 &                                   
  & 0.53\dag & 0.50 & 59.44 & 1.05  & \BestPerf{138.59}|148.71\dag   & \SecBestPerf{126.61}|137.61\dag  \\
\quad GNoME                      &                                   
  & 1.28\dag & 0.73 & 15.60 & 0.50  & \SecBestPerf{19.05}|18.46\dag   & \BestPerf{22.03}|21.81\dag   \\
\quad hMOF                       &                                   
  & 0.65\dag & 0.67 & \BestPerf{1.48} & 0.66  & \SecBestPerf{1.05}|1.04\dag    & {0.99}|0.99\dag    \\
\quad Cantor-HEA                &                                   
  & 0.77\dag & 0.45 & \BestPerf{8.40} & 0.87  & \SecBestPerf{7.40}|7.38\dag    & {7.18}|7.02\dag    \\
\quad QMOF                       &                                   
  & 1.24\dag & 0.86 & 2.05 & 0.93  & \BestPerf{8.55}|8.76\dag    & \SecBestPerf{8.07}|7.94\dag    \\
\quad OQMD                       &                                   
  & 1.20\dag & 0.65 & \BestPerf{6.02} & 1.16  & {5.29}|5.50\dag    & \SecBestPerf{5.43}|5.67\dag    \\
\quad OMDB                       &                                   
  & 0.82\dag & 0.93 & \BestPerf{1.55} & 1.01  & \SecBestPerf{1.40}|1.41\dag    & {1.39}|1.42\dag    \\
\midrule
\addlinespace[0.8em]
% 任务分类标题跨列数从9→8
\multicolumn{8}{l}{\raisebox{1.2ex}[0pt][0pt]{\bfseries Biology Multi-omics Property Regression}}\\[-0.8ex]
% 逐行删除原GPT-o3列的数值（如原“-0.43\dag & 0.78\dag & 3.20”改为“-0.43\dag & 3.20”）
\quad Fluorescence                    & Spearman $\uparrow$     & -0.43\dag & 3.20 & \SecBestPerf{69.00} & 2.57  & \BestPerf{69.36}|70.82\dag  & 66.24|68.49\dag  \\
\quad Stability$_\text{Biology-Instructions}$                       & Spearman $\uparrow$     & -9.15\dag & -7.79 & \BestPerf{79.00}  & 60.25  & 63.25|55.67\dag  & \SecBestPerf{64.41}|59.03\dag  \\
\quad Thermostability                 & Spearman $\uparrow$     & 5.47\dag & 6.29  & \BestPerf{78.00} & 45.07  & 56.25|53.17\dag  & \SecBestPerf{58.55}|54.45\dag  \\
\quad Enhancer Activity Prediction    & PCC $\uparrow$          & -4.24\dag & -1.29 & \BestPerf{68.00} & 57.24  & \SecBestPerf{65.07}|70.20\dag  & {64.39}|52.08\dag  \\
\quad APA Isoform Prediction          & R2 $\uparrow$           & 0.13\dag & 0.01  & 50.82 & 59.01  & \BestPerf{87.94}|86.19\dag  & \SecBestPerf{85.87}|87.92\dag  \\
\quad Mean Ribosome Loading Prediction& R2 $\uparrow$           & 0.00\dag & 0.02  & \BestPerf{78.00} & 47.64  & 50.82|54.56\dag  & \SecBestPerf{60.44}|64.5\dag  \\
\quad Programmable RNA Switches       & R2 $\uparrow$           & 0.01\dag & 0.01  & \BestPerf{55.67} & 26.65  & 39.00|38.1\dag  & \SecBestPerf{43.55}|29.20\dag  \\
\quad CRISPR On Target Prediction     & Spearman $\uparrow$     & -3.64 & 6.21  & \BestPerf{44.10} & 2.87  & 14.35  & \SecBestPerf{25.81}  \\
\quad siRNA Efficiency Prediction     & Mixed-score $\uparrow$  & 38.75\dag & 40.99 & 49.38 & 42.92  & \BestPerf{62.56}|64.13\dag  & \SecBestPerf{59.67}|61.94\dag  \\
\bottomrule
\end{tabular}
}
\caption{Evaluation results of property prediction tasks. 
\BestPerf{Bold} indicates the best performance, and \SecBestPerf{underline} indicates the second best. The names of expert models and specialist LLMs for each task are provided in Appendix Table.~\ref{tab:appendix_property_prediction}. \dag~denotes that only randomly sampled 1000 test samples were utilized for evaluation in the corresponding task. This measure is adopted because extremely large-scale test sets exist for certain tasks, which would lead to substantial testing costs.}
\label{tab:property_prediction}
\end{table}

\subsection{Property Prediction Tasks}
We benchmark our single instruction-tuned model on a unified suite of composition-to-property and sequence-to-property regressions spanning materials, molecules, and multi-omics. Inputs include chemical formulas, SMILES, and DNA/RNA/protein sequences, with targets covering structural, electronic, thermodynamic, and biochemical properties. The results are listed in Tab.~\ref{tab:property_prediction}.

\paragraph{Molecular Property Regression.}  
We consider three molecular regression tasks from SMolInstruct~\cite{yu2024llasmol} and Mol-Instructions~\cite{fang2023mol}: \ding{172} ESOL Prediction, \ding{173} LIPO Prediction, and \ding{174} Physicochemical Prediction. These tasks perform single-output regression to quantitatively predict fundamental molecular characteristics directly from biochemical representations. 
\begin{itemize}
    \item \textbf{Estimated Solubility (ESOL) Prediction}. The ESOL Property Prediction task  on property prediction for a molecule. The model receives a chemical structure in the form of a SMILES string. Its goal is to predict the aqueous solubility for that molecule. This is a regression task, requiring a single continuous value as output. The predicted value represents the logarithm of solubility (logS) in mol/L.
    \item \textbf{Lipophilicity (LIPO) Prediction}. The LIPO Property Prediction task predicts the octanol-water distribution coefficient (logD). The model takes a molecule's SMILES string as input. It calculates the logD value under the specific condition of pH 7.4. The final output is a single numerical value for this coefficient.
    \item \textbf{Physicochemical Prediction.} Physicochemical prediction leverages a molecule's structural details to estimate its core physicochemical traits. By inferring unmeasured attributes, it guides the early prioritization of promising candidates and reduces experimental burden. 
\end{itemize} 

The above task takes an SMILES string as input, and then requires the model to output a specific property, as shown in the following example for the Physicochemical Prediction task. 

\begin{framed}
        \small\textit{Example:}\\
        \texttt{Instructions: Could you give me the LUMO energy value of this molecule? <SMILES> CC1=CC2C(CO)C2C1 </SMILES>}\\
        \texttt{Response: 0.0191}
\end{framed}

\textbf{Evaluation Metric.} For ESOL Prediction and LIPO Prediction, we use RMSE as the evaluation metric. For Physicochemical Prediction, we use MAE as the evaluation metric. 

\paragraph{Material Property Regression}
Follow LLM4MAT~\cite{rubungo2025llm4mat}, we evaluate the model using data sources from multiple databases, i.e.,  \ding{172} MP$_\text{regression}$~\cite{jainmaterials}, \ding{173} SNUMAT$_\text{regression}$~\cite{snumat}, \ding{174} JARVIS-QETB~\cite{garrity2023fast}, \ding{175} JARVIS-DFT~\cite{choudhary2017high, choudhary2018computational}, \ding{176} GNoME~\cite{merchant2023scaling}, \ding{177} Cantor-HEA~\cite{li2024efficient}, \ding{178} QMOF~\citep{rosen2021machine, rosen2022high}, \ding{179} OQMD~\citep{kirklin2015open},\ding{180} OMDB~\cite{kirklin2015open}, and \ding{181} hMOF~\cite{wilmer2012large}, organized into a ``structure-to-property'' task designed to predict various crystal properties based on input chemical formulas. 
\begin{itemize}
    \item For Materials Project (MP), we predict band gap (eV), unit‐cell volume (Å$^3$), formation energy per atom (eV/atom), energy above hull (eV/atom), total energy per atom (eV/atom), thermodynamic stability flag, mass density (g/cm³), atomic density (atoms/Å$^3$) and Fermi energy (eV). 
    \item For SNUMAT, we predict both HSE and GGA band gaps (eV), the corresponding optical band gaps (eV), spin–orbit coupling flag. 
    \item For Jarvis-QETB, we predict energy per atom (eV/atom), indirect gap (eV), Fermi energy (eV) and final energy (eV). 
    \item For Jarvis-DFT, our model outputs formation energy per atom (eV/atom), band gap (eV), spectroscopic limited maximum efficiency (SLME), electronic spillage, total energy (eV), static dielectric constant (x-direction) at the meta-GGA TBmBJ level of theory (MEPSX), maximum electric field gradient (V/Å²), average electron mass, maximum piezoelectric coefficient (C/m² and pC/N), dielectric constant, n- and p-type Seebeck coefficients (µV/K), power factors (µW/cm·K²), exfoliation energy (meV/atom), bulk and shear moduli (GPa), modified Becke–Johnson band gap (eV) and energy above hull (eV/atom).

    \item For GNoME, we predict formation and decomposition energies (eV/atom), band gap (eV), corrected total energy (eV), volume (Å³) and density (g/cm³). 
    \item For Cantor-HEA, we estimate formation energy per atom (eV/atom), energy above hull (eV/atom), volume per atom (Å$^3$/atom) and energy per atom (eV/atom).

    \item For QMOF, we predict total energy (eV), band gap (eV), largest cavity diameter and pore‐limiting diameter (both Å). 
    \item For OMDB and OQMD, we predict band gap (eV) and, we also predict formation energy (eV/atom) for OQMD. 
    \item For hMOF, we predict maximum and minimum CO$_2$ adsorption capacities (mmol/g), largest cavity diameter (Å), pore‐limiting diameter (Å), void fraction, and both gravimetric (m$^2$/g) and volumetric (m$^2$/cm$^3$) surface areas.

\end{itemize}

We follow the instruction template in LLM4MAT~\cite{rubungo2025llm4mat}. 
\begin{framed}
\small\textit{Example:}\\
\texttt{Instructions: You are a material scientist. Look at the chemical composition and predict the requested property. Respond in a JSON format. \\chemical formula: KPrMnNbO6 property name: volume}\\
\texttt{Response: \{volume : 128.62405239223665\}}
\end{framed}

\textbf{Evaluation Metric.} For the above tasks, the metric is: 
\begin{equation}
    \text{MAD}=\frac{1}{n}\sum_{i=1}^n|y_i-\bar y|
\qquad
\text{MAE}=\frac{1}{n}\sum_{i=1}^n|\hat y_i- y_i|
\qquad
\frac{\text{MAD}}{\text{MAE}}=\frac{\sum_{j=1}^m 
|S_j|\frac{\text{MAD}_j}{\text{MAE}_j}}{\sum_{j=1}^m 
|S_j|}
\end{equation}

\begin{figure}[t]
\centering
\begin{subfigure}{0.48\textwidth}
\includegraphics[width=\linewidth]{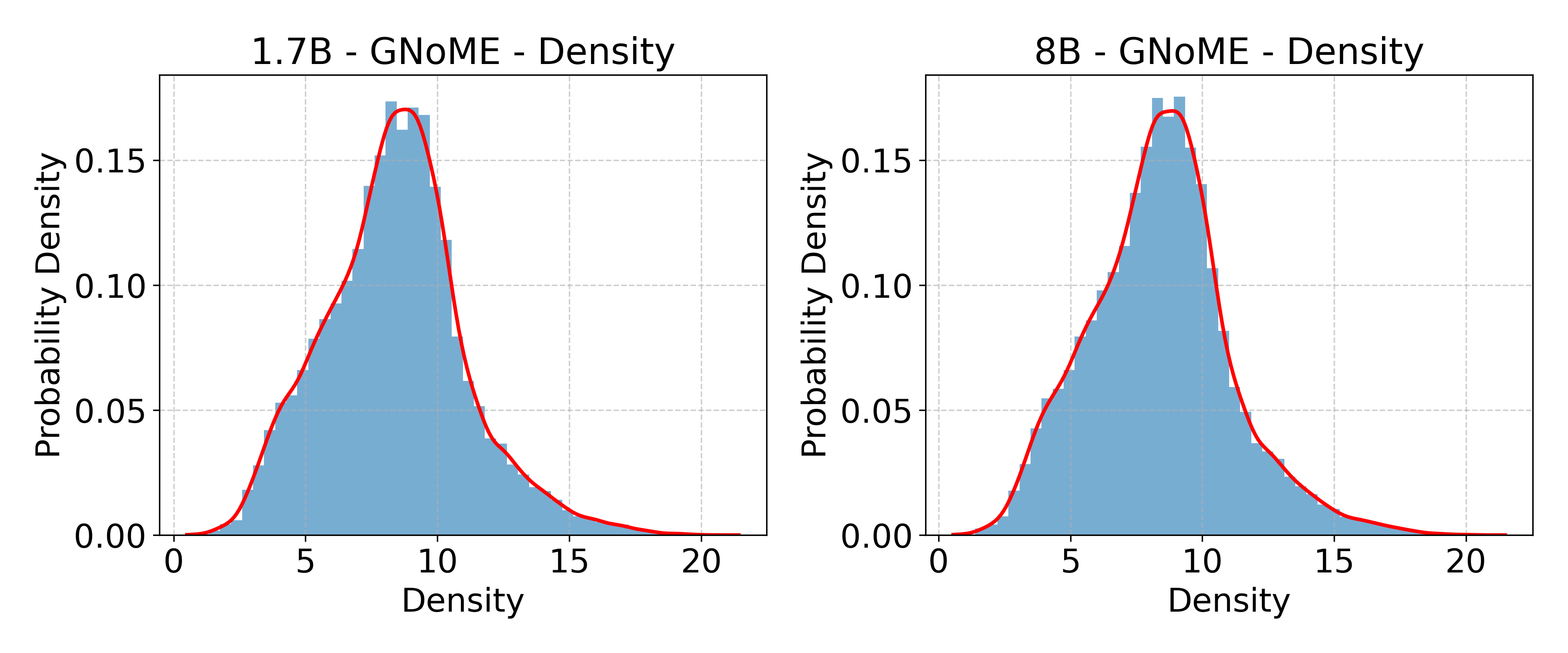}
\subcaption{\small Density}
\end{subfigure}\hfill
\begin{subfigure}{0.48\textwidth}
\includegraphics[width=\linewidth]{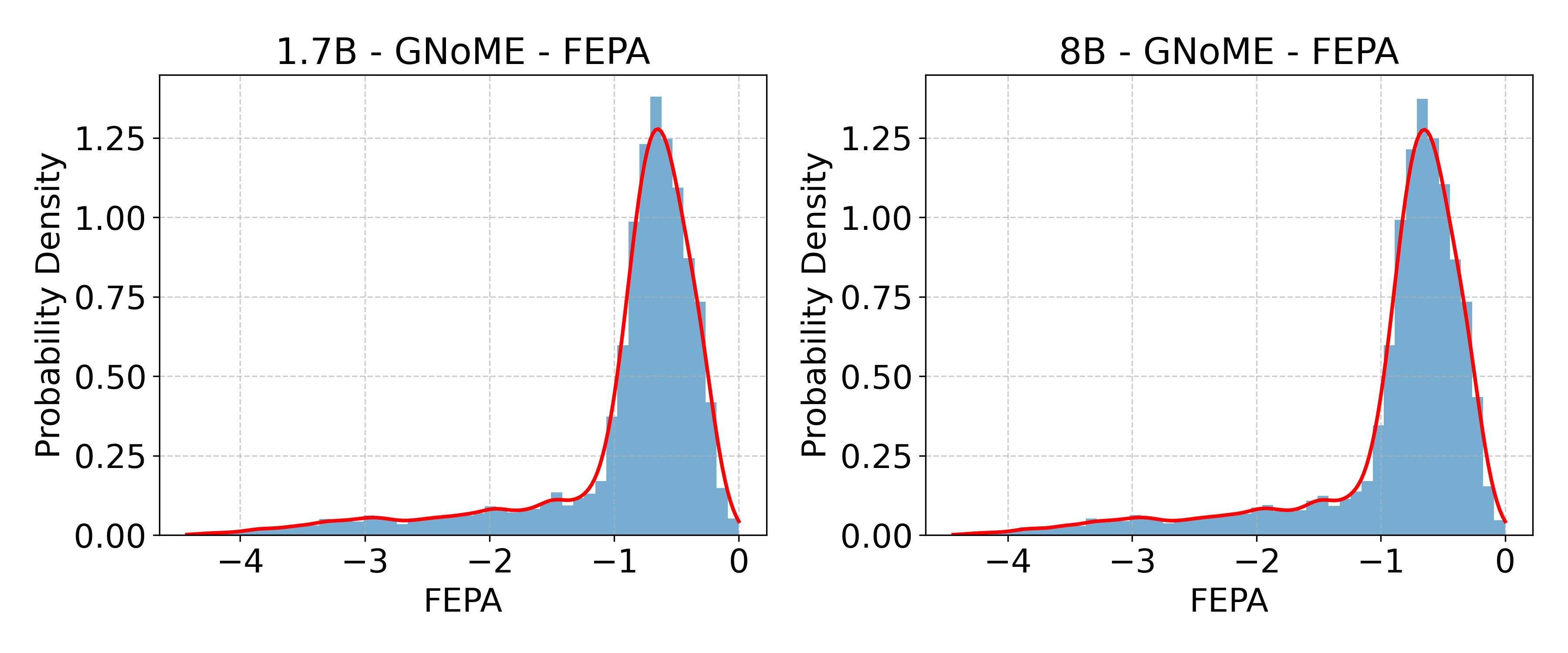}
\subcaption{\small FEPA}
\end{subfigure}
\caption{GNoME distributions (Density vs FEPA)}
\label{fig:material_regression_figure_a}
\end{figure}

\begin{figure}[t]
\centering
\begin{subfigure}{0.48\textwidth}
\includegraphics[width=\linewidth]{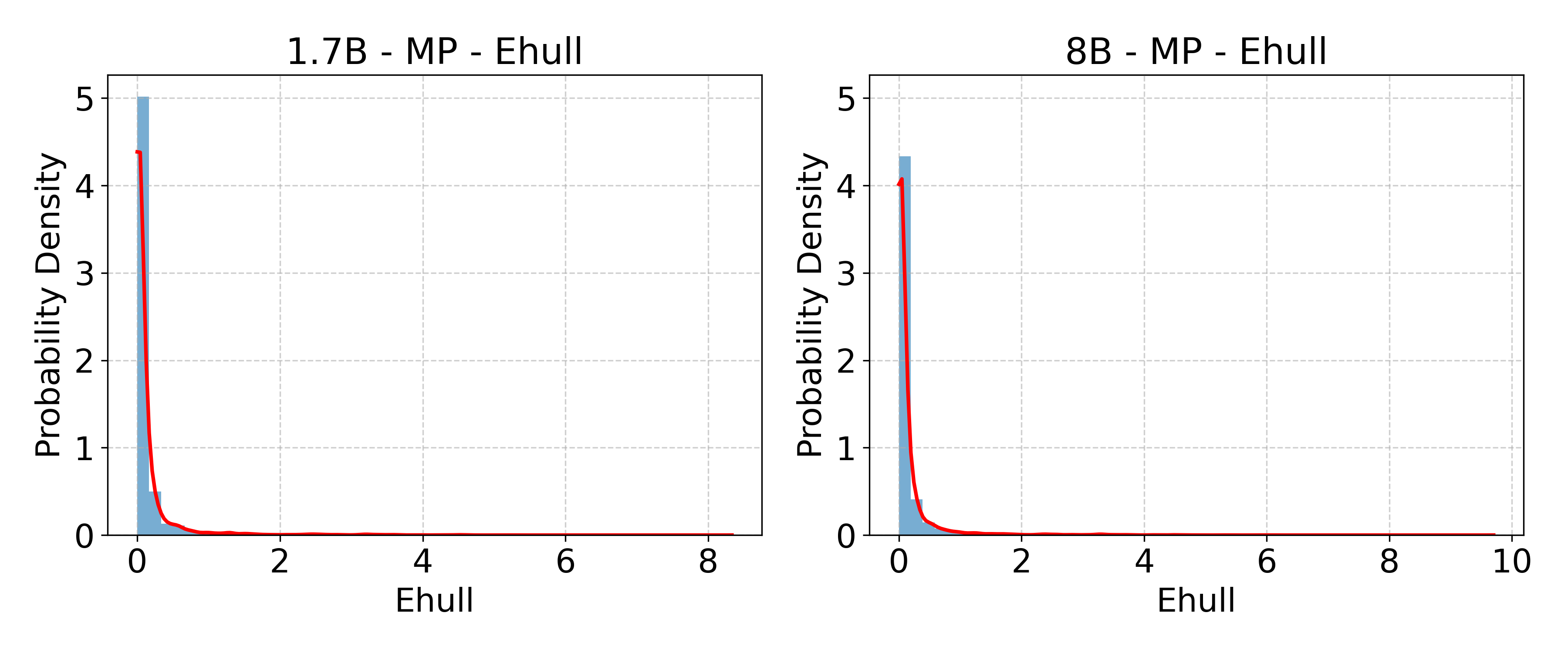}
\subcaption{\small Ehull}
\end{subfigure}\hfill
\begin{subfigure}{0.48\textwidth}
\includegraphics[width=\linewidth]{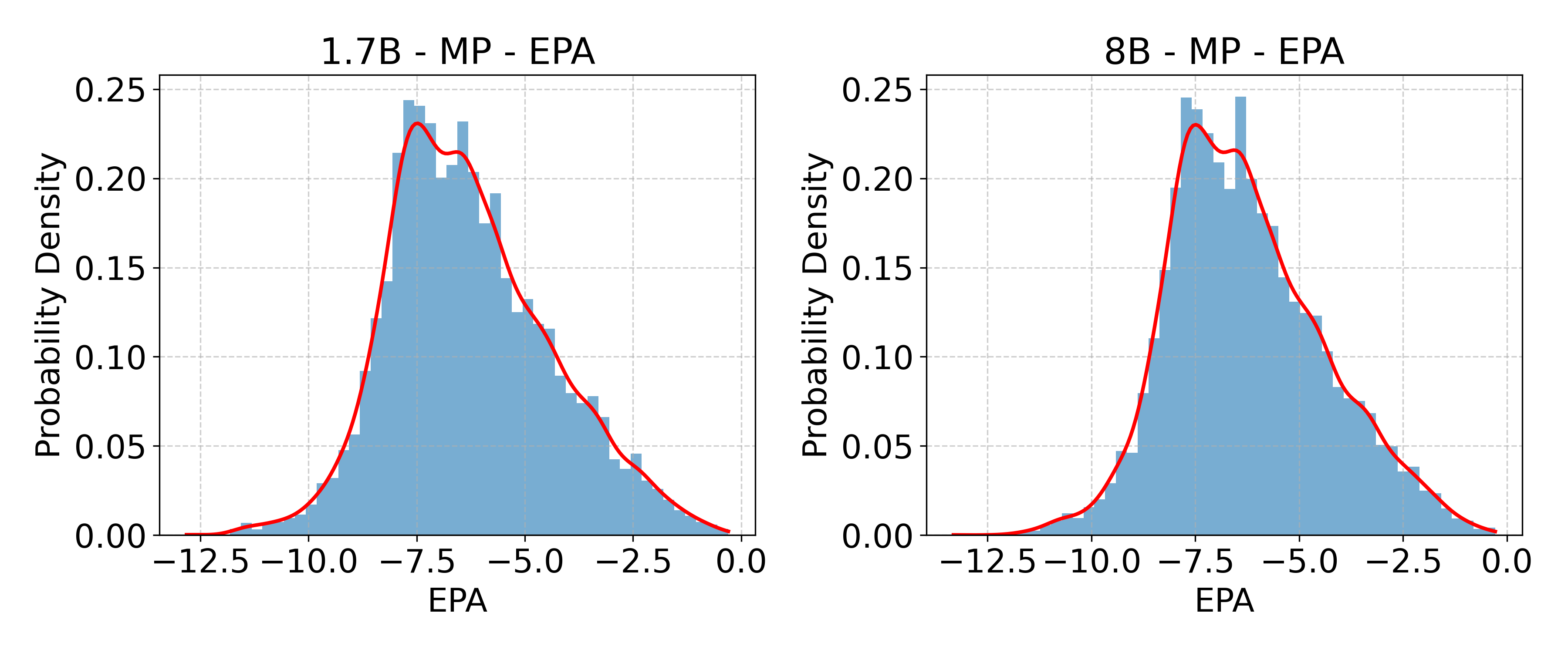}
\subcaption{\small EPA}
\end{subfigure}
\caption{Materials Project distributions (Ehull vs EPA)}
\label{fig:material_regression_figure_2}
\end{figure}

As shown in Fig.~\ref{fig:material_regression_figure_a}, and Fig.~\ref{fig:material_regression_figure_2}, our model is capable of predicting a broad spectrum of structural, electronic and thermodynamic properties across multiple materials databases. 

\begin{figure}[t]
\centering
\begin{subfigure}{0.48\textwidth}
\includegraphics[width=\linewidth]{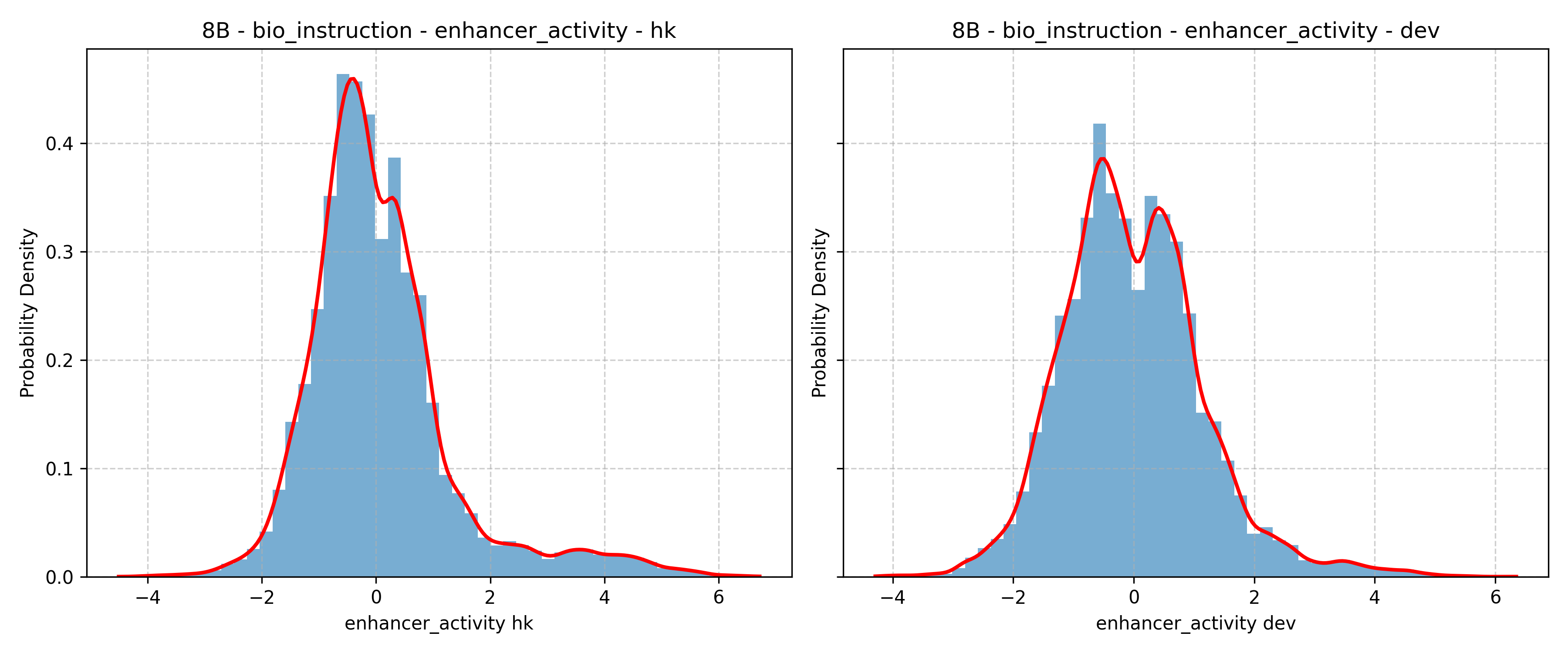}
\subcaption{\small Enhancer Activity Prediction}
\end{subfigure}\hfill
\begin{subfigure}{0.48\textwidth}
\includegraphics[width=\linewidth]{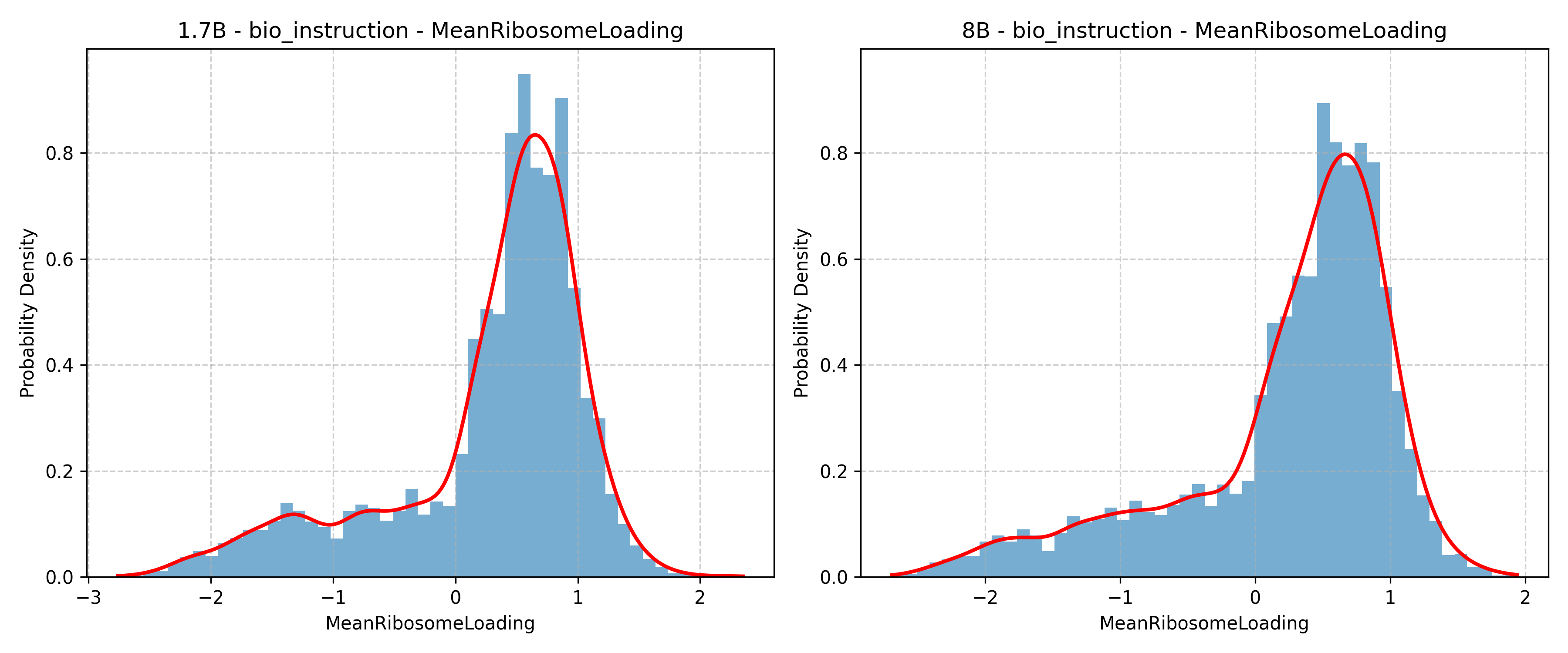}
\subcaption{\small Mean Ribosome Loading Prediction}
\end{subfigure}
\caption{Biology Multi-omics distributions. }
\end{figure}

\paragraph{Biology Multi-omics Property Regression} 
For multi-omics (i.e., DNA, DNA and protein) tasks, we consider 9 regression tasks in Biology-Instructions~\cite{he2024biology}: 
\ding{172} Fluorescence Prediction, 
\ding{173} Stability Prediction, 
\ding{174} Thermostability Prediction, 
\ding{175} Enhancer Activity Prediction, 
\ding{176} Alternative Aolyadenylation (APA) Isoform Prediction,
\ding{177} Mean Ribosome Loading Prediction, 
\ding{178} Programmable RNA Switches, 
\ding{179} CRISPR On Target Prediction, and 
\ding{180} siRNA Efficiency Prediction, they perform single-output or multi-output regression to predict diverse biological properties. 

\begin{itemize}
    \item \textbf{Fluorescence Prediction}. This task presents a regression task to predict the fluorescence of higher-order green fluorescent protein (GFP) mutants. Each protein sequence is mapped to the logarithm of its fluorescence intensity.
    \item \textbf{Stability Prediction}. This task presents a regression task to predict the intrinsic stability of proteins. The model maps each protein sequence to a continuous stability score reflecting its ability to maintain its fold above a specific concentration threshold (e.g., the EC50 value). 
    \item \textbf{Thermostability Prediction}. This task is to predict protein thermostability, defined as a protein's stability at elevated temperatures. 
    \item \textbf{Enhancer Activity Prediction. } This task is a multi-output regression task to predict the activity levels of enhancer regions from DNA sequences. The model output two numeric values corresponding to the enhancer's housekeeping and developmental activity levels. 
    \item \textbf{Alternative Aolyadenylation (APA) Isoform Prediction. } This tesk presents a regression task to predict APA isoform usage from RNA sequences. The model outputs a proportion between 0 and 1 for each isoform, which represents its relative expression and captures the variability in polyadenylation signal processing. 
    \item \textbf{Mean Ribosome Loading Prediction. } This task presents a regression task to predict ribosome loading efficiency from RNA sequences. The model outputs a continuous value representing the mean ribosome loading, reported to two decimal places. Accurate prediction of ribosome loading is essential for understanding how cis-regulatory elements, particularly 5' untranslated regions (UTRs), influence translation efficiency. 
    \item \textbf{Programmable RNA Switches. } This task addresses a multi-label regression task to predict the behavior of programmable RNA switches from their sequences. The model outputs three continuous values representing the ``ON'', ``OFF'', and ``ON/OFF'' states, each reported to two decimal places. 
    \item \textbf{CRISPR On Target Prediction.} This task involves a regression task to predict the on-target knockout efficacy of single guide RNA (sgRNA) sequences for use in CRISPR systems. 
    \item \textbf{siRNA Efficiency Prediction. } The model predicts the gene-silencing efficiency of modified small interfering RNA (siRNA). Based on the siRNA and corresponding target sequences, the model outputs a numeric value representing the percentage of mRNA remaining after treatment. 
    
\end{itemize}

\textbf{Evaluation Metric.} For Fluorescence Prediction, Stability Prediction, Thermostability, and CRISPR On Target Prediction, the Spearman score is used as the evaluation metric. 
For APA Isoform Prediction, Mean Ribosome Loading Prediction, and Programmable RNA Switches, the R2 score is adopted as the evaluation metric. 

For Enhancer Activity Prediction, PCC is used as the evaluation metric:  
\begin{equation}
\text{PCC} = \frac{\sum_{i=1}^{n}(x_i - \bar{x})(y_i - \bar{y})}{\sqrt{\sum_{i=1}^{n}(x_i - \bar{x})^2} \sqrt{\sum_{i=1}^{n}(y_i - \bar{y})^2}}
\end{equation}

For siRNA Efficiency Prediction, we consider the metric of Mixed-Score: 

    \begin{align}
         \text{Mixed-Score} &= 50\% \cdot (1 - \frac{\text{MAE}}{100}) + 50\% \cdot \text{F1} \cdot (1 - \frac{\text{Range\_MAE}}{100})  \quad \text{where} \\
        \text{F1} &= 2 \cdot \frac{\text{Precision} \cdot \text{Recall}}{\text{Precision} + \text{Recall}} \quad \text{and} \quad \text{Range\_MAE} = \frac{1}{m} \sum_{j=1}^{m} |y_j - \hat{y}_j|
    \end{align}

\paragraph{Summary of Evaluation Results} 
Across \emph{materials}, \emph{molecules}, and \emph{multi-omics}, scaling our  model from \textbf{\ProjectName-1.7B} to \textbf{\ProjectName-8B} yields consistent accuracy gains, i.e., lower MAE/RMSE for numeric regressions and higher rank-based correlations (e.g., Spearman/PCC) for sequence-to-property tasks. 
Relative to baselines, the \textbf{\ProjectName-8B} model \emph{matches or surpasses} the strongest non-LLM encoder baselines and/or Specialist LLMs on multiple benchmarks in Tab.~\ref{tab:property_prediction}, especially on long-context or multi-output targets.

\begin{itemize}
  \item \textbf{Materials (composition\(\rightarrow\)property).} The \ProjectName-8B model improves across MP, SNUMAT, JARVIS, GNoME, Cantor-HEA, OQMD/OMDB, QMOF, and hMOF, with notable MAE drops on harder thermodynamic/electronic properties (e.g., energy-above-hull, band gap, density/surface-area). On a subset of properties (see Tab.~\ref{tab:property_prediction}), \ProjectName-8B ties or outperforms encoder-based baselines / Specialist LLMs.
  \item \textbf{Molecules (SMILES\(\rightarrow\)property).} For \emph{ESOL} and \emph{LIPO} (RMSE\(\downarrow\)) and \emph{physicochemical} endpoints (MAE\(\downarrow\)), \ProjectName-8B consistently surpasses \ProjectName-1.7B, with the largest gains on endpoints sensitive to subtle substructure/charge effects. In several cases, \ProjectName-8B attains or exceeds Specialist LLMs, while \ProjectName-1.7B remains competitive but lags at the extremes (very hydrophilic/hydrophobic regimes).
  \item \textbf{Biology multi-omics (DNA/RNA/protein\(\rightarrow\)property).} The \ProjectName-8B model yields broad lifts in correlation-based metrics—\emph{Fluorescence}, \emph{Stability/Thermostability}, \emph{CRISPR on-target}, and \emph{Mean Ribosome Loading}—and improves composite outcomes such as the \emph{siRNA Mixed-Score}. The largest \ProjectName-1.7B\(\rightarrow\)\ProjectName-8B jumps appear in multi-output or context-dependent tasks (e.g., \emph{Enhancer Activity} with housekeeping/developmental readouts, \emph{APA isoform usage}).
\end{itemize}

\subsection{Property Classification Tasks}
Property classification tasks evaluate a model's capability to assign discrete categorical labels to small molecular, material, genetic, and multi-omics inputs, thereby enabling systematic assessment of structure–property relationships across various domains. We show the results of property classification in Tab.~\ref{tab:mat_gene_cls} and Tab.~\ref{tab:bio_prop_cls_long}.

\begin{table}[t]
\centering
\scriptsize
\begin{tabular}{l c c c c c c c} 
\toprule
Task & Metric & \makecell[c]{Gemini\\2.5-pro} & \makecell[c]{GPT\\oss} & \makecell[c]{Non-LLM \\ Expert Models} & \makecell[c]{Specialist\\ LLMs} & \ProjectName-1.7B & \ProjectName-8B \\
\midrule
\addlinespace[0.8em]
\multicolumn{8}{l}{\raisebox{1.2ex}[0pt][0pt]{\bfseries Molecular  Classification}}\\[-0.8ex]
\quad BBBP      & ACC@1 $\uparrow$      & 71.57\% & 46.70\%  & \textemdash{}  & 75.10\%    & \SecBestPerf{79.70\%}   & \BestPerf{82.74\%}   \\
\quad Clintox   & ACC@1 $\uparrow$      & 26.39\% & 52.78\%  & \textemdash{}  & \BestPerf{93.10\%}    & \SecBestPerf{92.36\%}   & 91.67\%    \\
\quad HIV       & ACC@1 $\uparrow$      & 26.7\%\dag & 69.86\%  & \textemdash{}  & 96.70\%    & \SecBestPerf{96.79\%}|92.40\%\dag   & \BestPerf{96.81\%}|95.90\%\dag   \\
\quad Sider     & ACC@1 $\uparrow$      & 67.7\%\dag & 38.25\%  & \textemdash{}  & \BestPerf{70.70\%}    & \SecBestPerf{70.00\%}|66.20\%\dag & {68.53\%}|68.70\%\dag   \\
\midrule
\addlinespace[0.8em]
\multicolumn{8}{l}{\raisebox{1.2ex}[0pt][0pt]{\bfseries Material Property Classification}}\\[-0.8ex]
\quad MP$_\text{classification}$                 & AUC$\uparrow$        & 0.58\dag & 0.53 & \BestPerf{0.72} & 0.51 & \SecBestPerf{0.70}|0.69\dag  & \BestPerf{0.72}|0.71\dag \\
\quad SNUMAT$_\text{classification}$             & AUC$\uparrow$        & 0.59\dag & 0.60 & \BestPerf{0.72} & 0.58 & \SecBestPerf{0.69}|0.69  & {0.66}|0.66 \\
\midrule
\addlinespace[0.8em]
\multicolumn{8}{l}{\raisebox{1.2ex}[0pt][0pt]{\bfseries Gene-centric Association Classification}}\\[-0.8ex]
\multirow{3}{*}{\quad gSymbol2Tissue}    & Precision$\uparrow$ & 0.01 & 0.03 & \textemdash{} & \SecBestPerf{0.39}      & \BestPerf{0.41}  & \BestPerf{0.41} \\
                                         & Recall$\uparrow$    & 0.01 & 0.09 & \textemdash{} & \BestPerf{0.91}      & 0.82  & \SecBestPerf{0.83} \\
                                         & F1 score$\uparrow$  & 0.01 & 0.04 & \textemdash{} & \BestPerf{0.53}      & 0.51  & \SecBestPerf{0.52} \\
\cdashline{1-8}[.6pt/2pt] 
\multirow{3}{*}{\quad gSymbol2Cancer}    & Precision$\uparrow$ & 0.00 & 0.01 & \textemdash{} & 0.36      & \SecBestPerf{0.75}  & \BestPerf{0.81} \\
                                         & Recall$\uparrow$    &0.00  & 0.02 & \textemdash{} & 0.32      & \SecBestPerf{0.72}  & \BestPerf{0.81} \\
                                         & F1 score$\uparrow$  &0.00 & 0.01 & \textemdash{} & 0.32      & \SecBestPerf{0.73}  & \BestPerf{0.81} \\
\cdashline{1-8}[.6pt/2pt] 
\multirow{3}{*}{\quad gName2Cancer}      & Precision $\uparrow$ &0.00  & 0.01 & \textemdash{} & 0.27      & \SecBestPerf{0.68}  & \BestPerf{0.80} \\
                                         & Recall $\uparrow$    &0.00  & 0.04 & \textemdash{} & 0.26      & \SecBestPerf{0.67}  & \BestPerf{0.79} \\
                                         & F1 score $\uparrow$  &0.00  & 0.01 & \textemdash{} & 0.25      & \SecBestPerf{0.67}  & \BestPerf{0.79} \\
\bottomrule
\end{tabular}
\caption{Evaluation results of molecular classification tasks, material property classification tasks, and gene-centric association classification tasks. \BestPerf{Bold} indicates the best performance, and \SecBestPerf{underline} indicates the second best. The names of expert models and specialist LLMs for each task are provided in Appendix Table.~\ref{tab:appendix_mat_gene_cls}. \dag~denotes that only randomly sampled 1000 test samples were utilized for evaluation in the corresponding task. This measure is adopted because extremely large-scale test sets exist for certain tasks, which would lead to substantial testing costs.}
\label{tab:mat_gene_cls}
\end{table}

\paragraph{Molecular Classification}
Molecular classification tasks evaluate a model's ability to infer discrete biological or pharmacological properties of molecules directly from their structural representations, enabling rapid assessment of drug permeability, toxicity, antiviral potential, and adverse effect risks that are central to drug discovery and safety evaluation. 

\begin{itemize}
    \item \textbf{Blood-Brain Barrier Permeability (BBBP) Prediction}. The BBBP property prediction task is designed to predict whether a given molecule can penetrate the blood-brain barrier. The model uses the SMILES string of a molecule as input. It then outputs a binary classification result, indicating if the molecule is blood-brain barrier permeable or not. This is a crucial property for drug discovery, as it helps determine if a potential drug can be effective for central nervous system targets.
    \item \textbf{Clinical Toxicity (ClinTox) Prediction}. The ClinTox property prediction task is designed to assess the toxicity of a molecule. The model takes the molecule's SMILES string as input. It then provides a binary classification indicating whether the molecule is toxic or not based on its chemical structure. This prediction is vital in the early stages of drug development to filter out potentially harmful compounds.
    \item \textbf{HIV Prediction}. The HIV property prediction task determines whether a molecule can inhibit the replication of the HIV virus. The model uses a molecule's SMILES string as input. It then performs a binary classification to predict if the molecule is an HIV replication inhibitor or not. This is an essential step in identifying potential antiviral drugs for treating HIV/AIDS.
    \item \textbf{Side Effect Resource (SIDER) Prediction}. The SIDER property prediction task predicts whether a molecule causes a specific side effect. The model takes a molecule's SMILES string as input. It then provides a binary classification indicating if the molecule is associated with a given side effect. This task is crucial for drug safety, helping researchers to identify potential adverse drug reactions early in the development process.
\end{itemize}

\paragraph{Material Property Classification} 
We consider two material property classification tasks,   MP$_\text{classification}$ and SNUMAT$_\text{classification}$, which evaluate the ability of a model to classify various crystal properties of an input chemical formula. 
\begin{itemize}
    \item For the Materials Project (MP),  it consists of two sub-tasks. One of them is to predict whether the gap of the crystal is formed directly, and the other is to predict whether a crystal is stable. 
    \begin{framed}
    \small\textit{Example:}\\
    \texttt{Instructions: You are a material scientist. Look at the chemical composition and predict whether the material is stable. Respond in JSON. chemical formula: KPrMnNbO6 property name: is\_stable}\\
    \texttt{Response: \{is\_stable: False\}}
    \end{framed}
    \item For the SNUMAT dataset, it outputs direct/indirect gap flag. 
    \begin{framed}
\small\textit{Example:}\\
\texttt{Instructions: You are a material scientist. Given the chemical composition, predict whether the band gap is direct (True) or indirect (False). Respond in JSON. chemical formula: PbS property name: is\_direct}\\
\texttt{Response: \{is\_direct: True\}}
\end{framed}
\end{itemize}

\textbf{Evaluation Metric.} We use AUC as the evaluation metric for these two tasks. 

\paragraph{Gene-centric Association Classification.}  
We assess gene-centric knowledge with three multi-label tasks mapping a gene identifier (symbol or official name) to tissue expression profiles or curated gene–cancer associations as in OPI~\cite{xiao2024opi}. 

\begin{itemize}
    \item \textbf{Tissue Location Prediction from Gene Symbol (gSymbol2Tissue).} This task involves predicting the tissue expression profile of a protein-coding gene given its gene symbol. Predicting spatial expression patterns aids in understanding protein-specific biological roles and informs biomarker discovery and therapeutic targeting. 
    \item \textbf{Cancer Prediction from Gene Symbol (gSymbol2Cancer).} Given a gene symbol, this task asks the model to identify associated cancer types, using data from the Cancer Gene Census. This task evaluates the model's capability to mine gene-cancer associations from existing biomedical knowledge, supporting cancer gene discovery and precision oncology efforts.
    \item \textbf{Cancer Prediction from Gene Name (gName2Cancer).} Given a gene symbol, this task asks the model to identify associated cancer types, using data from the Cancer Gene Census. This task evaluates the model's capability to mine gene-cancer associations from existing biomedical knowledge, supporting cancer gene discovery and precision oncology efforts.
\end{itemize}

\textbf{Evaluation Metric.} We use Precision, Recall, and F1 score as the evaluation metrics for these three tasks. 

We present the case study of the gSymbol2Cancer task as follows: 

\begin{figure}[htbp]
\centering
\begin{tcolorbox}[title={Cancer Prediction from Gene Symbol Case}]
\textbf{Input}\\
\texttt{Get the cancer name that is related to the given gene symbol: PLAG1 }

\medskip
\renewcommand{\arraystretch}{1.08}
\begin{tabularx}{\linewidth}{@{} l Y @{}}
\toprule
\textbf{Reference} &
salivary adenoma, lipoblastoma \\
\midrule
\textbf{Our model} &
salivary adenoma; lipoblastoma \\
\midrule
\textbf{GPT-5} &
Pleomorphic adenoma (salivary gland).\\
\bottomrule
\end{tabularx}
\end{tcolorbox}
\caption{Case study of cancer prediction from a gene symbol (PLAG1). 
The reference lists two associated tumors: \emph{salivary adenoma} and \emph{lipoblastoma}. 
Our model reproduces both exactly (precision = 1.00, recall = 1.00, F1 = 1.00). 
GPT-5 returns only \emph{pleomorphic adenoma (salivary gland)}, which maps to the \emph{salivary adenoma} category but misses \emph{lipoblastoma} (precision = 1.00, recall = 0.50), indicating partial coverage.}
\label{fig:cancer_prediction_case}
\end{figure}

\setlength{\tabcolsep}{6pt}
\begin{table}[]
    \centering
    \scriptsize
    \begin{tabular}{l c c c c c c c}
\toprule
Task & Metric & \makecell[c]{Gemini\\2.5-pro} & \makecell[c]{GPT\\oss} & \makecell[c]{Non-LLM \\ Expert Models} & \makecell[c]{Specialist \\ LLMs} & \makecell[c]{\ProjectName\\1.7B} & \makecell[c]{\ProjectName\\8B} \\
\midrule

\addlinespace[0.8em]
\multicolumn{8}{l}{\raisebox{1.2ex}[0pt][0pt]{\bfseries Biology Multi-omics Property Classification (binary)}}\\[-0.8ex]

\quad Solubility$_\text{Biology-Instructions}$ & ACC $\uparrow$ & $51.4$\dag  & $51.22$ & \BestPerf{77.00}& $63.02$ & $62.67$|67.3\dag & \SecBestPerf{71.61}|69.4\dag \\
\quad Antibody\mbox{-}Antigen & MCC $\uparrow$ & $-5.05$\dag  & $-2.78$ & \BestPerf{54.90} & $1.06$ & 41.59|45.43\dag & \SecBestPerf{45.5}|48.84\dag \\
\quad RNA\mbox{-}Protein & MCC $\uparrow$ & $1.20$\dag  & $-5.23$ & \BestPerf{93.2}  & $74.26$ & \SecBestPerf{86.47}|85.97\dag & $84.97$|88.8\dag \\
\quad Epigenetic Marks Prediction & MCC $\uparrow$ & 0.52\dag  & 0.84 & \BestPerf{58.83} & 3.64 & 14.16|18.09\dag & \SecBestPerf{17.24}|21.26\dag \\
\quad TF\mbox{-}Mouse$_0$ & MCC $\uparrow$ & $0.20$\dag  & $4.75$ & \BestPerf{64.23} & 26.50  & 41.79|39.77\dag & \SecBestPerf{43.52}|40.42\dag \\
\quad TF\mbox{-}Mouse$_1$ & MCC $\uparrow$ & $-1.24$\dag  & $-0.38$ & \BestPerf{86.28} & 66.96  & 68.95|72.07\dag & \SecBestPerf{71.03}|75.98\dag \\
\quad TF\mbox{-}Mouse$_2$ & MCC $\uparrow$ & $0.41$\dag  & $6.99$ & \BestPerf{81.28} & 63.98  & 71.57|69.08\dag & \SecBestPerf{75.07}|68.52\dag \\
\quad TF\mbox{-}Mouse$_3$ & MCC $\uparrow$ & $2.72$\dag  & $-9.63$ & \BestPerf{73.49} & 32.02  & 44.81|45.10\dag & \SecBestPerf{52.98}|53.31\dag \\
\quad TF\mbox{-}Mouse$_4$ & MCC $\uparrow$ & $6.61$\dag  & $-2.07$ & \BestPerf{52.66} & 9.18  & 27.34|28.45\dag & \SecBestPerf{28.62}|28.20\dag \\
\quad Enhancer\mbox{-}Promoter Interaction & MCC $\uparrow$ & $-3.75$\dag  & $-1.69$ & \BestPerf{53.59} & $4.13$ & $8.18$|6.75\dag & \SecBestPerf{14.40}|8.12\dag \\
\quad Promoter\mbox{-}300$_\text{all}$ & MCC $\uparrow$ & 24.12  & -3.56 & \BestPerf{91.01} & 78.97 & 82.94 & \SecBestPerf{83.01} \\
\quad Promoter\mbox{-}300$_\text{notata}$ & MCC $\uparrow$ & 16.65  & -2.40 & \BestPerf{94.34} & 83.45 & \SecBestPerf{88.46} & 87.04 \\
\quad Promoter\mbox{-}300$_\text{tata}$ & MCC $\uparrow$ & 11.98  & 13.04 & \BestPerf{79.43} & 38.50 & 44.38 & \SecBestPerf{51.37} \\
\quad Promoter\mbox{-}core$_\text{all}$ & MCC $\uparrow$ & 12.38  & -3.57 & \BestPerf{70.92} & 59.98 & \SecBestPerf{61.49} & 62.56 \\
\quad Promoter\mbox{-}core$_\text{notata}$ & MCC $\uparrow$ & 15.64  & -6.37 & \BestPerf{71.58} & 62.85 & 61.72 & \SecBestPerf{63.52} \\
\quad Promoter\mbox{-}core$_\text{tata}$ & MCC $\uparrow$ & 13.02  & 17.97 & \BestPerf{78.15} & 39.36 & \SecBestPerf{53.67} & 46.94 \\
\quad TF\mbox{-}Human$_0$ & MCC $\uparrow$ & -3.46  & 3.24 & \BestPerf{71.99} & 56.56 & 59.33 & \SecBestPerf{61.07} \\
\quad TF\mbox{-}Human$_1$ & MCC $\uparrow$ & 12.23  & 3.85 & \BestPerf{76.06} & 61.17 & 62.22 & \SecBestPerf{63.47} \\
\quad TF\mbox{-}Human$_2$ & MCC $\uparrow$ & 2.29  & 5.19 & \BestPerf{66.52} & 46.92 & 47.80 & \SecBestPerf{48.02} \\
\quad TF\mbox{-}Human$_3$ & MCC $\uparrow$ & -8.44  & -1.27 & \BestPerf{58.54} & 34.16 & 38.96 & \SecBestPerf{40.89} \\
\quad TF\mbox{-}Human$_4$ & MCC $\uparrow$ & 7.56  & 15.06 & \BestPerf{77.43} & 51.86 & \SecBestPerf{56.43} & 56.10 \\

\quad Solubility$_\text{PEER}$ & ACC $\uparrow$ & 58.40\dag  & 10.91 & \BestPerf{74.37} & 71.40 & 70.14|70.8\dag & \SecBestPerf{73.99}|75.4\dag \\
\quad Stability$_\text{PEER}$ & ACC $\uparrow$ & 49.60\dag  & 38.21 & 58.30 & {63.50} & \BestPerf{68.41}|68.90\dag & \SecBestPerf{64.70}|64.3\dag \\
\quad Human PPI & ACC $\uparrow$ & 52.74  & 58.2 & \BestPerf{85.09} & \SecBestPerf{84.8} & {76.79} & 73.42 \\
\quad Yeast PPI & ACC $\uparrow$ & 51.52  & 44.90 & \BestPerf{66.16} & {$60.40$} & 53.81 & \SecBestPerf{60.41} \\

\midrule
\addlinespace[0.8em]
\multicolumn{8}{l}{\raisebox{1.2ex}[0pt][0pt]{\bfseries Biology Multi-omics Property Classification (multi-label)}}\\[-0.8ex]
\quad Protein Function   & ROUGE-L   & 0.20\dag  & 0.05 & {0.57} & 0.43 & \SecBestPerf{0.63}|0.63 & \BestPerf{0.97}|0.96 \\
\quad Domain/Motif       & ROUGE-L   & 0.19\dag  & 0.02 & 0.53 & 0.46 & \SecBestPerf{0.56}|0.56 & \BestPerf{0.97}|0.90 \\
\quad FunctionEC         & Fmax$\uparrow$ & $3.3$\dag  & $22.86$  & \BestPerf{88.9} & $19.79$ & 60.65|60.33\dag & \SecBestPerf{77.26}|77.56\dag \\
\quad Non\mbox{-}coding RNA Func. Classif. & ACC $\uparrow$ & $17.8$\dag  & $11.54$ & \SecBestPerf{85.73} & $63.09$ & 80.39|88.8\dag & \BestPerf{86.65}|93.4\dag \\
\quad Modification Prediction & AUC $\uparrow$ & $51.19$ & 50.05 & \BestPerf{84.00} & 59.06 & 53.06 & \SecBestPerf{59.94} \\
\quad Subcellular Localization & Accuracy$\uparrow$ & 0.02 & 0 & \textemdash{} & 0.78  & \SecBestPerf{0.84}  & \BestPerf{0.91}  \\
\quad Fold Type & Accuracy$\uparrow$ & 0.01 & 0 & \textemdash{} & \BestPerf{0.61}  & 0.34  & \SecBestPerf{0.48}  \\
\cdashline{1-8}[.6pt/2pt] % 虚线分隔：从第1列到第9列（原10列→9列）
\quad Numerical EC (New\mbox{-}392) & \makecell{Precision$\uparrow$\\Recall$\uparrow$\\F1$\uparrow$} & \makecell{0\\0\\0}  & \makecell{0\\0\\0} & \makecell{\BestPerf{0.59}\\\SecBestPerf{0.48}\\\SecBestPerf{0.49}} & \makecell{0.37\\0.34\\0.35} & \makecell{0.40\\0.39\\0.39} & \makecell{\SecBestPerf{0.55}\\\BestPerf{0.55}\\\BestPerf{0.54}} \\
\quad Numerical EC (Price\mbox{-}149) & \makecell{Precision$\uparrow$\\Recall$\uparrow$\\F1$\uparrow$} & \makecell{0\\0\\0}  & \makecell{0\\0\\0} & \makecell{\BestPerf{0.58}\\\BestPerf{0.47}\\\BestPerf{0.49}} & \makecell{0.07\\0.07\\0.07} & \makecell{0.10\\0.10\\0.10} & \makecell{\SecBestPerf{0.21}\\\SecBestPerf{0.23}\\\SecBestPerf{0.22}} \\
\cdashline{1-8}[.6pt/2pt] % 虚线分隔范围调整为9列
\quad Keywords (CASPSimilarSeq) & \makecell{Precision$\uparrow$\\Recall$\uparrow$\\F1$\uparrow$} & \makecell{0.04\\0.03\\0.03}  & \makecell{0\\0.02\\0.01} & \makecell{-\\-\\-} & \makecell{0.81\\0.74\\0.76} & \makecell{\SecBestPerf{0.83}\\\SecBestPerf{0.80}\\\SecBestPerf{0.81}} & \makecell{\BestPerf{0.86}\\\BestPerf{0.83}\\\BestPerf{0.84}} \\
\quad Keywords (IDFilterSeq) & \makecell{Precision$\uparrow$\\Recall$\uparrow$\\F1$\uparrow$} & \makecell{0.03\\0.02\\0.01}  & \makecell{0\\0.02\\0.01} & \makecell{-\\-\\-} & \makecell{0.84\\0.80\\0.82} & \makecell{\SecBestPerf{0.86}\\\SecBestPerf{0.85}\\\SecBestPerf{0.85}} & \makecell{\BestPerf{0.89}\\\BestPerf{0.89}\\\BestPerf{0.88}} \\
\quad Keywords (UniProtSeq) & \makecell{Precision$\uparrow$\\Recall$\uparrow$\\F1$\uparrow$} & \makecell{0.01\\0\\0}  & \makecell{0.01\\0.03\\0.01} & \makecell{-\\-\\-} & \makecell{0.86\\0.82\\0.83} & \makecell{\SecBestPerf{0.85}\\\SecBestPerf{0.90}\\\SecBestPerf{0.88}} & \makecell{\BestPerf{0.92}\\\BestPerf{0.91}\\\BestPerf{0.91}} \\
\cdashline{1-8}[.6pt/2pt] 
\quad GO Terms (CASPSimilarSeq) & \makecell{Precision$\uparrow$\\Recall$\uparrow$\\F1$\uparrow$} & \makecell{0\\0\\0}  & \makecell{0\\0\\0} & \makecell{-\\-\\-} & \makecell{0.76\\0.74\\0.75} & \makecell{\SecBestPerf{0.81}\\\SecBestPerf{0.81}\\\SecBestPerf{0.80}} & \makecell{\BestPerf{0.86}\\\BestPerf{0.86}\\\BestPerf{0.86}} \\
\quad GO Terms (IDFilterSeq) & \makecell{Precision$\uparrow$\\Recall$\uparrow$\\F1$\uparrow$} & \makecell{0\\0\\0}  & \makecell{0\\0\\0} & \makecell{-\\-\\-} & \makecell{0.74\\0.73\\0.72} & \makecell{\SecBestPerf{0.78}\\\SecBestPerf{0.77}\\\SecBestPerf{0.77}} & \makecell{\BestPerf{0.83}\\\BestPerf{0.82}\\\BestPerf{0.81}} \\
\quad GO Terms (UniProtSeq) & \makecell{Precision$\uparrow$\\Recall$\uparrow$\\F1$\uparrow$} &\makecell{0\\0\\0}   & \makecell{0\\0\\0} & \makecell{-\\-\\-} & \makecell{0.76\\0.74\\0.74} & \makecell{\SecBestPerf{0.84}\\\SecBestPerf{0.82}\\\SecBestPerf{0.81}} & \makecell{\BestPerf{0.88}\\\BestPerf{0.87}\\\BestPerf{0.86}} \\
\bottomrule
\end{tabular}
    \caption{Evaluation results of biology multi-omics property classification tasks. \BestPerf{Bold}: best, \SecBestPerf{underline}: the second best. Details of expert models and specialist LLMs are provided in Appendix Table.~\ref{tab:appendix_bio_prop_cls_long}. \dag~denotes that only randomly sampled 1000 test samples were utilized for evaluation in the corresponding task to avoid substantial testing costs.}
\label{tab:bio_prop_cls_long}
\end{table}

\paragraph{Biology Multi-omics Property Classification} 
For biology-related (DNA/RNA/protein) classification tasks, we include the following 26 tasks from the Biology-Instructions~\cite{he2024biology} and Mol-Instructions~\cite{fang2023mol}. These tasks span binary and multi-label classification problems covering molecular interactions, regulatory elements, and functional annotation. 

Specifically, we first consider 11 binary classification task: 
\begin{itemize}
    \item \textbf{Solubility Classification}. For a given protein sequence, this task addresses a binary classification task to predict whether a it is soluble or insoluble. 
    \item \textbf{Antibody-Antigenqing}. This is a binary classification task to predict interactions between antibody and antigen sequences. The ultimate goal is to leverage this understanding to predict effective antibodies against new viral variants. 
    \item \textbf{RNA-Protein Interaction Prediction}. This is a binary classification task to predict interactions between non-coding RNAs (ncRNAs) and proteins based on their respective sequences. Most ncRNAs must interact with proteins to perform their biological functions. Therefore, identifying these interactions is crucial for elucidating the mechanisms underlying ncRNA-mediated biological activities . 
    \item \textbf{Epigenetic Marks Predictions}. For a given DNA sequence, this binary classification task predicts the presence of chemical modifications in DNA sequence.  
    \item \textbf{TF-Mouse}. This is a binary classification task, to determine whether specific regions with transcription factors binding in the DNA sequences or not. These transcription binding sites (TBS) are critical for controlling the initiation, enhancement, or repression of transcription. 
    \item \textbf{Enhancer-Promoter Interaction Prediction}. This task addresses a binary classification task focused on identifying interactions between enhancer and promoter regions within a given pair of DNA sequences. Predicting these enhancer-promoter interactions is crucial for understanding the complex regulatory networks that govern gene activity.
    \item \textbf{TF-Human Prediction}. This is a binary classification task, to determine whether specific regions with transcription factors binding in the DNA sequences or not.

    \item \textbf{Promoter-300\&Promoter-Core}. These two tasks are binary classification problems focused on identifying the presence or absence of promoter regions in DNA sequences. The first task, Promoter Detection 300, involves detecting these regions within a 300 base pair (bp) window that includes both the core promoter and surrounding regulatory elements. The second task, Promoter Detection Core, focuses on detecting a shorter, core sequence (typically 50-100 bp) located directly upstream of the transcription start site. 
    \item \textbf{Human PPI}. This task presents a binary classification task to predict protein-protein interactions (PPI) in humans. Based on pairs of human proteins, the model outputs Positive/Negetive sentences indicating whether the two proteins interact . 
    \item \textbf{Yeast PPI}. This task presents a binary classification task to predict protein-protein interactions (PPI) in yeast. Based on pairs of yeast proteins, the model outputs Positive/Negetive sentences indicating whether the two proteins interact. 
    
\end{itemize}
In summary, the above binary classification task is to evaluate whether the given DNA/RNA/protein sequence is positive or negative  for a specific task. Using solubility classification as an example:
\begin{framed}
        \small\textit{Example:}\\
        \texttt{Instructions: <protein>MKALCLLLLPVLGLLVSSKTLCSMEEAINERIQEVAGSLIFRAISSIGLECQSVT\\SRGDLATCPRGFAVTGCTCGSACGSWDVRAETTCHCQCAGMDWTGARCCRVQPLEHHHHHH</protein> This task predicts if a protein is likely to be soluble or insoluble. Can this protein be expected to be soluble?}\\
        \texttt{Response: Solubility is not predicted for this protein.}
    \end{framed}
More example please refer to the Appendix. 

Then, we consider 15 multi-label\&multi-class classification tasks: 

\begin{itemize}
    \item \textbf{Domain/Motif Classification}. This task requires models to pinpoint folded structural domains and conserved motifs within a protein sequence. These annotations provide a concise information of protein architecture that aids functional inference and downstream analyses. 
    \item \textbf{Protein Function classification}. Protein function classification maps an input protein sequence onto standardized annotations across the three GO ontologies: cellular component, biological process, and molecular function. This could enable high‐throughput, comprehensive annotation of novel proteins.
    \item \textbf{Function-EC Classification}. For a give protein sequence, this multi-label classification task predicts enzyme function by annotating protein sequences with their corresponding Enzyme Commission (EC). 
    \item \textbf{Non-coding RNA Function Classification}. This task presents a multi-class classification task for assigning non-coding RNA (ncRNA) sequences to one of 13 functional classes (e.g., 'tRNA', 'miRNA', 'riboswitch'). Accurate classification is essential for understanding the diverse regulatory roles of ncRNAs in biological processes and disease.   
    \item \textbf{Modification Prediction}. This task addresses a multi-label classification task to predict post-transcriptional modifications in RNA sequences, identifying one or more types from a set of 12 common modifications (e.g., ``m6A'', ``m1A'', ``m5C''). The precise identification of these modification sites is essential for understanding RNA's regulatory mechanisms and its roles in diverse biological processes.

    \item \textbf{Fold Type Classification.} This task involves assigning a protein to one of 1195 fold types based on its amino acid sequence. Fold types reflect fundamental tertiary structural patterns, even across proteins with low sequence similarity.  

    \item \textbf{Subcellular Localization Classification.} This task requires predicting the subcellular localization (e.g., nucleus, cytoplasm, membrane) of a protein in eukaryotic cells based on its sequence. Subcellular location is tightly linked to protein function and disease relevance, making this task vital for biological interpretation and drug discovery.

    \item \textbf{Numerical EC Classification.} This task entails predicting the GO terms associated with a protein sequence. GO annotations describe protein functions across three dimensions: molecular function, biological process, and cellular component. Unlike the Function-EC Classification task, this task outputs four numerical levels of EC number system. Moreover, \emph{since the test data comes from two distinct sets, i.e., New-392 and Price-149, we therefore consider these two datasets are two sub-tasks of Numerical EC Classification.} We add extra spaces between the numbers and the periods to make sure the the four numbers are tokenized properly, for example, 
        \begin{framed}
        \small\textit{Example:}\\
        \texttt{Instructions: What is the input protein sequence's EC identifier? <protein>MTAEPATKKIKLELSDPSEPLTQSDVIAFQKEALFRCINRRRVDFEALRKQYELSRRE...</pro\\tein>}\\
        \texttt{Response: 2 . 3 . 2 . 27}
        \end{framed}

    \item \textbf{Gene Ontology (GO) Terms Classification.} This task entails predicting the GO terms associated with a protein sequence. GO annotations describe protein functions across three dimensions: molecular function, biological process, and cellular component. This tasks include three different data for testing, \textbf{(i)} CASPSimilarSeq: the sequences whose are similar to the sequences released by CASP14. \textbf{(ii)} IDFilterSeq: this version uses ID-filtered sequences to reduce redundancy. \textbf{(iii)} UniProtSeq: this version's protein sequences are from the UniProt database. We therefore consider the above three data soucres are three sub-tasks of GO Terms Classification. 

    \item \textbf{Function Keywords Predictions. } This task targets the prediction of UniProtKB keywords associated with a protein, spanning 10 categories: biological process, cellular component, coding sequence diversity, developmental stage, disease, domain, ligand, molecular function, post-translational modification, and technical term. These keywords provide concise yet informative summaries of protein characteristics and functions. Similar to the Terms Classification, we regard the \textbf{(i)} CASPSimilarSeq, \textbf{(ii)} IDFilterSeq, and \textbf{(iii)} UniProtSeq are three sub-tasks of Function Keywords Prediction.

\end{itemize}

\begin{figure}[t]
\centering
\begin{subfigure}{0.33\textwidth}
\includegraphics[width=\linewidth]{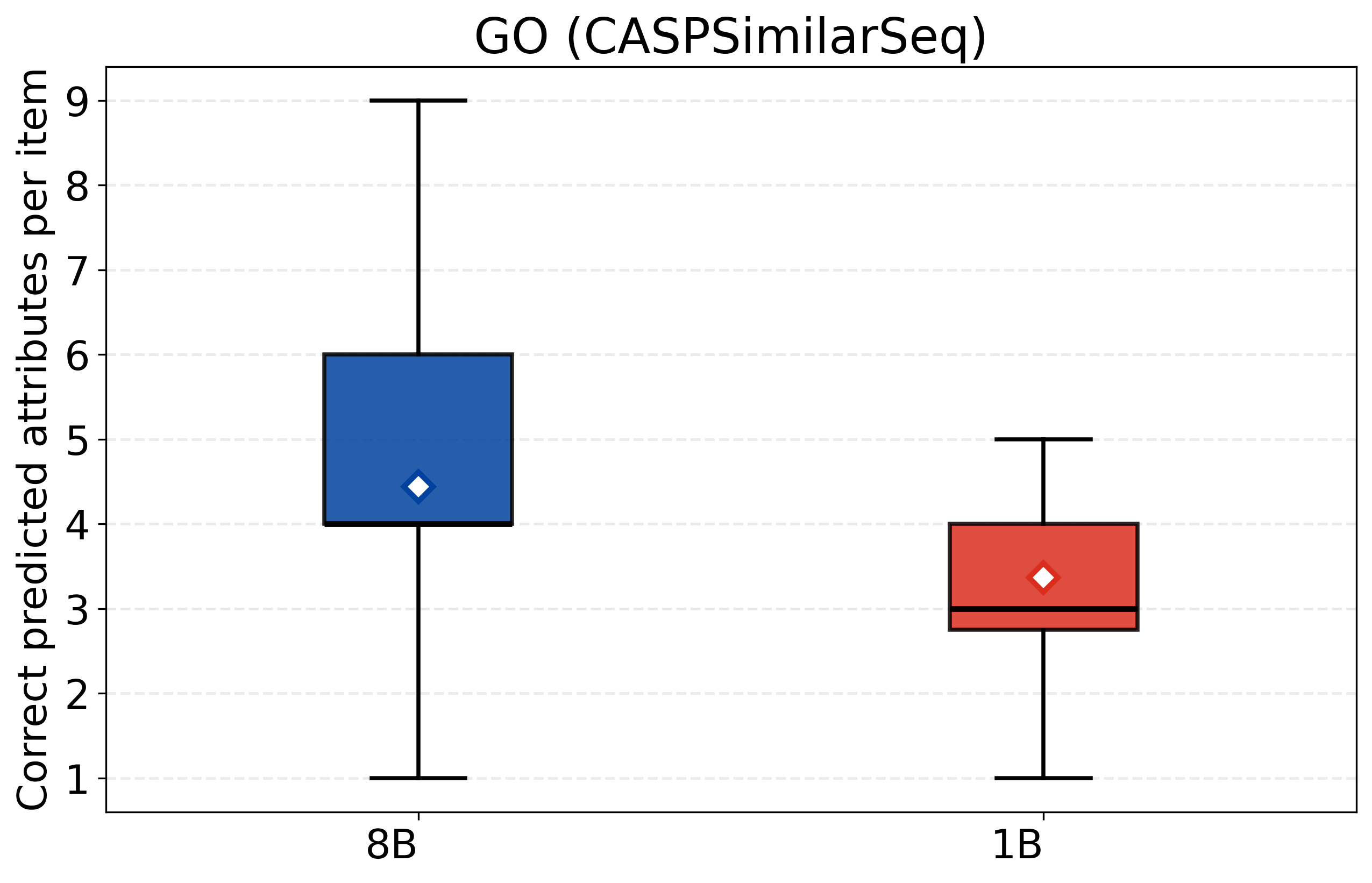}
\subcaption{\small GO (CASPSimilarSeq)}
\end{subfigure}\hfill
\begin{subfigure}{0.33\textwidth}
\includegraphics[width=\linewidth]{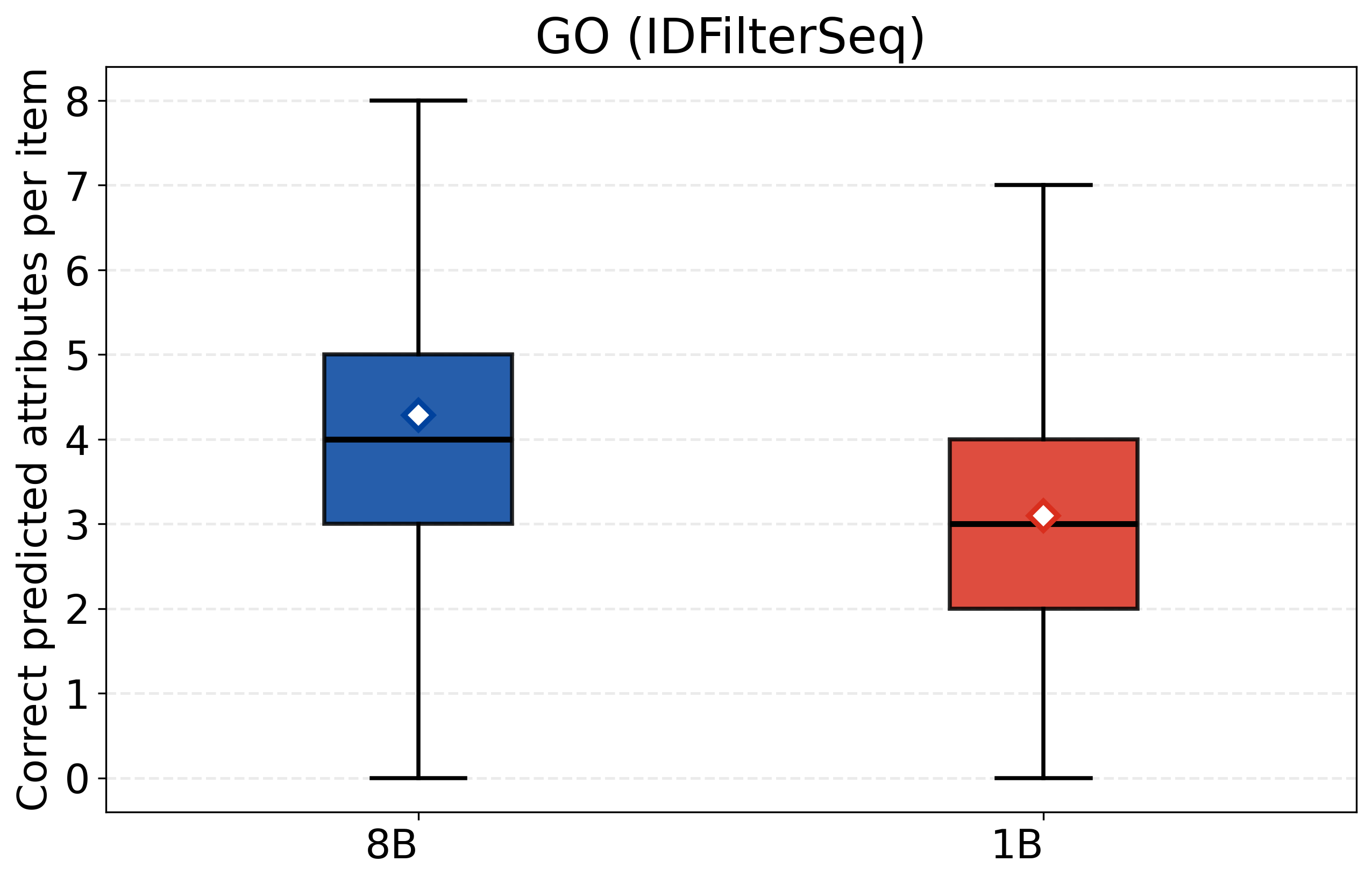}
\subcaption{\small GO (IDFilterSeq)}
\end{subfigure}
\begin{subfigure}{0.33\textwidth}
\includegraphics[width=\linewidth]{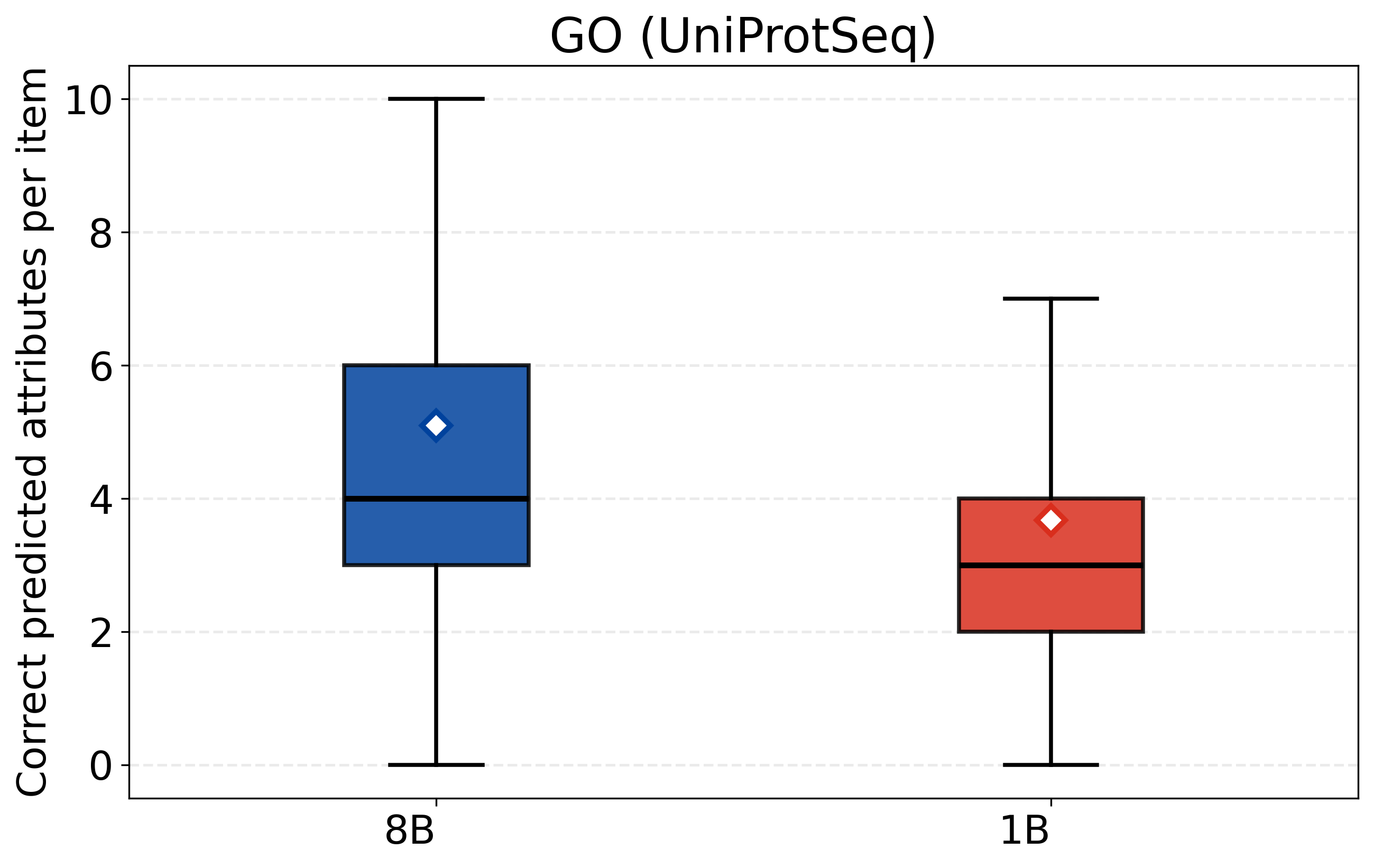}
\subcaption{\small GO (UniProtSeq)}
\end{subfigure}
\caption{Gene Ontology (GO) term classification measured as the number of correctly predicted attributes per sequence (higher is better). Box–whisker plots compare two model sizes (\ProjectName-8B, blue; \ProjectName-1.7B, red) on three evaluation sets: (A) \textit{CASPSimilarSeq}, (B) \textit{IDFilterSeq}, and (C) \textit{UniProtSeq}. Diamonds mark the mean. Across all datasets, the \ProjectName-8B model shows higher medians and upper quartiles than the \ProjectName-1.7B model, with longer upper whiskers (reaching 9–10 correct terms vs.\ 5–8), indicating more high-quality predictions. Gains are most pronounced on \textit{IDFilterSeq} and \textit{UniProtSeq}, suggesting stronger generalization beyond CASP-similar sequences.}
\label{fig:go_classification_figure_1}
\end{figure}

In Fig.~\ref{fig:go_classification_figure_1}, we compare Gene Ontology (GO) term classification performance between the \ProjectName-1.7B and \ProjectName-8B models across three evaluation sets. The box–whisker plots show that the \ProjectName-8B model consistently achieves higher medians and upper quartiles of correctly predicted attributes per sequence, with upper whiskers extending to 9–10 correct terms. These improvements are most evident on IDFilterSeq and UniProtSeq, highlighting stronger generalization beyond CASP-similar sequences. 

We present the case study of the Function Keywords Prediction task in Fig.~\ref{fig:functional_keywords_case}. 

\begin{figure}[htbp]
\centering
\begin{tcolorbox}[title={Function Keywords Prediction Case}]
\textbf{Input}\\
\texttt{Please give the functional keywords of the following protein sequences. <protein>MVEAFCATWKLTNSQNFDEYMKALGVGFATRQVGNVTKPTVIISQEGDKVV\\IRTLSTFKNTEISFQLGEEFDETTADDRNCKSVVSLDGDKLVHIQKWDGKETNFVREIKDGKMVMTLTFGDV\\VAVRHYEKA</protein> }

\medskip
\renewcommand{\arraystretch}{1.08}
\begin{tabularx}{\linewidth}{@{} l Y @{}}
\toprule
\textbf{Reference} &
3D-structure ; Acetylation ; Alternative splicing ; Cytoplasm ; Direct protein sequencing ; Lipid-binding ; Reference proteome ; Transport \\
\midrule
\textbf{Our model} &
\Match{Acetylation}; \Match{Cytoplasm}; \Match{Direct protein sequencing}; \Match{Lipid-binding}; \Match{Reference proteome}; \Match{Transport} \\
\midrule
\textbf{GPT-5} &
Binding; Metal-binding; Oxidoreductase\\
\bottomrule
\end{tabularx}
\end{tcolorbox}
\caption{Case study of functional keyword prediction for a single protein sequence. The curated reference lists eight UniProtKB keywords. Our model correctly recovers six of them (\emph{Acetylation}, \emph{Cytoplasm}, \emph{Direct protein sequencing}, \emph{Lipid-binding}, \emph{Reference proteome}, \emph{Transport}), yielding 100\% precision and 75\% recall. By contrast, GPT-5 returns generic terms (\emph{Binding}, \emph{Metal-binding}, \emph{Oxidoreductase}) with no overlap with the reference.}
\label{fig:functional_keywords_case}
\end{figure}

\paragraph{Summary of Evaluation Results} 
Across all classification tasks, scaling from \ProjectName-1.7B to \ProjectName-8B consistently improves performance, often surpassing GPT baselines (GPT-o3/oss) and approaching specialist systems.  

\begin{itemize}
    \item \textbf{Molecular Classification.} On BBBP, ClinTox, HIV, and SIDER, the 8B model achieves higher accuracy and AUC than 1.7B, reducing toxicity false positives and improving antiviral recall, showing stronger capture of structure–activity relationships.  
    \item \textbf{Material Property Classification.} For MP\textsubscript{classification} and SNUMAT\textsubscript{classification}, 8B improves AUC by about 3\% over 1.7B, narrowing the gap with expert material models and outperforming general-purpose baselines.  
    \item \textbf{Gene-centric Association Classification.} On gSymbol2Tissue, gSymbol2Cancer, and gName2Cancer, the 8B model gains up to 15 F1 over 1.7B, recovering curated gene–disease associations more completely than GPT-5 or smaller models. 
    \item \textbf{Biology Multi-omics Classification.} The largest relative gains appear here: 8B improves MCC/accuracy by 2--10 points across several tasks, and achieves state-of-the-art in multi-label GO and UniProt keyword prediction, with broader and higher-quality annotation coverage.  
\end{itemize}

Overall, \textbf{1.7B $\rightarrow$ 8B} scaling delivers robust gains across molecules, materials, genes, and multi-omics, positioning the 8B model as best or second-best on most tasks and validating unified cross-domain pretraining for discrete property inference.  

\subsection{Generation and Design Tasks}
We leverage the generation and design tasks to assess  models' ability to create novel scientific entities—ranging from molecules, proteins, RNAs to materials—either in an unconditional manner or under explicit structural and functional constraints, thereby providing a rigorous test of generative priors and controllable design capabilities across diverse scientific domains. We show the results of generation and design tasks in Tab.~\ref{tab:generation_design}.

\begin{table}[t]
\centering
\scriptsize

\resizebox{\linewidth}{!}{
\begin{tabular}{l c c c c c c c}
\toprule
Task & Metric & \makecell[c]{Gemini\\2.5-pro} & \makecell[c]{GPT\\oss} & \makecell[c]{Non-LLM \\ Experts}& \makecell[c]{Specialist\\ LLMs} & \makecell[c]{\ProjectName\\1.7B} & \makecell[c]{\ProjectName\\8B} \\
\midrule
\addlinespace[0.8em]
\multicolumn{8}{l}{\raisebox{1.2ex}[0pt][0pt]{\bfseries Unconditional Generation}}\\[-0.8ex]
\multirow{2}{*}{\quad Uncondi. Molecular Generation}
  & Validity $\uparrow$ & 0.568 & 0.952 & \textemdash{} & \SecBestPerf{0.968} & 0.864 & \BestPerf{0.971} \\ 
  & Unique $\uparrow$   & 0.042 & 0.318 & \textemdash{} & 0.966 & \BestPerf{0.988} & \SecBestPerf{0.982} \\  
  \cdashline{1-8}[.6pt/2pt]
  
\multirow{3}{*}{\quad Uncondi. Protein Generation} 
& Avg. Length$\uparrow$ & 128.87 & \BestPerf{318.91} &\textemdash{}  & 284.50 & 280.10 & \SecBestPerf{284.70} \\  
& Diversity$\uparrow$ & 0.62 & \BestPerf{1.00} &\textemdash{}  & \SecBestPerf{0.97}& \BestPerf{1.00} & 0.91 \\  
& Validity$\uparrow$ & \SecBestPerf{0.93} & \BestPerf{0.99} & \textemdash{} & \textemdash{} & 0.80 & \BestPerf{0.99} \\ 
\cdashline{1-8}[.6pt/2pt]

\quad Uncondi. RNA Generation     & Avg. MFE$\downarrow$ & N/A & \BestPerf{-417.58} & \textemdash{} & -177.1 & -178.85 & \SecBestPerf{-200.14} \\  
\cdashline{1-8}[.6pt/2pt]
\quad Uncondi. Material Generation & SMACT$\uparrow$ & \BestPerf{89.70} & N/A & \textemdash{} & 66.07 & 55.06 & \SecBestPerf{88.18} \\ 
\midrule
\addlinespace[0.8em]
\multicolumn{8}{l}{\raisebox{1.2ex}[0pt][0pt]{\bfseries Conditional Material Generation}}\\[-0.8ex]
\multirow{3}{*}{\quad Composition to Material}
  & SMACT  $\uparrow$   & N/A & N/A & \textemdash{} & 83.36 & \BestPerf{86.07} & \SecBestPerf{85.85} \\
  & Precision $\uparrow$& N/A & N/A & \textemdash{} & 98.44 & \SecBestPerf{99.89} & \BestPerf{99.96} \\
  & Novelty $\downarrow$& N/A & N/A & \textemdash{} & \BestPerf{97.13} & \SecBestPerf{85.64} & 84.81 \\
  \cdashline{1-8}[.6pt/2pt]
\multirow{2}{*}{\quad Bulk to Material}
  & SMACT  $\uparrow$   & N/A & N/A & \textemdash{} & \BestPerf{94.75} & \SecBestPerf{88.92} & 87.49 \\
  & Success rate $\uparrow$       & N/A & N/A & \textemdash{} & \textemdash{} & \BestPerf{7.12}  & \SecBestPerf{7.05} \\
\midrule
\addlinespace[0.8em]
\multicolumn{8}{l}{\raisebox{1.2ex}[0pt][0pt]{\bfseries Conditional Small Molecular Generation}}\\[-0.8ex]
\quad Forward Synthesis Prediction        & Exact Match $\uparrow$& 0.32\dag & 0.33 & \textemdash{} & 0.63 & \SecBestPerf{0.69}|0.81 & \BestPerf{0.75}|0.83 \\
\quad Forward Reaction Prediction         & Exact Match $\uparrow$            &0.55 & 0.36 & 0.86 & 0.54 & \SecBestPerf{0.96} & \BestPerf{0.99} \\
\quad Reagent Prediction                  & Exact Match $\uparrow$            & 0.04 & 0.00 & \BestPerf{0.26} & 0.13 & 0.19 & \SecBestPerf{0.22} \\
\quad Retrosynthesis$_\text{Mol-Instructions}$           & Exact Match $\uparrow$            & 0.20 & 0.05 & 0.64 & 0.41 & \SecBestPerf{0.72} & \BestPerf{0.83} \\
\quad Retrosynthesis$_\text{USPTO-50K}$           & Exact Match $\uparrow$            & 0.23\dag & 0.12 & 0.61 & \SecBestPerf{0.70} & {0.68}|0.72 & \BestPerf{0.72}|0.74 \\
\quad Retrosynthesis$_\text{SMol-Instruction}$           & Exact Match $\uparrow$            & 0.12\dag & 0.08 & \textemdash{} & 0.33 & \SecBestPerf{0.41}|0.43\dag & \BestPerf{0.46}|0.47\dag \\
\quad Molecule Generation                 & Exact Match $\uparrow$ & 0.25\dag & 0.03 & \textemdash{} & 0.19 & 0.39|0.34\dag & 0.48|0.46\dag \\
\quad Description-Guided Molecule  & Exact Match $\uparrow$            &0.06 &0.04  & \SecBestPerf{0.11} & 0.002 & 0.10 & \BestPerf{0.12} \\
\midrule
\addlinespace[0.8em]
\multicolumn{8}{l}{\raisebox{1.2ex}[0pt][0pt]{\bfseries Conditional Biochemical Generation}}\\[-0.8ex]
\quad Function-guided Protein Design    &\makecell[c]{Max.\\Normalized SW} $\uparrow$    &0.00\dag &0.00\dag  & \SecBestPerf{0.97} & 0.43 & {0.96}|0.90\dag & \BestPerf{1.00}|1.00\dag \\
\quad Catalytic Activity         & ROUGE-L $\uparrow$  & 0.25\dag & 0.02 & 0.72 & 0.52 & \SecBestPerf{0.77}|0.78\dag & \BestPerf{0.99}|0.95\dag \\
\bottomrule
\end{tabular}
}
\caption{Evaluation results of generation and design tasks. N/A indicates that the model cannot generate samples that meet the required specifications. \BestPerf{Bold} indicates the best performance, and \SecBestPerf{underline} indicates the second best. The names of expert models and specialist LLMs for each task are provided in Appendix Table.~\ref{tab:appendix_generation_design}. \dag~denotes that only randomly sampled 1000 test samples were utilized for evaluation in the corresponding task. This measure is adopted because extremely large-scale test sets exist for certain tasks, which would lead to substantial testing costs.}
\label{tab:generation_design}
\end{table}

\paragraph{Unconditional Generation Tasks.} 
To probe the model's generative prior, we evaluate unconditional generation across four domains: RNA sequences, inorganic material compositions, small molecules, and proteins. 
In this setting, the model receives no task-specific conditioning beyond minimal format requirements (e.g., valid alphabet/grammar, length or syntax ranges) and must sample plausible candidates directly from its learned distribution. 
Furthermore, these tasks could provide seed candidates for downstream, property-conditioned design.

\begin{figure}[t]
\centering
\begin{subfigure}{0.31\textwidth}
\includegraphics[width=\linewidth]{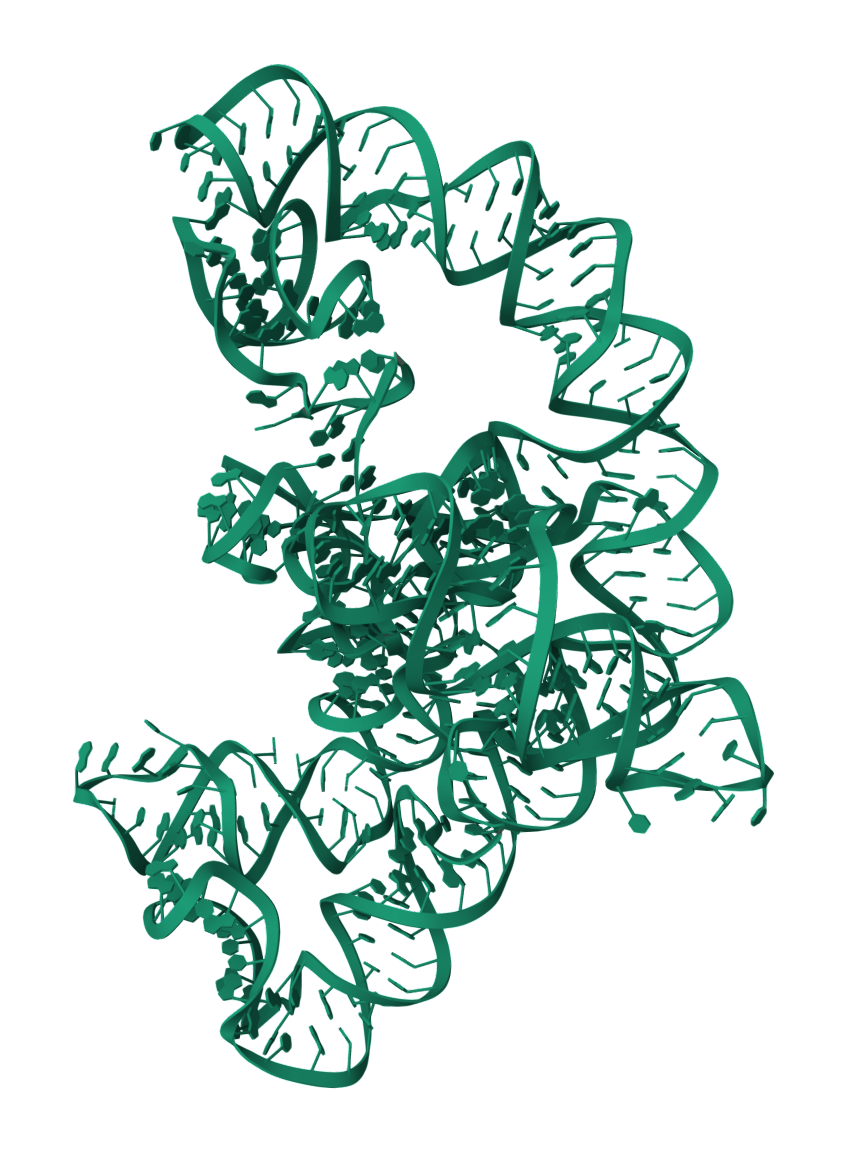}
\end{subfigure}\hfill
\begin{subfigure}{0.31\textwidth}
\includegraphics[width=\linewidth]{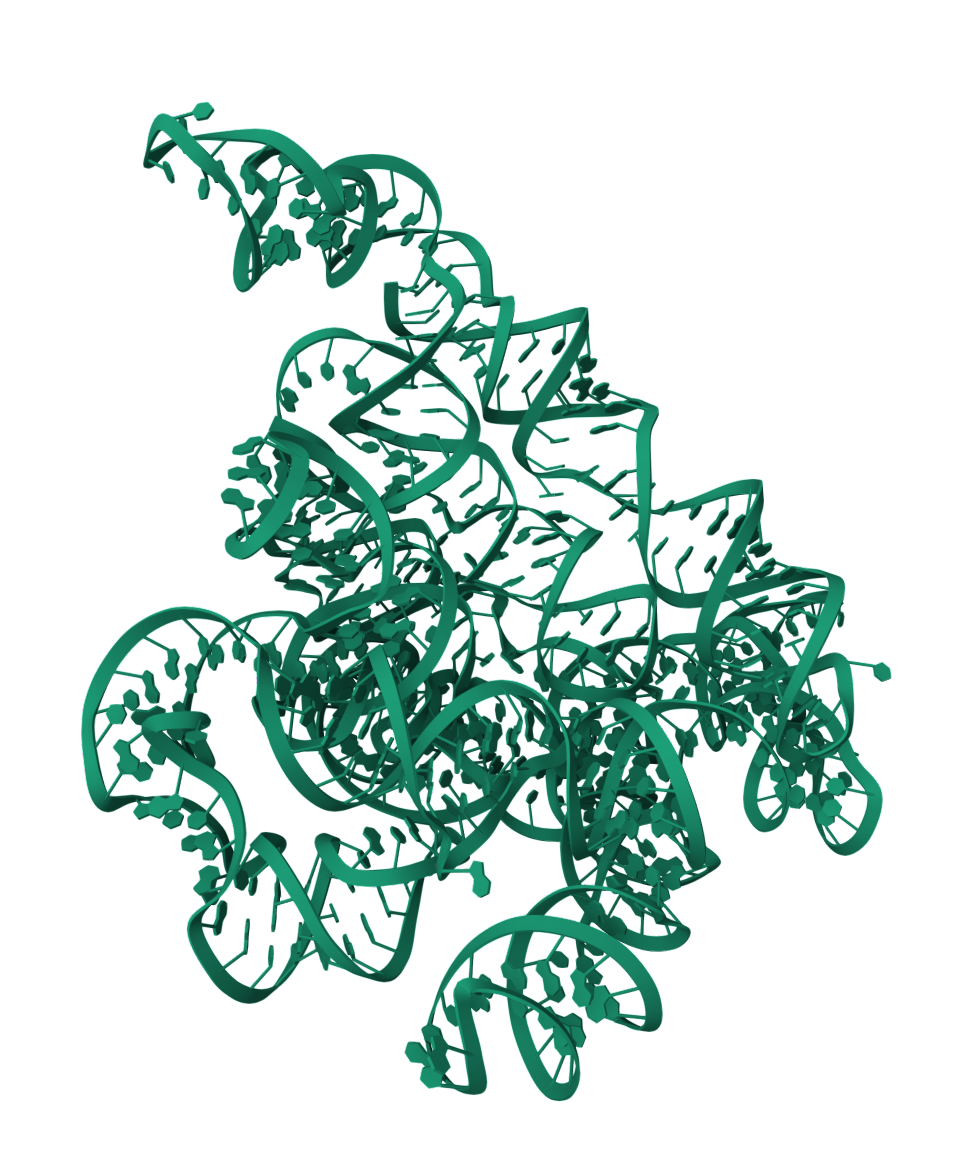}
\end{subfigure}\hfill
\begin{subfigure}{0.31\textwidth}
\includegraphics[width=\linewidth]{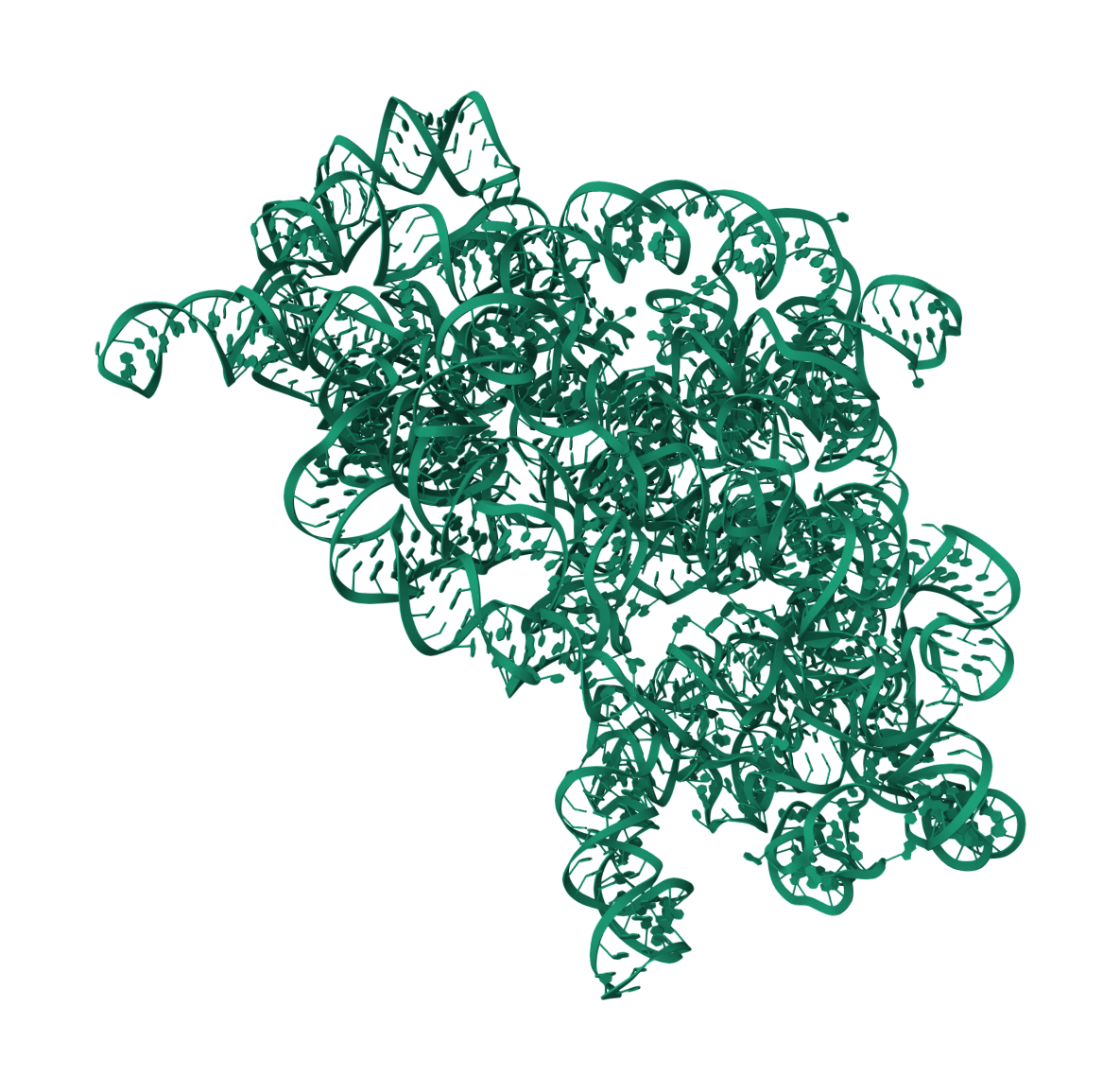}
\end{subfigure}\hfill
\caption{Visualization of Unconditional RNA Generation. 
The pLDDT scores of these three RNA sequences are 74.6, 72.3, and 71.5, respectively. These values indicate that the generated structures achieve moderate confidence and highlight the model's ability to produce diverse yet structurally plausible candidates. 
} 
\label{fig:rna_generation_figure}
\end{figure}

\begin{itemize}
    \item \textbf{Unconditional RNA generation. } The unconditional RNA generation task aims to produce novel RNA sequences without any input conditions, using the four types of nucleotides. We show the visualization of unconditional RNA generation in Fig.~\ref{fig:rna_generation_figure}. 
    \begin{framed}
    \small\textit{Example:}\\
    \texttt{Instructions: Please generate a novel RNA sequence of length 50 nt.}\\
    \texttt{Response: AUGGCUACUGGUAACCGGAAUCCGGAUUCCGGAAUCCGGAAUCCGGAAUCCGGA}
    \end{framed}

    \item \textbf{Unconditional Material Generation. } The unconditional material generation task aims to generate novel material compositions without any input constraints.  
    \begin{framed}
    \small\textit{Example:}\\
    \texttt{Instructions: Produce a material composition with any bulk modulus.}\\
    \texttt{Response: \{composition: "Li2O·TiO2·MnO2", bulk\_modulus: 120.5 GPa\}}
    \end{framed}

    \item \textbf{Unconditional Small Molecular Generation. } This task aims to produce novel molecules that are both chemically plausible and structurally diverse, without any guiding input conditions. 
    \begin{framed}
    \small\textit{Example:}\\
    \texttt{Instructions: Generate a molecule with <SMILES>.}\\
    \texttt{Response:  <SMILES> CC(=O)CCCCCC(C)C </SMILES>}
    \end{framed}

    \item \textbf{Unconditional Protein Generation. } This task is a generative task focused on creating new protein sequences or structures without any specific constraints. The purpose of this task is to test our model's understanding of the fundamental principles of protein biology and to generate completely new proteins with biological feasibility from scratch. 

\end{itemize}

\textbf{Evaluation Metric.} 
\begin{itemize}
    
    \item For unconditional molecular generation, 800 sequences are randomly and independently generated, and the proportions of valid sequences and unique sequences is evaluated by using RDKit. 
    \item For  unconditional protein generation, 500 sequences are generated independently at random, and the average length, diversity (non-repetition rate), and validity (whether only containing valid characters) of these sequences are evaluated.

    \item For unconditional RNA generation, 5000 sequences are generated independently at random, and the average minimum free energy (MFE) of the valid sequences is evaluated. 
    \item For unconditional material generation, 5000 materials are generated independently at random, and the percentage of valid materials is evaluated by Semiconducting Materials by Alternative Chemical Toolkit (SMACT). 
    
    % \item For conditional material generation, one candidate is generated per prompt given natural-language composition constraints, and we evaluate the format validity, SMACT validity, composition precision, and novelty.

\end{itemize}

\paragraph{Conditional Material Generation Tasks}
In conditional material generation tasks, the model is required to generate candidate material compositions that satisfy given constraints, such as specific elemental requirements or target physical properties. These tasks not only assess the model's generative capability but also its ability to adhere to underlying chemical and physical principles in material design.
\begin{itemize}
    \item \textbf{Composition to Material Generation. }This task asks the model to design materials containing specific elements (e.g., Li, Ti, Mn, Fe, and O).  
    \begin{framed}
    \small\textit{Example:}\\
    \texttt{Instructions: Given F, In, Tc, Zr, design a material composition.}\\
    \texttt{Response: \{composition: "FeInTcZrO4", stability: "metastable"\}}
    \end{framed}

    \item \textbf{Bulk to Material Generation.} In this task, the model is asked to generate materials with a desired bulk modulus.  
    \begin{framed}
    \small\textit{Example:}\\
    \texttt{Instructions: Given a bulk modulus of 75.321 GPa, propose a material composition.}\\
    \texttt{Response: \{composition: "Al3Ti", predicted\_bulk\_modulus: 74.8 GPa\}}
    \end{framed}
        
\end{itemize}

\begin{figure}
    \centering
    \includegraphics[width=1.0\linewidth]{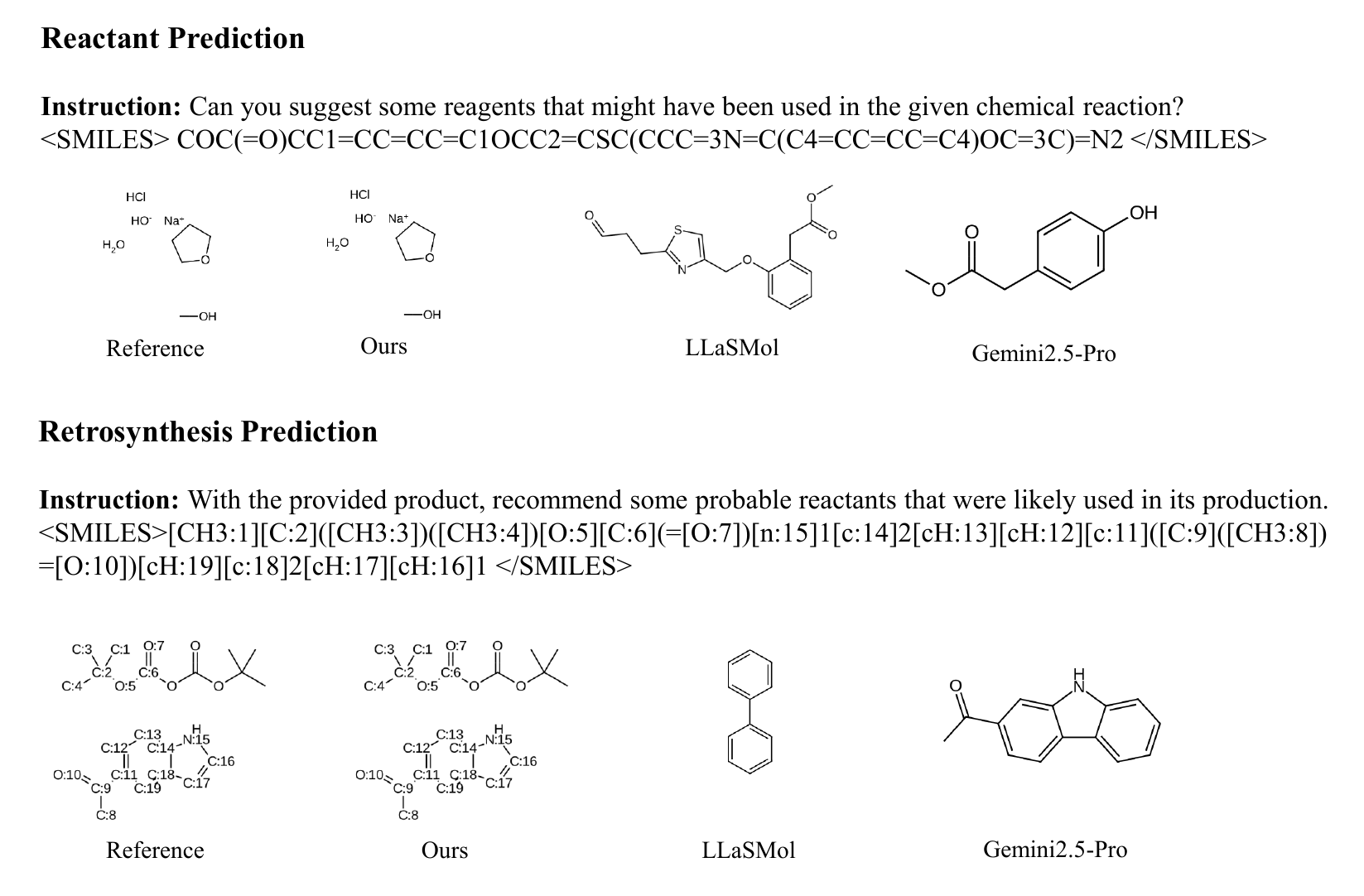}
    \caption{Visualization comparison on molecular generation tasks. Top (Reactant Prediction): given a product SMILES (instruction shown), models propose likely reagents. Bottom (Retrosynthesis Prediction): given a product SMILES, models suggest probable reactants/precursors. Our model recovers reactants that align with the reference chemistry, whereas baseline models propose off-target species.}
    \label{fig:mol_prediction_translation}
\end{figure}

\textbf{Evaluation Metrics.} 
In addition to SMACT, we evaluate model performance using three complementary metrics: Precision, Novelty, and Success Rate. 
\begin{itemize}
    \item \textbf{Precision} quantifies the proportion of correctly predicted elements relative to the total number of ground-truth elements, averaged across all samples. It reflects how accurately the predicted material matches the specified composition.
    \item \textbf{Novelty} measures the proportion of predictions that do not appear in the ground-truth material set, thereby capturing the model’s ability to generate unseen candidates.
    \item \textbf{Success Rate} denotes the fraction of predictions that exactly reproduce all ground-truth elements of a sample. It assesses whether the model can fully recover the intended material.
\end{itemize}
\paragraph{Conditional Small Molecular Generation Tasks} 
We evaluate controllable, chemistry‐aware generation under explicit conditioning signals, including reactants/reagents, textual descriptions, scaffolds, and target property constraints. 
These tasks probe whether a model can follow chemical instructions, satisfy constraints (e.g., validity, synthesizability, and target properties), and balance novelty with correctness, and we follow the task settings in Mol-Instruction~\cite{fang2023mol} and SMolInstruct~\cite{yu2024llasmol}. 

\begin{itemize}
    \item \textbf{Forward Synthesis Prediction.} The Forward Synthesis task predicts the product of a chemical reaction given a set of reactants and reagents. The model takes the SMILES strings of the starting materials as input. Its goal is to generate the most probable product molecule, effectively simulating a chemical reaction. This task is fundamental to drug discovery and materials science, as it helps chemists predict the outcome of a synthesis and explore new reaction pathways. 

    The Molecule Generation task is a generative task that creates a molecule based on a given set of properties and constraints. The model receives a natural language description as input. Its goal is to generate a valid SMILES string for a molecule that satisfies all the specified conditions. This task is highly valuable in drug discovery, as it allows for the design of novel molecules with specific desired characteristics.

    \item \textbf{Description Guided Molecule Design.} This is a text-conditioned generation task that produces candidate molecules that satisfy explicit, user-specified criteria expressed in natural language (e.g., target activity). By conditioning the generative process on these constraints, the chemical search space is narrowed toward compounds consistent with the desired profile, improving the efficiency of design and optimization in applications such as drug discovery.
    
    \item \textbf{Forward Reaction Prediction.} Forward reaction prediction forecasts the expected products of a chemical reaction from specified reactants and reagents. 

    \item \textbf{Reagent Prediction.} Reagent prediction identifies the optimal catalysts, solvents and auxiliary agents for any given reaction. 

    \item \textbf{Retrosynthesis Prediction.} Retrosynthesis prediction task is to invert the synthetic process—beginning with the target molecule and working backward to identify viable precursors and reaction sequences.

\end{itemize}

\textbf{Evaluation Metric.}  The exact match metric enforces the strictest criterion: a prediction is considered correct only if the predicted molecular formula string is identical to one of the ground-truth formulas in both element symbols and their order. Formally,  
\[
    M_{\mathrm{exact}} = \frac{1}{N} \sum_{i=1}^{N} \mathbb{I}\!\left( \exists j \;\; P_i = G_{i,j} \right),
\]
where $N$ denotes the total number of samples, $P_i$ the predicted formula for the $i$-th sample, $\{G_{i,j}\}$ the corresponding set of ground-truth formulas, and $\mathbb{I}(\cdot)$ the indicator function that equals 1 if the condition holds and 0 otherwise.

We visualize the outputs in Fig.~\ref{fig:mol_prediction_translation}. Specifically, compared with LLaSMol and Gemini-2.5-Pro, our method recovers reagents/reactants that more closely match the reference chemistry in both reactant- and retrosynthesis-prediction settings.

\paragraph{Conditional Biochemical Generation Tasks} 
Follow Mol-Instructions~\cite{fang2023mol}, we assess the model's capacity to generate biochemical outputs conditioned on specific inputs, ranging from function-guided protein design to enzyme reaction description from sequences. 
\begin{itemize}
    \item \textbf{Function-guided Protein Design.} Given a functional prompt (e.g., ``DNA binding''), the model outputs a corresponding amino-acid sequence wrapped in <protein> tags. This evaluates its ability to map explicit functional requirements to viable sequences. Function-driven protein design speeds up directed evolution, enzyme engineering and drug-target discovery. 

 \item \textbf{Catalytic Activity Prediction. } Catalytic activity prediction aims to infer the exact chemical transformation an enzyme performs directly from its amino acid sequence. 
    \begin{framed}
        \small\textit{Example:}\\
        \texttt{Instructions: Using the protein sequence supplied, identify and describe the enzymatic catalytic activity, with emphasis on the chemical reaction it accelerates:  <protein>...</protein>}\\
        \texttt{Response: Based on the provided protein sequence, the enzyme appears to facilitate the chemical reaction: (-)-secoisolariciresinol + NADP(+) = (+)-lariciresinol + H(+)   + NADPH.}
    \end{framed}
\end{itemize}

\textbf{Evaluation Metric.} For Function-guided protein design, we use Normalized Smith-Waterman score (Normalized SW) as the metric. 

\paragraph{Summary of Evaluation Results}
Scaling from \textbf{\ProjectName-1.7B} to \textbf{\ProjectName-8B} strengthens both \emph{unconditional} and \emph{conditional} generation across molecules, proteins, RNAs, and materials.

\begin{figure}[t]
\centering
\begin{subfigure}{0.31\textwidth}
\includegraphics[width=\linewidth]{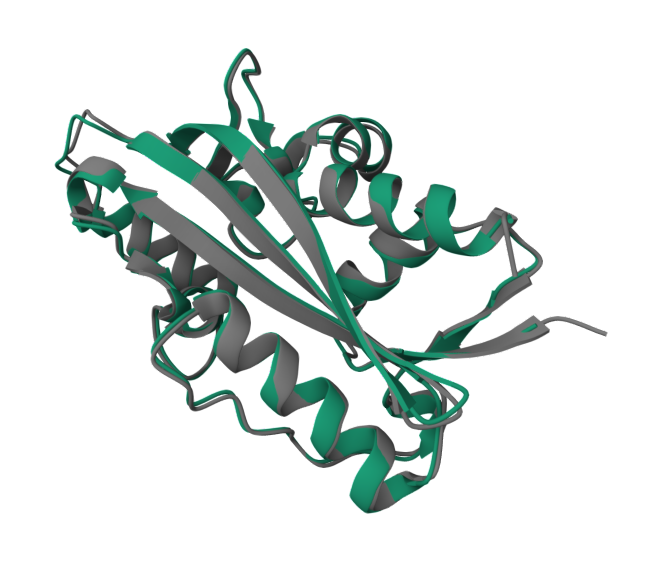}
\end{subfigure}\hfill
\begin{subfigure}{0.31\textwidth}
\includegraphics[width=\linewidth]{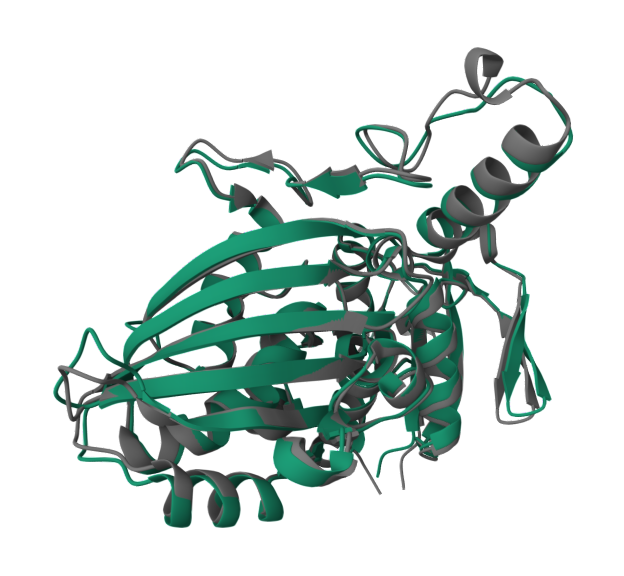}
\end{subfigure}\hfill
\begin{subfigure}{0.31\textwidth}
\includegraphics[width=\linewidth]{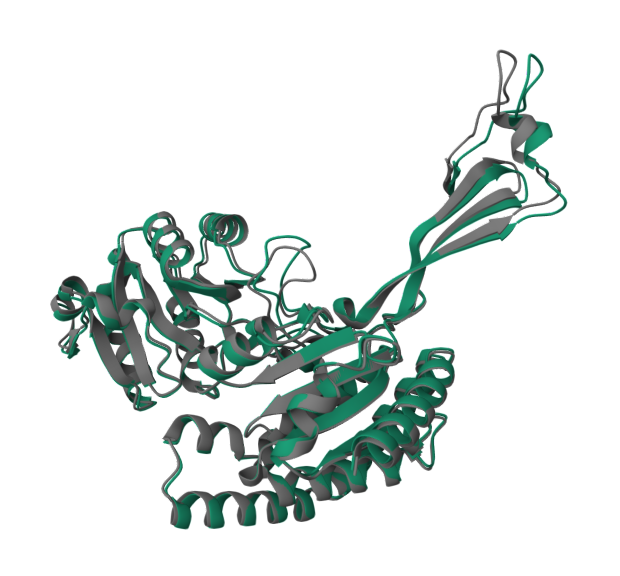}
\end{subfigure}\hfill
\caption{Structural overlay of the proteins: the reference structure is shown in gray and \ProjectName-8B's prediction in green. 
% The prediction recapitulates the overall domain organization and secondary-structure topology, with larger deviations mainly in loop/linker regions.
}
\label{fig:protein_translation_figure}
\end{figure}

\begin{itemize}
  \item \textbf{Unconditional generation.} \ProjectName-8B attains SOTA-level molecular validity while retaining very high uniqueness; it also improves material feasibility (SMACT) beyond prior Specialist LLMs. RNA samples exhibit lower (better) free energies than \ProjectName-1.7B and the LLM baseline. 
  \item \textbf{Conditional material design.} With element-set constraints, \ProjectName-8B maintains high chemical validity with near-perfect precision and improved (lower) novelty; under bulk-modulus conditioning, validity is comparable to \ProjectName-1.7B with a small trade-off in success rate.
  \item \textbf{Conditional molecular generation.} \ProjectName-8B consistently surpasses both \ProjectName-1.7B and prior Specialist LLMs across forward synthesis/reaction prediction, reagent selection, retrosynthesis, and text-guided molecule design, with the largest gains in retrosynthesis and top-1 exact matching.
  \item \textbf{Conditional Protein Design.} Function-guided protein generation reaches near-ceiling sequence similarity and substantially outperforms prior LLM baselines. As shown in \ref{fig:protein_translation_figure}, \ProjectName-8B can generate proteins that are very close to the reference protein. More results please refer to the Appendix. 
\end{itemize}

\noindent Overall, scaling primarily amplifies constraint satisfaction and chemical/structural validity under conditioning, while also strengthening the generative prior in unconditional regimes.

% \subsection{Case Study}

\subsection{Reasoning Examples}

\begin{figure}[t]
\centering
\begin{tcolorbox}[colback=white,colframe=black,title={SNUMAT Classification}]

\textbf{Input.} You are a material scientist. Look at the chemical formula of the given crystalline material and predict its property. The output must be in a json format. For example: {property\_name:predicted\_property\_value}. Answer as precise as possible and in as few words as possible. \\
chemical formula: RbCuCl3\\
property name: Direct\_or\_indirect\_HSE

\textbf{Thinking Content:}\\[-0.25em]
\small
<think>\\
The compound RbCuCl3 is analyzed based on its chemical formula. It consists of \hl{Rb$^{+}$, Cu, and Cl$^{-}$ ions}, likely forming \hl{ionic or coordination bonds}. The Cu likely has a \hl{+2 oxidation state}, giving it a \hl{d$^{9}$ configuration}. Considering its probable crystal structure and electronic arrangement, RbCuCl3 is predicted to be an \hl{indirect bandgap semiconductor} due to the ionic bonding and the potential for indirect electron transitions in its structure. \\
\{  "Direct\_or\_indirect\_HSE": "indirect"\}
\\
</think>
\normalsize

\textbf{Answer:}\\[-0.25em]
{\small \{Direct\_or\_indirect\_HSE : Indirect\}}

\textbf{Reference:}\\[-0.25em]
{\small Indirect}

\end{tcolorbox}
\caption{ \textbf{SNUMAT Classification Example.} This figure demonstrates the model's reasoning process for predicting the bandgap type of the material \textbf{RbCuCl3}. The thinking content highlights a logical chain of deductions: from analyzing the constituent ions and chemical bonding, to identifying the crucial \textbf{+2 oxidation state} and resulting \textbf{d$^{9}$ electronic configuration} of the Cu ion.}
\label{cot_t1}
\end{figure}

This section provides a detailed exposition of the model's reasoning capabilities across diverse scientific domains. Each example illustrates the sophisticated chain-of-thought processing employed by the model to arrive at precise predictions and assessments, underscoring its efficacy in tackling complex scientific inquiries. These cases, ranging from materials science to molecular biology, demonstrate the model's ability to interpret intricate inputs and generate well-supported conclusions. For each instance, the input prompt, the model's detailed thinking content, the final answer, and a reference are provided to offer a transparent view of the complete reasoning trajectory.

\subsubsection{Material Task: SNUMAT Classification}
As shown in Fig.~\ref{cot_t1}, the SNUMAT Classification example showcases a sophisticated ability to infer complex electronic properties from basic chemical information. 
Given the chemical formula RbCuCl3 and a request for its Direct\_or\_indirect\_HSE, the model initiates its reasoning by correctly deconstructing the formula into its ionic components. Critically, it accurately deduces the +2 oxidation state of copper, which leads to the identification of its d$^{9}$ electronic configuration. By integrating these pieces of information—ionic nature, specific oxidation states, and key electronic configurations—the model constructs a logical pathway to its final, correct prediction. This example demonstrates the model's capacity to not just identify individual chemical features, but to connect them in a physically meaningful sequence to determine a material's fundamental electronic character.

\begin{figure}[t]
\centering
\begin{tcolorbox}[colback=white,colframe=black,title={Promoter-300}]

\textbf{Input.} 
<dna>TTCTTTTTCCGGATTTTCAGCCGGGTCTTCCGGGGATGGAGAGCAAAAGGACTTGG\\
TGCTCTCGGAGAGAGCCTGCAGGGGGCGGGGTGTTGGTAGAAGGAGTGGGAGGGGTTTTACTGTAG\\
CAAGGCGGGGCCTCAGTTGAGTGGGTGGGGCCTGGAGGCAACTAACGCGGCATAGGGCGAGGGCG\\
GGGCCGGCAGGCCGGGGCGGGGCTGGGAAGTAGTCGGGCGGGGTTGTGAGACGCCGCGCTCAGCT\\
TCCATCGCTGGGCGGTCAACAAGTGCGGGCCTGGCTCAGCGCGGGGGG</dna>: Could you assess the potential of this 300 bp DNA as a promoter region?

\textbf{Thinking Content:}\\[-0.25em]
\small
<think>\\
The 300 bp DNA sequence exhibits several features that suggest it might function as a promoter region, \hl{though definitive confirmation requires further analysis}. Here's a structured summary of the evaluation:

1. \textbf{Length and Position}: \hl{The sequence length of 300 bp is within the typical range for promoter regions}, located upstream of the transcription start site.

2. \textbf{GC Content}: The sequence is GC-rich, \hl{which is characteristic of GC-rich promoters}, potentially indicating a strong promoter element.

3. \textbf{CpG Islands}: The presence of multiple CG dinucleotides suggests a CpG island, \hl{commonly found in promoter regions}, supporting the possibility of being a promoter.

4. \textbf{Transcription Factor Binding Sites}: While specific motifs like the TATA box are not evident, the sequence may contain other transcription factor binding sites that could drive gene expression.

5. \textbf{Conservation Analysis}: Without checking against known promoters, it's difficult to assess conservation, but this could provide additional evidence.

6. \textbf{Recommendation for Further Analysis}: \hl{To confirm its potential as a promoter, it is recommended to use bioinformatics tools like PromoScanner or compare it with known promoters in databases}.

In conclusion, while the sequence shows features consistent with a promoter region, such as appropriate length, GC content, and CpG islands, \hl{further analysis with specialized tools is necessary for confirmation}.\\
</think>

\normalsize

\textbf{Answer:}\\[-0.25em]
{\small 
Promoter motifs are detectable in the given DNA fragment.}

\textbf{Reference:}\\[-0.25em]
{\small The 300 bp DNA sequence includes a recognizable promoter region.}

\end{tcolorbox}
\caption{\textbf{DNA Promoter Region Assessment Example.} This figure demonstrates the model's structured approach to evaluate a 300 bp DNA sequence for its potential as a promoter region. The thinking content meticulously breaks down the analysis into key biological features and provides a recommendation for further experimental validation.}
\label{cot_t2}
\end{figure}

\subsubsection{DNA Task: Promoter-300}
As shown in Fig.~\ref{cot_t2}, the model is tasked with assessing the promoter potential of a 300 bp DNA sequence. The reasoning sequence meticulously evaluates several critical characteristics pertinent to promoter regions. It correctly identifies the sequence length as being within the typical range for promoters and notes the GC-rich content and the presence of CpG islands—both established indicators of promoter activity. While acknowledging the absence of an explicit TATA box, the model prudently suggests the potential for other transcription factor binding sites. Importantly, it recommends further bioinformatics analysis using specialized tools or databases, demonstrating a nuanced understanding of the empirical requirements for definitive promoter identification. This structured, multi-faceted analysis highlights the model's ability to synthesize biological cues into a well-reasoned assessment, concluding with a biologically plausible interpretation.

\begin{figure}[t]
\centering
\begin{tcolorbox}[colback=white,colframe=black,title={Solubility}]

\textbf{Input.}
<protein>MDAQTIAPGFESVAELFGRFLSEDREYSAQLAAYHRGVKVLDISGGPHRRPDSVTG\\
VFSCSKGVSGLVIALLVQDGFLDLDAEVVKYWPEFGAEGKATITVAQLLSHQAGLLGVEGGLTLAEYNNS\\
ELAAAKLAQMRPLWKPGTAFGYHALTIGVFMEELCRRITGSTLQEIYEQRIRSVTGAHFFLGLPESEEPRY\\
ATLRWAADPSQPWIDPASHFGLSANSAVGDILDLPNLREVRAAGLSSAAGVASAEGMARVYAAALTGLA\\
ANGDRAAVAPLLSEETIQTVTAEQVFGIDRVFGETSCFGTVFMKSHARSPYGSYRAFGHDGASASLGFA\\
DPVYELAFGYVPQQAEPGGAGCRNLELSAAVRKAVTELAQ</protein> [Solubility Prediction Binary Classification] Will the protein generated from this sequence be soluble?

\textbf{Thinking Content:}\\[-0.25em]
\small
<think>\\
Based on the analysis of the provided protein sequence, the protein is likely soluble. Here's a concise summary of the reasoning:

\textbf{1. Amino Acid Composition:} The sequence contains a significant number of \hl{hydrophilic and charged amino acids} such as \hl{Arginine (R), Lysine (K), Glutamic acid (E), and Aspartic acid (D)}. These residues \hl{promote solubility by interacting with water}.

\textbf{2. Absence of Long Hydrophobic Regions:} While there are some hydrophobic residues, they \hl{do not form extensive stretches that could lead to aggregation or membrane insertion}, which might reduce solubility.

\textbf{3. No Obvious Transmembrane Domains:} The sequence \hl{lacks regions indicative of transmembrane domains}, which are typically \hl{hydrophobic and contribute to insolubility}.

\textbf{4. Presence of Charged Residues:} The abundance of \hl{charged residues} suggests potential for \hl{ionic interactions with water}, enhancing solubility.

\textbf{5. Aromatic Residues:} While aromatic residues can sometimes reduce solubility, their \hl{distribution in this sequence does not appear to be a significant issue}. Given these factors, the protein is predicted to be soluble. 

\textbf{Conclusion:} The protein is \hl{likely soluble based on its composition and sequence analysis}.\\
</think>

\normalsize

\textbf{Answer:}\\[-0.25em]
{\small
The solubility task forecasts a protein's ability to dissolve in solution. This protein is highly soluble.}

\textbf{Reference:}\\[-0.25em]
{\small This task evaluates if a protein will dissolve in a solution. Yes, this protein is soluble.}

\end{tcolorbox}
\caption{\textbf{Protein Solubility Prediction Example.} This figure illustrates the model's reasoning for predicting protein solubility. The thinking content details an analysis of key factors, such as the high proportion of hydrophilic and charged amino acids and the absence of long hydrophobic regions or transmembrane domains, which collectively lead to the prediction of a \textbf{soluble} protein.}
\label{cot_t3}
\end{figure}

\subsubsection{Protein Task: Solubility}
As shown in Fig.~\ref{cot_t3}, the Solubility prediction task exemplifies the model's analytical prowess in protein biochemistry. Provided with a protein sequence, the model systematically dissects its characteristics to predict solubility. The reasoning primarily focuses on the high proportion of hydrophilic and charged amino acids (e.g., Arginine, Lysine, Glutamic acid, Aspartic acid), identifying them as key factors that promote solubility through favorable interactions with water. It further reinforces this conclusion by noting the absence of long hydrophobic regions or transmembrane domains, which are common drivers of aggregation and insolubility. This detailed consideration of amino acid composition and its implications for protein-solvent interactions underscores a sophisticated understanding of the biophysical principles governing protein solubility.

% \subsection{Discussion}
\subsection{Effectiveness of Pretraining}

\begin{table}[t]
\centering
\begin{tabular}{@{} l l r r r r @{}}
\toprule
\textbf{Task} & \textbf{Metric} & \textbf{Before} & \textbf{After} & \textbf{$\Delta$ (pp / RMSE)} & \textbf{Relative $\Delta$} \\
\midrule
Promoter Detection & MCC & 0.451 & 0.719 & 26.8 & $\uparrow$59.4\% \\
Core Promoter Detection & MCC & 0.515 & 0.589 & 7.4 & $\uparrow$14.4\% \\
TF-Human & MCC & 0.447 & 0.529 & 8.2 & $\uparrow$18.3\% \\
ESOL & RMSE $\downarrow$ & 94.54 & 1.21 & -93.33 & $\downarrow$98.7\%  \\
Protein Function Generation & ROUGE-L & 0.755 & 0.823 & 6.7 & $\uparrow$8.9\% \\
Antibody–Antigen Interaction & MCC & 0.129 & 0.416 & 28.7 & $\uparrow$221.9\% \\
RNA–Protein Interaction & MCC & 0.507 & 0.865 & 35.8 & $\uparrow$70.6\% \\
APA Isoform Usage & R2 & 0.311 & 0.879 & 56.8 & $\uparrow$182.3\% \\
GO (CASP-SimilarSeq) & Precision & 0.690 & 0.840 & 15.0 & $\uparrow$21.7\% \\
GeneName $\rightarrow$ Cancer & Precision & 0.610 & 0.740 & 13.0 & $\uparrow$21.3\% \\
\bottomrule
\end{tabular}
\caption{Ablation of pretraining on 10 tasks. Larger is better for MCC/ROUGE-L/Precision/R; lower is better for RMSE. Values are on the [0,1][0,1] scale after rescaling percent-formatted entries; ``pp'' denotes percentage points.}
\label{tab:pretrain_ablation}

\par\smallskip\emph{Notes.} (i) Percent-like entries in the raw results (e.g., \texttt{12.92}) were rescaled to proportion (e.g., 0.12920.1292). (ii) ``pp'' =100×(after-before)=100$\times$(\text{after}-\text{before}) for non-RMSE metrics. (iii) Relative $\Delta$ is computed with respect to the ``before'' score; for RMSE we report relative \emph{reduction}.
\end{table}

We ablate the contribution of pretraining on ten representative tasks that span gene-regulatory sequence understanding (promoter detection at 300\,bp and core promoters at 50--100\,bp), transcription factor binding (binary classification), small-molecule property prediction (ESOL; regression), natural-language protein function generation, two sequence–sequence interaction tasks (antibody–antigen and RNA–protein; binary), alternative APA isoform usage (regression), GO prediction, and gene-name to cancer mapping. Metrics are Matthews correlation coefficient (MCC), ROUGE-L, precision, and R2 (``higher is better''), and RMSE for ESOL (``lower is better'').

Pretraining yields consistent gains on \emph{all} tasks. Across the nine ``higher is better'' tasks, the \textbf{median absolute improvement} is \textbf{+15.0 percentage points (pp)} and the \textbf{mean improvement} is \textbf{+22.0 pp}. The ESOL regression error drops by \textbf{98.7\%} (RMSE 94.54$\rightarrow$1.2194.54$\rightarrow$1.21). The largest absolute gains appear on sequence–sequence \emph{interaction} and \emph{regulatory} tasks (APA, RNA–protein, antibody–antigen), indicating that pretraining most strongly enhances representations that require long-range dependencies, cross-sequence compatibility, and motif composition. Promoter detection at 300\,bp benefits more than core-promoter detection at 50–100\,bp, consistent with the hypothesis that broader context amplifies the value of pretraining. Language-centric tasks (function generation; structured GO terms; gene-name$\rightarrow$cancer) also improve steadily, showing transfer from pretrained biochemical/biomedical knowledge to text- and ontology-grounded outputs.

Tab.~\ref{tab:pretrain_ablation} summarizes per-task results. For proportion-like metrics reported in percent in the raw logs, we rescale to [0,1][0,1]. Absolute deltas are expressed in pp for those metrics and in raw units for RMSE; relative deltas follow the usual convention of (after-before)/before(\text{after}-\text{before})/\text{before} for higher-is-better metrics and (before-after)/before(\text{before}-\text{after})/\text{before} for RMSE.
\vspace{0.5em}

\section{Conclusion}
We introduced a scientific language foundation model that bridges general-purpose large language modeling with the heterogeneous data and reasoning workflows of the natural sciences. Our framework couples multi-representation pre-training with instruction-driven alignment and a reasoning stage that supports both fast responses and step-by-step deliberation. By mapping natural language, DNA/RNA/protein sequences, molecular strings, and materials representations into a shared backbone via task-aware tokenization and consistent input–output schemas, the model moves beyond narrow, discipline-specific solutions and limited task menus. This design enables a single system to support four core capability families central to scientific work—property prediction/classification, sequence generation and design under constraints, scientific translation across formats, and text/knowledge extraction—while grounding its outputs through retrieval and tool execution for improved fidelity and verifiability.

\begingroup
\sloppy
\printbibliography[heading=bibintoc]
\endgroup

\clearpage
\section*{Appendix}
\appendix

\begin{landscape} 
\begin{table}[t]
\centering
\scriptsize
\resizebox{\linewidth}{!}{

\begin{tabular}{l c c c c c c c c c} 
\toprule

Task & Metric & \makecell[c]{Gemini\\2.5-pro} & \makecell[c]{GPT\\o3} & \makecell[c]{DeepSeek\\r1} & \makecell[c]{GPT\\oss} & \makecell[c]{Non-LLM \\ Expert Models} & \makecell[c]{Specialist\\LLMs} & \ProjectName-1.7B & \ProjectName-8B \\
\midrule
\addlinespace[0.8em]
\multicolumn{9}{l}{\raisebox{1.2ex}[0pt][0pt]{\bfseries Molecular Representation Translation}}\\[-0.8ex]
\quad SMILES to IUPAC  & \makecell[c]{Top1 \\Split Match} {\normalsize $\uparrow$} & 4.70\%\dag & 1.50\%\dag & 2.20\% & 0.17\% & \textemdash{} & 29.00\%(LlaSMol-7B~\cite{yu2024llasmol}) & \SecBestPerf{49.52\%}|50.20\%\dag & \BestPerf{56.63\%}|57.10\%\dag\\
\quad SMILES to Formula & \makecell[c]{Top1 \\Element Match} {\normalsize $\uparrow$} & 55.20\%\dag & 64.81\%\dag & 34.00\% & 19.88\% & \textemdash{} & 93.20\%(LlaSMol-7B~\cite{yu2024llasmol}) & \SecBestPerf{95.92\%}|87.1\%\dag & \BestPerf{96.39\%}|96.00\%\dag \\
\quad IUPAC to SMILES & \makecell[c]{Top1 \\Split Match} {\normalsize $\uparrow$} & 38.20\%\dag & 21.82\%\dag & 26.9\% & 7.58\% & \textemdash{} & 70.10\%(LlaSMol-7B~\cite{yu2024llasmol}) & \SecBestPerf{78.85\%}|86.50\%\dag & \BestPerf{84.40\%}|90.70\%\dag \\
\quad IUPAC to Formula & \makecell[c]{Top1 \\Element Match} {\normalsize $\uparrow$} & 88.50\%\dag & 76.50\%\dag & 68.80\% & 33.24\% & \textemdash{} & 87.90\%(LlaSMol-7B~\cite{yu2024llasmol}) & \SecBestPerf{91.81\%}|88.10\%\dag & \BestPerf{92.65\%}|93.00\%\dag \\
\midrule
\addlinespace[0.8em]
\multicolumn{9}{l}{\raisebox{1.2ex}[0pt][0pt]{\bfseries Natural Language Translation of Molecule}}\\[-0.8ex]
\quad Molecular Description            & ROUGE-L $\uparrow$ & 0.23 & 0.07 & 0.06 & 0.05 & \SecBestPerf{0.75}(BioT5+~\cite{pei2024biot5+}) & 0.29(Mol-Instructions-7B~\cite{fang2023mol}) & \SecBestPerf{0.75} & \BestPerf{0.78} \\
\quad Molecular Captioning             & MENTOR $\uparrow$  & 0.42\dag & 0.28\dag &0.25 & 0.15 & \textemdash{} & 0.45(LlaSMol-7B~\cite{yu2024llasmol}) & \SecBestPerf{0.59}|0.59\dag & \BestPerf{0.60}|0.58\dag \\
\midrule
\addlinespace[0.8em]
\multicolumn{9}{l}{\raisebox{1.2ex}[0pt][0pt]{\bfseries Protein Functional Translation}}\\[-0.8ex]
\quad CASPSimilarSeq                   & ROUGE-L$\uparrow$  & 0.01 & 0.01 &0.02 & 0.02 & \textemdash{} & 0.74 (OPI-Galactica-6.7B~\cite{xiao2024opi}) & \SecBestPerf{0.83} & \BestPerf{0.85}  \\
\quad IDFilterSeq                      & ROUGE-L$\uparrow$  & 0.01 & 0.01 &0.01 & 0.03 & \textemdash{} & 0.70 (OPI-Galactica-6.7B~\cite{xiao2024opi}) & \SecBestPerf{0.77} & \BestPerf{0.82} \\
\quad UniProtSeq                       & ROUGE-L$\uparrow$  &  0.01 & 0.02 &0.01 & 0.03 & \textemdash{} & 0.71 (OPI-Galactica-6.7B~\cite{xiao2024opi}) & \SecBestPerf{0.84} & \BestPerf{0.88}  \\
\quad Mol-Instructions                 & ROUGE-L$\uparrow$  & 0.21\dag & 0.08\dag &0.05 & 0.07 & 0.68(BioT5+~\cite{pei2024biot5+}) & 0.44(Mol-Instructions-7B~\cite{fang2023mol}) & \SecBestPerf{0.83}|0.82\dag & \BestPerf{0.99}|0.98\dag  \\
\bottomrule
\end{tabular}
}
\caption{Evaluation results of scientific translation tasks. 
\BestPerf{Bold} indicates the best performance, and \SecBestPerf{underline} indicates the second best.
\dag~denotes that only randomly sampled 1000 test samples were utilized for evaluation in the corresponding task. This measure is adopted because extremely large-scale test sets exist for certain tasks, which would lead to substantial testing costs.
}
\label{tab:appendix_scientific_translation}
\end{table}
\end{landscape}

\begin{landscape}
\begin{table}[t]
\centering
\scriptsize
\begin{tabular}{l c c c c c c c c c} 
\toprule
Task & Metric & \makecell[c]{Gemini\\2.5-pro} & \makecell[c]{GPT\\o3} & \makecell[c]{DeepSeek\\r1} & \makecell[c]{GPT\\oss} & \makecell[c]{Non-LLM \\ Expert Models}& Specialist LLMs & \ProjectName-1.7B & \ProjectName-8B \\
\midrule
\addlinespace[0.8em]
\multicolumn{9}{l}{\raisebox{1.2ex}[0pt][0pt]{\bfseries Scientific Knowledge Extraction and QA}}\\[-0.8ex]
\quad Chemical Entity Recognition & F1 $\uparrow$ & 0.71 & 0.66 &0.05 & 0.55 & \textemdash{} & 0.75(Mol-Instructions-7B~\cite{fang2023mol}) & \SecBestPerf{0.81} & \BestPerf{0.92} \\
\quad Chemical–Protein Interaction Extraction & F1 $\uparrow$ & 0.12 & 0.06 &0.05 & 0.05 & \textemdash{} & 0.22(Mol-Instructions-7B~\cite{fang2023mol}) & \SecBestPerf{0.46} & \BestPerf{0.83} \\
\quad Chemical–Disease Interaction Extraction & F1 $\uparrow$ & 0.35 & 0.33 & 0.18 & 0.23 & \textemdash{} & 0.40(Mol-Instructions-7B~\cite{fang2023mol}) & \SecBestPerf{0.66} & \BestPerf{0.97} \\
\quad General Multiple Choice Question & ACC $\uparrow$ & \SecBestPerf{0.94} & 0.71 & 0.26& 0.87 & \textemdash{}  & 0.65(Mol-Instructions-7B~\cite{fang2023mol}) & {0.92} & \BestPerf{0.99} \\
\quad True/False Question & ACC $\uparrow$ & \SecBestPerf{0.67} & 0.62 &0.66 & {0.66} & \textemdash{} & 0.55(Mol-Instructions-7B~\cite{fang2023mol}) & {0.64} & \BestPerf{0.68} \\
\quad Open-Ended Question & BertScore $\uparrow$ & 0.84 & 0.82 &0.79 & 0.80 & \textemdash{} & \SecBestPerf{0.84}(Mol-Instructions-7B~\cite{fang2023mol}) & 0.83 & \BestPerf{0.87} \\
\bottomrule
\end{tabular}
\caption{Evaluation results of scientific knowledge extraction and QA tasks. 
\BestPerf{Bold} indicates the best performance, and \SecBestPerf{underline} indicates the second best.}
\label{tab:appendix_scientific_knowledge_extraction}
\end{table}
\end{landscape}

\begin{landscape}
    \begin{table}[t]
\centering
\scriptsize
\resizebox{\linewidth}{!}{
\begin{tabular}{l c c c c c c c c c} 
\toprule
Task & Metric & \makecell[c]{Gemini\\2.5-pro} & \makecell[c]{GPT\\o3} & \makecell[c]{DeepSeek\\r1} & \makecell[c]{GPT\\oss} & \makecell[c]{Non-LLM \\ Expert Models} & \makecell[c]{Specialist\\LLMs} & \ProjectName-1.7B & \ProjectName-8B \\
\midrule
\addlinespace[0.8em]
\multicolumn{9}{l}{\raisebox{1.2ex}[0pt][0pt]{\bfseries Molecular Property Regression}}\\[-0.8ex]
\quad ESOL      & RMSE $\downarrow$        & 1.28 & 1.58 & 1.48 & 3.45  & \textemdash{}  & \BestPerf{1.04}(LlaSMol-7B~\cite{yu2024llasmol})    & 1.21     & \SecBestPerf{1.08}   \\
\quad LIPO      & RMSE  $\downarrow$       & 1.21 & 1.44 &2.00 & 1.65  & \textemdash{}  & 1.01(LlaSMol-7B~\cite{yu2024llasmol})               & \SecBestPerf{0.94}   & \BestPerf{0.82}   \\
\quad Physicochemical Prediction & MAE  $\downarrow$     & 0.027 & 1.879 &2.384 & 1.329   & \textemdash{}  & 0.013(Mol-Instructions-7B~\cite{fang2023mol})             & \SecBestPerf{0.003}   & \BestPerf{0.002}   \\
\midrule
\addlinespace[0.8em]
\multicolumn{9}{l}{\raisebox{1.2ex}[0pt][0pt]{\bfseries Material Property Regression}}\\[-0.8ex]
\quad MP$_\text{Regression}$     
  & \multirow{10}{*}{$\frac{\mathbf{MAD}}{\mathbf{MAE}}$$\uparrow$}
  & 1.32\dag & 1.22\dag &0.66 & 0.95 & \BestPerf{5.32}(MatBERT-109M~\cite{rubungo2025llm4mat}) & 0.76(Gemma 2-9b-it:5S~\cite{rubungo2025llm4mat})  & 4.98|4.67\dag   & \SecBestPerf{5.12}|5.21\dag   \\
\quad SNUMAT$_\text{Regression}$ &                                   
  & 1.07\dag & 1.14\dag & 1.12 & 1.00 & 1.83(MatBERT-109M~\cite{rubungo2025llm4mat}) & 1.16(Llama 2-7b-chat:5S~\cite{rubungo2025llm4mat})  & \BestPerf{2.18}|2.18\dag   & \SecBestPerf{1.96}|2.22\dag   \\
\quad JARVISDFT                  &                                   
  & 1.09\dag & 1.2\dag &0.79 &  0.88 & 4.10(MatBERT-109M~\cite{rubungo2025llm4mat}) & 0.70(Llama 2-7b-chat:5S~\cite{rubungo2025llm4mat})  & \SecBestPerf{5.11}|5.28\dag   & \BestPerf{5.21}|5.59\dag   \\
\quad JARVISQETB                 &                                   
  & 0.53\dag & 0.53\dag &0.35 &  0.50 & 59.44(LLM-Prop-35M~\cite{rubungo2025llm4mat}) & 1.05(Llama 2-7b-chat:5S~\cite{rubungo2025llm4mat})  & \BestPerf{138.59}|148.71\dag   & \SecBestPerf{126.61}|137.61\dag  \\
\quad GNoME                      &                                   
  & 1.28\dag & 1.19\dag &0.84 &  0.73 & 15.60(LLM-Prop-35M~\cite{rubungo2025llm4mat}) & 0.50(Llama 2-7b-chat:5S~\cite{rubungo2025llm4mat})  & \SecBestPerf{19.05}|18.46\dag   & \BestPerf{22.03}|21.81\dag   \\
\quad hMOF                       &                                   
  & 0.65\dag & 0.80\dag &0.61 &  0.67 & \BestPerf{1.48}(LLM-Prop-35M~\cite{rubungo2025llm4mat}) & 0.66(Llama 2-7b-chat:5S~\cite{rubungo2025llm4mat})  & \SecBestPerf{1.05}|1.04\dag    & {0.99}|0.99\dag    \\
\quad Cantor-HEA                &                                   
  & 0.77\dag & 0.91\dag &0.53 &  0.45 & \BestPerf{8.40}(LLM-Prop-35M~\cite{rubungo2025llm4mat}) & 0.87(Llama 2-7b-chat:5S~\cite{rubungo2025llm4mat})  & \SecBestPerf{7.40}|7.38\dag    & {7.18}|7.02\dag    \\
\quad QMOF                       &                                   
  & 1.24\dag & 1.41\dag &0.51 &  0.86 & 2.05(MatBERT-109M~\cite{rubungo2025llm4mat}) & 0.93(Llama 2-7b-chat:5S~\cite{rubungo2025llm4mat})  & \BestPerf{8.55}|8.76\dag    & \SecBestPerf{8.07}|7.94\dag    \\
\quad OQMD                       &                                   
  & 1.20\dag & 1.25\dag &0.77 &  0.65 & \BestPerf{6.02}(LLM-Prop-35M~\cite{rubungo2025llm4mat}) & 1.16(Llama 2-7b-chat:5S~\cite{rubungo2025llm4mat})  & {5.29}|5.50\dag    & \SecBestPerf{5.43}|5.67\dag    \\
\quad OMDB                       &                                   
  & 0.82\dag & 0.89\dag &0.88 &  0.93 & \BestPerf{1.55}(MatBERT-109M~\cite{rubungo2025llm4mat}) & 1.01(Llama 2-7b-chat:5S~\cite{rubungo2025llm4mat})  & \SecBestPerf{1.40}|1.41\dag    & {1.39}|1.42\dag    \\
\midrule
\addlinespace[0.8em]
\multicolumn{9}{l}{\raisebox{1.2ex}[0pt][0pt]{\bfseries Biology Multi-omics Property Regression}}\\[-0.8ex]
\quad Fluorescence                    & Spearman $\uparrow$     & -0.43\dag & 0.78\dag & 0.72&  3.20 & \SecBestPerf{69.00}(ShallowCNN~\cite{shanehsazzadeh2020transfer}) & 2.57(ChatMultiOmics-8B~\cite{he2024biology})  & \BestPerf{69.36}|70.82\dag  & 66.24|68.49\dag  \\
\quad Stability$_\text{Biology-Instructions}$                       & Spearman $\uparrow$     & -9.15\dag & -7.96 &-1.65 &  -7.79 & \BestPerf{79.00}(Evoformer~\cite{rossi2019ncrna})  & 60.25(ChatMultiOmics-8B~\cite{he2024biology})  & 63.25|55.67\dag  & \SecBestPerf{64.41}|59.03\dag  \\
\quad Thermostability                 & Spearman $\uparrow$     & 5.47\dag & 15.00 & -3.65&  6.29  & \BestPerf{78.00}(ESM-1v~\cite{meier2021language}) & 45.07(ChatMultiOmics-8B~\cite{he2024biology})  & 56.25|53.17\dag  & \SecBestPerf{58.55}|54.45\dag  \\
\quad Enhancer Activity Prediction    & PCC $\uparrow$          & -4.24\dag & 0.49\dag & -2.27 & -1.29 & \BestPerf{68.00}(DeepSTARR~\cite{de2022deepstarr}) & 57.24(ChatMultiOmics-8B~\cite{he2024biology})  & \SecBestPerf{65.07}|70.20\dag  & {64.39}|52.08\dag  \\
\quad APA Isoform Prediction          & R2 $\uparrow$           & 0.13\dag & 0.24\dag & 0.03 & 0.01  & 50.82(APARENT~\cite{bogard2019deep}) & 59.01(ChatMultiOmics-8B~\cite{he2024biology})  & \BestPerf{87.94}|86.19\dag  & \SecBestPerf{85.87}|87.92\dag  \\
\quad Mean Ribosome Loading Prediction& R2 $\uparrow$           & 0.00\dag & 0.00\dag &0.06 & 0.02  & \BestPerf{78.00}(Optimus~\cite{sample2019human}) & 47.64(ChatMultiOmics-8B~\cite{he2024biology})  & 50.82|54.56\dag  & \SecBestPerf{60.44}|64.5\dag  \\
\quad Programmable RNA Switches       & R2 $\uparrow$           & 0.01\dag & 0.15\dag & 0.08 & 0.01  & \BestPerf{55.67}(MLP-O~\cite{angenent2020deep}) & 26.65(ChatMultiOmics-8B~\cite{he2024biology})  & 39.00|38.1\dag  & \SecBestPerf{43.55}|29.20\dag  \\
\quad CRISPR On Target Prediction     & Spearman $\uparrow$     & -3.64 & -2.21 & 0.74&  6.21  & \BestPerf{44.10}(SCC~\cite{xu2015sequence}) & 2.87(ChatMultiOmics-8B~\cite{he2024biology})  & 14.35  & \SecBestPerf{25.81}  \\
\quad siRNA Efficiency Prediction     & Mixed-score $\uparrow$  & 38.75\dag & 34.02 &17.68 & 40.99 & $49.38$(Meta-Biomol~\cite{ren2024comet}) & 42.92(ChatMultiOmics-8B~\cite{he2024biology})  & \BestPerf{62.56}|64.13\dag  & \SecBestPerf{59.67}|61.94\dag  \\
\bottomrule
\end{tabular}
}
\caption{Evaluation results of property prediction tasks. 
\BestPerf{Bold} indicates the best performance, and \SecBestPerf{underline} indicates the second best. \dag~denotes that only randomly sampled 1000 test samples were utilized for evaluation in the corresponding task. This measure is adopted because extremely large-scale test sets exist for certain tasks, which would lead to substantial testing costs.}
\label{tab:appendix_property_prediction}
\end{table}
\end{landscape}

\begin{landscape}
    \begin{table}[t]
\centering
\scriptsize
\begin{tabular}{l c c c c c c c c c} 
\toprule
Task & Metric & \makecell[c]{Gemini\\2.5-pro} & \makecell[c]{GPT\\o3} & \makecell[c]{DeepSeek\\r1} &  \makecell[c]{GPT\\oss} & \makecell[c]{Non-LLM \\ Expert Models} & \makecell[c]{Specialist\\ LLMs} & \ProjectName-1.7B & \ProjectName-8B \\
\midrule
\addlinespace[0.8em]
\multicolumn{9}{l}{\raisebox{1.2ex}[0pt][0pt]{\bfseries Molecular  Classification}}\\[-0.8ex]
\quad BBBP      & ACC@1 $\uparrow$      & 71.57\% & 78.68\% & 59.90\% & 46.70\%  & \textemdash{}  & 75.10\%(LlaSMol-7B~\cite{yu2024llasmol})    & \SecBestPerf{79.70\%}   & \BestPerf{82.74\%}   \\
\quad Clintox   & ACC@1 $\uparrow$      & 26.39\% & 47.92\% & 70.14\% & 52.78\%  & \textemdash{}  & \BestPerf{93.10\%}(LlaSMol-7B~\cite{yu2024llasmol})    & \SecBestPerf{92.36\%}   & 91.67\%    \\
\quad HIV       & ACC@1 $\uparrow$      & 26.7\%\dag & 77.70\%\dag & 89.3\% & 69.86\%  & \textemdash{}  & 96.70\%(LlaSMol-7B~\cite{yu2024llasmol})    & \SecBestPerf{96.79\%}|92.40\%\dag   & \BestPerf{96.81\%}|95.90\%\dag   \\
\quad Sider     & ACC@1 $\uparrow$      & 67.7\%\dag & 70.50\%\dag &61.2\% & 38.25\%  & \textemdash{}  & \BestPerf{70.70\%}(LlaSMol-7B~\cite{yu2024llasmol})    & \SecBestPerf{70.00\%}|66.20\%\dag & {68.53\%}|68.70\%\dag   \\
\midrule
\addlinespace[0.8em]
\multicolumn{9}{l}{\raisebox{1.2ex}[0pt][0pt]{\bfseries Material Property Classification}}\\[-0.8ex]
\quad MP$_\text{classification}$                 & AUC$\uparrow$        & 0.58\dag & 0.56\dag &0.57 & 0.53 & \BestPerf{0.72}(MatBERT-109M~\cite{rubungo2025llm4mat}) & 0.51(Gemma 2-9b-it:5S~\cite{rubungo2025llm4mat}) & \SecBestPerf{0.70}|0.69\dag  & \BestPerf{0.72}|0.71\dag \\
\quad SNUMAT$_\text{classification}$             & AUC$\uparrow$        & 0.59\dag & 0.63\dag &0.57 & 0.60 & \BestPerf{0.72}(LLM-Prop-35M~\cite{rubungo2025llm4mat}) & 0.58(Gemma 2-9b-it:5S~\cite{rubungo2025llm4mat}) & \SecBestPerf{0.69}|0.69  & {0.66}|0.66 \\
\midrule
\addlinespace[0.8em]
\multicolumn{9}{l}{\raisebox{1.2ex}[0pt][0pt]{\bfseries Gene-centric Association Classification}}\\[-0.8ex]
\multirow{3}{*}{\quad gSymbol2Tissue}    & Precision$\uparrow$ & 0.01 & 0.38 &0.03 & 0.03 & \textemdash{} & \SecBestPerf{0.39}(OPI-Galactica-6.7B~\cite{xiao2024opi})      & \BestPerf{0.41}  & \BestPerf{0.41} \\
                                         & Recall$\uparrow$    & 0.01 & 0.68 &0.09 & 0.09 & \textemdash{} & \BestPerf{0.91}(OPI-Galactica-6.7B)~\cite{xiao2024opi}      & 0.82  & \SecBestPerf{0.83} \\
                                         & F1 score$\uparrow$  & 0.01 & 0.45 &0.04 & 0.04 & \textemdash{} & \BestPerf{0.53}(OPI-Galactica-6.7B)~\cite{xiao2024opi}      & 0.51  & \SecBestPerf{0.52} \\
\cdashline{1-9}[.6pt/2pt] 
\multirow{3}{*}{\quad gSymbol2Cancer}    & Precision$\uparrow$ & 0.00 & 0.11 &0.00 & 0.01 & \textemdash{} & 0.36(OPI-Galactica-6.7B~\cite{xiao2024opi})      & \SecBestPerf{0.75}  & \BestPerf{0.81} \\
                                         & Recall$\uparrow$    &0.00  & 0.10 &0.02 & 0.02 & \textemdash{} & 0.32(OPI-Galactica-6.7B~\cite{xiao2024opi})      & \SecBestPerf{0.72}  & \BestPerf{0.81} \\
                                         & F1 score$\uparrow$  &0.00 & 0.10 &0.00 & 0.01 & \textemdash{} & 0.32(OPI-Galactica-6.7B~\cite{xiao2024opi})      & \SecBestPerf{0.73}  & \BestPerf{0.81} \\
\cdashline{1-9}[.6pt/2pt] 
\multirow{3}{*}{\quad gName2Cancer}      & Precision $\uparrow$ &0.00  & 0.11 &0.00 & 0.01 & \textemdash{} & 0.27(OPI-Galactica-6.7B~\cite{xiao2024opi})      & \SecBestPerf{0.68}  & \BestPerf{0.80} \\
                                         & Recall $\uparrow$    &0.00  & 0.10 &0.03 & 0.04 & \textemdash{} & 0.26(OPI-Galactica-6.7B~\cite{xiao2024opi})      & \SecBestPerf{0.67}  & \BestPerf{0.79} \\
                                         & F1 score $\uparrow$  &0.00  & 0.10 &0.00  & 0.01 & \textemdash{} & 0.25(OPI-Galactica-6.7B~\cite{xiao2024opi})      & \SecBestPerf{0.67}  & \BestPerf{0.79} \\
\bottomrule
\end{tabular}
\caption{Evaluation results of molecular classification tasks, material property classification tasks, and gene-centric association classification tasks. \BestPerf{Bold} indicates the best performance, and \SecBestPerf{underline} indicates the second best. \dag~denotes that only randomly sampled 1000 test samples were utilized for evaluation in the corresponding task. This measure is adopted because extremely large-scale test sets exist for certain tasks, which would lead to substantial testing costs.}
\label{tab:appendix_mat_gene_cls}
\end{table}

\end{landscape}

\begin{landscape}
\centering
\scriptsize
\setlength{\tabcolsep}{4pt}
% \resizebox{\linewidth}{!}{
\begin{longtable}{l c c c c c c c c c}
\toprule
Task & Metric & \makecell[c]{Gemini\\2.5-pro} & \makecell[c]{GPT\\o3} & \makecell[c]{DeepSeek\\r1} & \makecell[c]{GPT\\oss} & \makecell[c]{Non-LLM \\ Expert Models} & \makecell[c]{Specialist \\ LLMs} & \makecell[c]{\ProjectName\\1.7B} & \makecell[c]{\ProjectName\\8B} \\
\midrule
\endhead 

\midrule
\multicolumn{8}{r}{\textit{Continued on next page}} \\
\endfoot 

\bottomrule
\caption{Evaluation results of biology multi-omics property classification tasks. Dashed lines separate tasks that report multiple metrics (Precision/Recall/F1). \BestPerf{Bold} indicates the best performance, and \SecBestPerf{underline} indicates the second best.}
\label{tab:appendix_bio_prop_cls_long}
\endlastfoot

\addlinespace[0.8em]
\multicolumn{9}{l}{\raisebox{1.2ex}[0pt][0pt]{\bfseries Biology Multi-omics Property Classification (binary)}}\\[-0.8ex]

\quad Solubility$_\text{Biology-Instructions}$ & ACC $\uparrow$ & $51.40$\dag & $50.83$ &51.82 & $51.22$ & \BestPerf{77.00}(DeepSol~\cite{khurana2018deepsol})& $63.02$(ChatMultiOmics-8B~\cite{he2024biology}) & $62.67$|67.3\dag & \SecBestPerf{71.61}|69.4\dag \\
\quad Antibody\mbox{-}Antigen & MCC $\uparrow$ & $-5.05$\dag  & $1.32$ &-6.05 & $-2.78$ &  \BestPerf{54.90}(DeepAAI~\cite{zhang2022predicting}) & $1.06$(ChatMultiOmics-8B~\cite{he2024biology}) & 41.59|45.43\dag & \SecBestPerf{45.5}|48.84\dag \\
\quad RNA\mbox{-}Protein & MCC $\uparrow$ & $1.20$\dag & $-4.40$ & -15.97 & $-5.23$ & \BestPerf{93.2}(ncRPI-LGAT~\cite{han2023ncrpi})  & $74.26$(ChatMultiOmics-8B~\cite{he2024biology}) & \SecBestPerf{86.47}|85.97\dag & $84.97$|88.8\dag \\
\quad Epigenetic Marks Prediction & $\uparrow$ & 0.52\dag  & 2.40\dag &3.23 & 0.84 & \BestPerf{58.83}(DNABert-2~\cite{zhou2023dnabert}) & 3.64(ChatMultiOmics-8B~\cite{he2024biology}) & 14.16|18.09\dag & \SecBestPerf{17.24}|21.26\dag \\
\quad TF\mbox{-}Mouse$_0$ & MCC $\uparrow$ & $0.20$\dag & $11.93$\dag &4.70 & $4.75$ & \BestPerf{64.23}(DNABert-2~\cite{zhou2023dnabert}) & 26.50 (ChatMultiOmics-8B~\cite{he2024biology}) & 41.79|39.77\dag & \SecBestPerf{43.52}|40.42\dag \\
\quad TF\mbox{-}Mouse$_1$ & MCC $\uparrow$ & $-1.24$\dag & $-3.27$\dag &1.95 & $-0.38$ & \BestPerf{86.28}(DNABert-2~\cite{zhou2023dnabert}) & 66.96 (ChatMultiOmics-8B~\cite{he2024biology}) & 68.95|72.07\dag & \SecBestPerf{71.03}|75.98\dag \\
\quad TF\mbox{-}Mouse$_2$ & MCC $\uparrow$ & $0.41$\dag & $5.97$\dag &21.34 & $6.99$ & \BestPerf{81.28}(DNABert-2~\cite{zhou2023dnabert}) & 63.98 (ChatMultiOmics-8B~\cite{he2024biology}) & 71.57|69.08\dag & \SecBestPerf{75.07}|68.52\dag \\
\quad TF\mbox{-}Mouse$_3$ & MCC $\uparrow$ & $2.72$\dag & $3.00$\dag &4.68 & $-9.63$ & \BestPerf{73.49}(DNABert-2~\cite{zhou2023dnabert}) & 32.02 (ChatMultiOmics-8B~\cite{he2024biology}) & 44.81|45.10\dag & \SecBestPerf{52.98}|53.31\dag \\
\quad TF\mbox{-}Mouse$_4$ & MCC $\uparrow$ & $6.61$\dag & $3.58$\dag &1.44 & $-2.07$ & \BestPerf{52.66}(DNABert-2~\cite{zhou2023dnabert}) & 9.18 (ChatMultiOmics-8B~\cite{he2024biology}) & 27.34|28.45\dag & \SecBestPerf{28.62}|28.20\dag \\
\quad Enhancer\mbox{-}Promoter Interaction & MCC $\uparrow$ & $-3.75$\dag & $-3.24$ & 0 & $-1.69$ & \BestPerf{53.59}(EPI-DLMH~\cite{min2021predicting}) & $4.13$(ChatMultiOmics-8B~\cite{he2024biology}) & $8.18$|6.75\dag & \SecBestPerf{14.40}|8.12\dag \\
\quad Promoter\mbox{-}300$_\text{all}$ & MCC $\uparrow$ & 24.12 & 13.04 &17.37 & -3.56 & \BestPerf{91.01}(NT-2500M-multi~\cite{zhou2023dnabert}) & 78.97(ChatMultiOmics-8B~\cite{he2024biology}) & 82.94 & \SecBestPerf{83.01} \\
\quad Promoter\mbox{-}300$_\text{notata}$ & MCC $\uparrow$ & 16.65 & 20.31 &16.60 & -2.40 & \BestPerf{94.34}(DNABert-2~\cite{zhou2023dnabert}) & 83.45(ChatMultiOmics-8B~\cite{he2024biology}) & \SecBestPerf{88.46} & 87.04 \\
\quad Promoter\mbox{-}300$_\text{tata}$ & MCC $\uparrow$ & 11.98 & 15.12 &12.60 & 13.04 & \BestPerf{79.43}(NT-2500M-multi~\cite{zhou2023dnabert}) & 38.50(ChatMultiOmics-8B~\cite{he2024biology}) & 44.38 & \SecBestPerf{51.37} \\
\quad Promoter\mbox{-}core$_\text{all}$ & MCC $\uparrow$ & 12.38 & -3.18 &-12.13 & -3.57 & \BestPerf{70.92}(DNABert 3-mer~\cite{zhou2023dnabert}) & 59.98(ChatMultiOmics-8B~\cite{he2024biology}) & \SecBestPerf{61.49} & 62.56 \\
\quad Promoter\mbox{-}core$_\text{notata}$ & MCC $\uparrow$ & 15.64 & 18.43 &-13.70 & -6.37 & \BestPerf{71.58}(NT-2500M-multi~\cite{zhou2023dnabert}) & 62.85(ChatMultiOmics-8B~\cite{he2024biology}) & 61.72 & \SecBestPerf{63.52} \\
\quad Promoter\mbox{-}core$_\text{tata}$ & MCC $\uparrow$ & 13.02 & 16.07 &11.90 & 17.97 & \BestPerf{78.15}(DNABert 3mer~\cite{zhou2023dnabert}) & 39.36(ChatMultiOmics-8B~\cite{he2024biology}) & \SecBestPerf{53.67} & 46.94 \\
\quad TF\mbox{-}Human$_0$ & MCC $\uparrow$ & -3.46 & 2.41 &5.97 & 3.24 & \BestPerf{71.99}(DNABert-2~\cite{zhou2023dnabert}) & 56.56(ChatMultiOmics-8B~\cite{he2024biology}) & 59.33 & \SecBestPerf{61.07} \\
\quad TF\mbox{-}Human$_1$ & MCC $\uparrow$ & 12.23 & 11.08 &4.42 & 3.85 & \BestPerf{76.06}(DNABert-2~\cite{zhou2023dnabert}) & 61.17(ChatMultiOmics-8B~\cite{he2024biology}) & 62.22 & \SecBestPerf{63.47} \\
\quad TF\mbox{-}Human$_2$ & MCC $\uparrow$ & 2.29 & 7.20 &0.61 & 5.19 & \BestPerf{66.52}(DNABert-2~\cite{zhou2023dnabert}) & 46.92(ChatMultiOmics-8B~\cite{he2024biology}) & 47.80 & \SecBestPerf{48.02} \\
\quad TF\mbox{-}Human$_3$ & MCC $\uparrow$ & -8.44 & 0.00 &-0.40 & -1.27 & \BestPerf{58.54}(DNABert-2~\cite{zhou2023dnabert}) & 34.16(ChatMultiOmics-8B~\cite{he2024biology}) & 38.96 & \SecBestPerf{40.89} \\
\quad TF\mbox{-}Human$_4$ & MCC $\uparrow$ & 7.56 & 15.66 &7.61 & 15.06 & \BestPerf{77.43}(DNABert-2~\cite{zhou2023dnabert}) & 51.86(ChatMultiOmics-8B~\cite{he2024biology}) & \SecBestPerf{56.43} & 56.10 \\

\quad Solubility$_\text{PEER}$ & ACC $\uparrow$ & 58.40\dag & 53.0\dag &57.58 & 10.91 & \BestPerf{74.37}(BioT5+~\cite{pei2024biot5+}) & 71.40(NatureLM~\cite{xia2025naturelm}) & 70.14|70.8\dag & \SecBestPerf{73.99}|75.4\dag \\
\quad Stability$_\text{PEER}$ & ACC $\uparrow$ & 49.60\dag & 45.40\dag &50.63 & 38.21 & 58.30(SFM-Protein~\cite{he2024sfm}) & {63.50}(NatureLM-8B~\cite{xia2025naturelm}) & \BestPerf{68.41}|68.90\dag & \SecBestPerf{64.70}|64.3\dag \\
\quad Human PPI & ACC $\uparrow$ & 52.74 & 57.4 &51.48 & 58.2 & \BestPerf{85.09}(BioT5+~\cite{pei2024biot5+}) & \SecBestPerf{84.8}(NatureLM-8B~\cite{xia2025naturelm}) & {76.79} & 73.42 \\
\quad Yeast PPI & ACC $\uparrow$ & 51.52 & 50.00 &56.60 & 44.90 & \BestPerf{66.16}(BioT5+~\cite{pei2024biot5+}) & {$60.40$}(NatureLM-8B~\cite{xia2025naturelm}) & 53.81 & \SecBestPerf{60.41} \\

\midrule
\addlinespace[0.8em]
\multicolumn{9}{l}{\raisebox{1.2ex}[0pt][0pt]{\bfseries Biology Multi-omics Property Classification (multi-label)}}\\[-0.8ex]
\quad Protein Function   & ROUGE-L   & 0.20\dag & $0.07$\dag &0.05 & 0.05 &  {0.57}(BioT5+~\cite{pei2024biot5+}) & 0.43(Mol-Instructions-7B~\cite{fang2023mol}) & \SecBestPerf{0.63}|0.63 & \BestPerf{0.97}|0.96 \\
\quad Domain/Motif       & ROUGE-L   & 0.19\dag & $0.04$\dag &0.03 & 0.02 & 0.53(BioT5+~\cite{pei2024biot5+}) & 0.46(Mol-Instructions-7B~\cite{fang2023mol}) & \SecBestPerf{0.56}|0.56 & \BestPerf{0.97}|0.90 \\
\quad FunctionEC         & Fmax$\uparrow$ & $3.3$\dag & $4.49$\dag &5.27 & $22.86$  & \BestPerf{88.9}(SaProt-GearNet~\cite{su2023saprot}) & $19.79$(ChatMultiOmics-8B~\cite{he2024biology}) & 60.65|60.33\dag & \SecBestPerf{77.26}|77.56\dag \\
\quad Non\mbox{-}coding RNA Func. Classif. & ACC $\uparrow$ & $17.8$\dag  & $6.20$\dag & 10.31& $11.54$ & \SecBestPerf{85.73}(GCN~\cite{rossi2019ncrna}) & $63.09$(ChatMultiOmics-8B~\cite{he2024biology}) & 80.39|88.8\dag & \BestPerf{86.65}|93.4\dag \\
\quad Modification Prediction & AUC $\uparrow$ & $51.19$ & 50.96  &49.92 & 50.05 & \BestPerf{84.00}(MultiRM~\cite{song2021attention}) & 59.06(ChatMultiOmics-8B~\cite{he2024biology}) & 53.06 & \SecBestPerf{59.94} \\
\quad Subcellular Localization & Accuracy$\uparrow$ & 0.02 & 0.08 &0.05 & 0 & \textemdash{} & 0.78(OPI-Galactica-6.7B~\cite{xiao2024opi})  & \SecBestPerf{0.84}  & \BestPerf{0.91}  \\
\quad Fold Type & Accuracy$\uparrow$ & 0.01 & 0.01&0.01 & 0 & \textemdash{} & \BestPerf{0.61}(OPI-Llama-3.1-8B-Instruct)~\cite{xiao2024opi}  & 0.34  & \SecBestPerf{0.48}  \\
\cdashline{1-9}[.6pt/2pt] 
\quad Numerical EC (New\mbox{-}392) & \makecell{Precision$\uparrow$\\Recall$\uparrow$\\F1$\uparrow$} & \makecell{0\\0\\0} & \makecell{0.01\\0\\0} & \makecell{0\\0\\0} & \makecell{0\\0\\0} & \makecell{\BestPerf{0.59}(CLEAN~\cite{yu2023enzyme})\\\SecBestPerf{0.48}(CLEAN~\cite{yu2023enzyme})\\\SecBestPerf{0.49}(CLEAN~\cite{yu2023enzyme})} & \makecell{0.37(OPI-Llama-3.1-8B-Instruct)~\cite{xiao2024opi})\\0.34(OPI-Llama-3.1-8B-Instruct)~\cite{xiao2024opi}\\0.35(OPI-Llama-3.1-8B-Instruct)~\cite{xiao2024opi}} & \makecell{0.40\\0.39\\0.39} & \makecell{\SecBestPerf{0.55}\\\BestPerf{0.55}\\\BestPerf{0.54}} \\
\quad Numerical EC (Price\mbox{-}149) & \makecell{Precision$\uparrow$\\Recall$\uparrow$\\F1$\uparrow$} & \makecell{0\\0\\0} & \makecell{0\\0\\0} & \makecell{0\\0\\0} & \makecell{0\\0\\0} & \makecell{\BestPerf{0.58}(CLEAN~\cite{yu2023enzyme})\\\BestPerf{0.47}(CLEAN~\cite{yu2023enzyme})\\\BestPerf{0.49}(CLEAN~\cite{yu2023enzyme})} & \makecell{0.07(OPI-Llama-3.1-8B-Instruct)~\cite{xiao2024opi}\\0.07(OPI-Llama-3.1-8B-Instruct)~\cite{xiao2024opi}\\0.07(OPI-Llama-3.1-8B-Instruct)~\cite{xiao2024opi}} & \makecell{0.10\\0.10\\0.10} & \makecell{\SecBestPerf{0.21}\\\SecBestPerf{0.23}\\\SecBestPerf{0.22}} \\
\cdashline{1-9}[.6pt/2pt] 
\quad Keywords (CASPSimilarSeq) & \makecell{Precision$\uparrow$\\Recall$\uparrow$\\F1$\uparrow$} & \makecell{0.04\\0.03\\0.03} & \makecell{0.21\\0.19\\0.19} & \makecell{0\\0.04\\0.01} & \makecell{0\\0.02\\0.01} & \makecell{-\\-\\-} & \makecell{0.81(OPI-Galactica-6.7B~\cite{xiao2024opi})\\0.74(OPI-Galactica-6.7B~\cite{xiao2024opi})\\0.76(OPI-Galactica-6.7B~\cite{xiao2024opi})} & \makecell{\SecBestPerf{0.83}\\\SecBestPerf{0.80}\\\SecBestPerf{0.81}} & \makecell{\BestPerf{0.86}\\\BestPerf{0.83}\\\BestPerf{0.84}} \\
\quad Keywords (IDFilterSeq) & \makecell{Precision$\uparrow$\\Recall$\uparrow$\\F1$\uparrow$} & \makecell{0.03\\0.02\\0.01} & \makecell{0.17\\0.16\\0.16} & \makecell{0.01\\0\\0} & \makecell{0\\0.02\\0.01} & \makecell{-\\-\\-} & \makecell{0.84(OPI-Galactica-6.7B~\cite{xiao2024opi})\\0.80(OPI-Galactica-6.7B~\cite{xiao2024opi})\\0.82(OPI-Galactica-6.7B~\cite{xiao2024opi})} & \makecell{\SecBestPerf{0.86}\\\SecBestPerf{0.85}\\\SecBestPerf{0.85}} & \makecell{\BestPerf{0.89}\\\BestPerf{0.89}\\\BestPerf{0.88}} \\
\quad Keywords (UniProtSeq) & \makecell{Precision$\uparrow$\\Recall$\uparrow$\\F1$\uparrow$} & \makecell{0.01\\0\\0} & \makecell{0.2\\0.18\\0.18} & \makecell{0.01\\0\\0} & \makecell{0.01\\0.03\\0.01} & \makecell{-\\-\\-} & \makecell{0.86(OPI-Galactica-6.7B~\cite{xiao2024opi})\\0.82(OPI-Galactica-6.7B~\cite{xiao2024opi})\\0.83(OPI-Galactica-6.7B~\cite{xiao2024opi})} & \makecell{\SecBestPerf{0.85}\\\SecBestPerf{0.90}\\\SecBestPerf{0.88}} & \makecell{\BestPerf{0.92}\\\BestPerf{0.91}\\\BestPerf{0.91}} \\
\cdashline{1-9}[.6pt/2pt] 
\quad GO Terms (CASPSimilarSeq) & \makecell{Precision$\uparrow$\\Recall$\uparrow$\\F1$\uparrow$} & \makecell{0\\0\\0} & \makecell{0\\0\\0} & \makecell{0\\0\\0} & \makecell{0\\0\\0} & \makecell{-\\-\\-} & \makecell{0.76(OPI-Galactica-6.7B~\cite{xiao2024opi})\\0.74(OPI-Galactica-6.7B~\cite{xiao2024opi})\\0.75(OPI-Galactica-6.7B~\cite{xiao2024opi})} & \makecell{\SecBestPerf{0.81}\\\SecBestPerf{0.81}\\\SecBestPerf{0.80}} & \makecell{\BestPerf{0.86}\\\BestPerf{0.86}\\\BestPerf{0.86}} \\
\quad GO Terms (IDFilterSeq) & \makecell{Precision$\uparrow$\\Recall$\uparrow$\\F1$\uparrow$} & \makecell{0\\0\\0} & \makecell{0\\0\\0} & \makecell{0\\0\\0} & \makecell{0\\0\\0} & \makecell{-\\-\\-} & \makecell{0.74(OPI-Galactica-6.7B~\cite{xiao2024opi})\\0.73(OPI-Galactica-6.7B~\cite{xiao2024opi})\\0.72(OPI-Galactica-6.7B~\cite{xiao2024opi})} & \makecell{\SecBestPerf{0.78}\\\SecBestPerf{0.77}\\\SecBestPerf{0.77}} & \makecell{\BestPerf{0.83}\\\BestPerf{0.82}\\\BestPerf{0.81}} \\
\quad GO Terms (UniProtSeq) & \makecell{Precision$\uparrow$\\Recall$\uparrow$\\F1$\uparrow$} &\makecell{0\\0\\0}  & \makecell{0\\0\\0} & \makecell{0\\0\\0} & \makecell{0\\0\\0} & \makecell{-\\-\\-} & \makecell{0.76(OPI-Galactica-6.7B~\cite{xiao2024opi})\\0.74(OPI-Galactica-6.7B~\cite{xiao2024opi})\\0.74(OPI-Galactica-6.7B~\cite{xiao2024opi})} & \makecell{\SecBestPerf{0.84}\\\SecBestPerf{0.82}\\\SecBestPerf{0.81}} & \makecell{\BestPerf{0.88}\\\BestPerf{0.87}\\\BestPerf{0.86}} \\
\end{longtable}
% }
\end{landscape}

\begin{landscape}
\setlength{\tabcolsep}{6pt}
\begin{table}[ht]
\centering
\scriptsize
\resizebox{\linewidth}{!}
{
\begin{tabular}{l c c c c c c c c c}
\toprule
Task & Metric & \makecell[c]{Gemini\\2.5-pro} & \makecell[c]{GPT\\o3} & \makecell[c]{DeepSeek\\r1} & \makecell[c]{GPT\\oss} & \makecell[c]{Non-LLM \\ Expert Models}& \makecell[c]{Specialist\\ LLMs} & \makecell[c]{\ProjectName\\1.7B} & \makecell[c]{\ProjectName\\8B} \\
\midrule
\addlinespace[0.8em]
\multicolumn{9}{l}{\raisebox{1.2ex}[0pt][0pt]{\bfseries Unconditional Generation}}\\[-0.8ex]
\multirow{2}{*}{\quad Unconditional Molecular Generation}
  & Validity $\uparrow$ & 0.568 & 0.715 & 0.698 & 0.952 & \textemdash{} & \SecBestPerf{0.968}(NatureLM-8B~\cite{xia2025naturelm}) & 0.864 & \BestPerf{0.971} \\
  & Unique $\uparrow$   & 0.042 & 0.364 & 0.093 & 0.318 & \textemdash{} & 0.966(NatureLM-8B~\cite{xia2025naturelm}) & \BestPerf{0.988} & \SecBestPerf{0.982} \\
  \cdashline{1-9}[.6pt/2pt] 
  
\multirow{3}{*}{\quad Unconditional Protein Generation} 
& Avg. Length$\uparrow$ & 128.87 & 153.56 & 260.20 & \BestPerf{318.91} & \textemdash{} & 284.50(NatureLM-8B~\cite{xia2025naturelm}) & 280.10 & \SecBestPerf{284.7} \\
& Diversity$\uparrow$ & 0.62 & \SecBestPerf{0.99} &0.93 & \BestPerf{1.00} & \textemdash{} & 0.97 (NatureLM-8B~\cite{xia2025naturelm}) & \BestPerf{1.00} & 0.91 \\
& Validity$\uparrow$ & 0.93 & {0.97} &\BestPerf{1.00} & \SecBestPerf{0.99} & \textemdash{} & \textemdash{} & 0.80 & \SecBestPerf{0.99} \\
\cdashline{1-9}[.6pt/2pt]

\quad Unconditional RNA Generation     & Avg. MFE$\downarrow$ & N/A  & -151.70 & N/A & \BestPerf{-417.58} & \textemdash{} & -177.1(NatureLM-8x7B~\cite{xia2025naturelm}) & -178.85 & \SecBestPerf{-200.14} \\
\cdashline{1-9}[.6pt/2pt]
\quad Unconditional Material Generation & SMACT$\uparrow$ & \BestPerf{89.70}  & N/A &  N/A & N/A & \textemdash{} & 66.07(NatureLM-8x7B~\cite{xia2025naturelm}) & 55.06 & \SecBestPerf{88.18} \\
\midrule
\addlinespace[0.8em]
\multicolumn{9}{l}{\raisebox{1.2ex}[0pt][0pt]{\bfseries Conditional Material Generation}}\\[-0.8ex]
\multirow{3}{*}{\quad Composition to Material Generation}
  & SMACT  $\uparrow$   & N/A & N/A &  N/A & N/A & \textemdash{} & 83.36(NatureLM-8B~\cite{xia2025naturelm}) & \BestPerf{86.07} & \SecBestPerf{85.85} \\
  & Precision $\uparrow$& N/A & N/A& N/A & N/A & \textemdash{} & 98.44(NatureLM-8B~\cite{xia2025naturelm}) & \SecBestPerf{99.89} & \BestPerf{99.96} \\
  & Novelty $\downarrow$& N/A & N/A & N/A & N/A & \textemdash{} & \BestPerf{97.13}(NatureLM-1B~\cite{xia2025naturelm}) & \SecBestPerf{85.64} & 84.81 \\
  \cdashline{1-9}[.6pt/2pt]
\multirow{2}{*}{\quad Bulk to Material Generation}
  & SMACT  $\uparrow$   & N/A & N/A & N/A & N/A & \textemdash{} & \BestPerf{94.75}(NatureLM-8x7B~\cite{xia2025naturelm}) & \SecBestPerf{88.92} & 87.49 \\
  & Success rate        & N/A & N/A & N/A & N/A & \textemdash{} & \textemdash{} & \BestPerf{7.12}  & \SecBestPerf{7.05} \\
\midrule
\addlinespace[0.8em]
\multicolumn{9}{l}{\raisebox{1.2ex}[0pt][0pt]{\bfseries Conditional Small Molecular Generation}}\\[-0.8ex]
\quad Forward Synthesis Prediction        & Exact Match $\uparrow$& 0.32\dag & 0.27\dag & 0.10& 0.33 & \textemdash{} & 0.63(LlaSMol-7B~\cite{yu2024llasmol}) & \SecBestPerf{0.69}|0.81 & \BestPerf{0.75}|0.83 \\
\quad Forward Reaction Prediction         & Exact Match $\uparrow$            &0.55  & 0.58 &0.48 & 0.36 & 0.86(BioT5+~\cite{pei2024biot5+}) & 0.54(InstructMol-GS-6.9B~\cite{cao2025instructmol}) & \SecBestPerf{0.96} & \BestPerf{0.99} \\
\quad Reagent Prediction                  & Exact Match $\uparrow$            & 0.04 &0.01 &0.01 & 0.00 & \BestPerf{0.26}(BioT5+~\cite{pei2024biot5+}) & 0.13(InstructMol-GS-6.9B~\cite{cao2025instructmol}) & 0.19 & \SecBestPerf{0.22} \\
\quad Retrosynthesis$_\text{Mol-Instructions}$           & Exact Match $\uparrow$            & 0.20 & 0.16 &0.05 & 0.05 & 0.64(BioT5+~\cite{pei2024biot5+}) & 0.41(InstructMol-GS-6.9B~\cite{cao2025instructmol}) & \SecBestPerf{0.72} & \BestPerf{0.83} \\
\quad Retrosynthesis$_\text{USPTO-50K}$           & Exact Match $\uparrow$            & 0.23\dag & 0.07\dag &0.03 & 0.12 & 0.61(EditRetro~\cite{han2024retrosynthesis}) & \SecBestPerf{0.70}(NatureLM-8B~\cite{xia2025naturelm}) & {0.68}|0.72 & \BestPerf{0.72}|0.74 \\
\quad Retrosynthesis$_\text{SMol-Instruction}$           & Exact Match $\uparrow$            & 0.12\dag & 0.04\dag &0.01 & 0.08 & \textemdash{} & 0.33(LlaSMol-7B~\cite{yu2024llasmol}) & \SecBestPerf{0.41}|0.43\dag & \BestPerf{0.46}|0.47\dag \\
\quad Molecule Generation                 & Exact Match $\uparrow$ & 0.25\dag & 0.16\dag &0.19 & 0.03 & \textemdash{} & 0.19(LlaSMol-7B~\cite{yu2024llasmol}) & 0.39|0.34\dag & 0.48|0.46\dag \\
\quad Description-Guided Molecule Design  & Exact Match $\uparrow$            &0.06  &0.04 &0.03 &0.04  & \SecBestPerf{0.11}(MolT5~\cite{edwards2022translation}) & 0.002(Mol-Instructions-7B~\cite{fang2023mol}) & 0.10 & \BestPerf{0.12} \\
\midrule
\addlinespace[0.8em]
\multicolumn{9}{l}{\raisebox{1.2ex}[0pt][0pt]{\bfseries Conditional Biochemical Generation}}\\[-0.8ex]
\quad Function-guided Protein Design      & Max. Normalized SW $\uparrow$    &0.00\dag  & 0.00\dag &0.00 &0.00\dag  & \SecBestPerf{0.97}(BioT5+~\cite{pei2024biot5+}) & 0.43(Mol-Instructions-7B~\cite{fang2023mol}) & {0.96}|0.90\dag & \BestPerf{1.00}|1.00\dag \\
\quad Catalytic Activity         & ROUGE-L $\uparrow$  & 0.25\dag & 0.09\dag &0.04 & 0.02 & 0.72(BioT5+~\cite{pei2024biot5+}) & 0.52(Mol-Instructions-7B~\cite{fang2023mol}) & \SecBestPerf{0.77}|0.78\dag & \BestPerf{0.99}|0.95\dag \\
\bottomrule
\end{tabular}
}
\caption{Evaluation results of generation and design tasks. N/A indicates that the model cannot generate samples that meet the required specifications. \BestPerf{Bold} indicates the best performance, and \SecBestPerf{underline} indicates the second best.}
\label{tab:appendix_generation_design}
\end{table}
\end{landscape}

\clearpage

\subsection*{Protein Generation}

\begin{figure}[h]
\centering
\begin{subfigure}{0.31\textwidth}
\includegraphics[width=\linewidth]{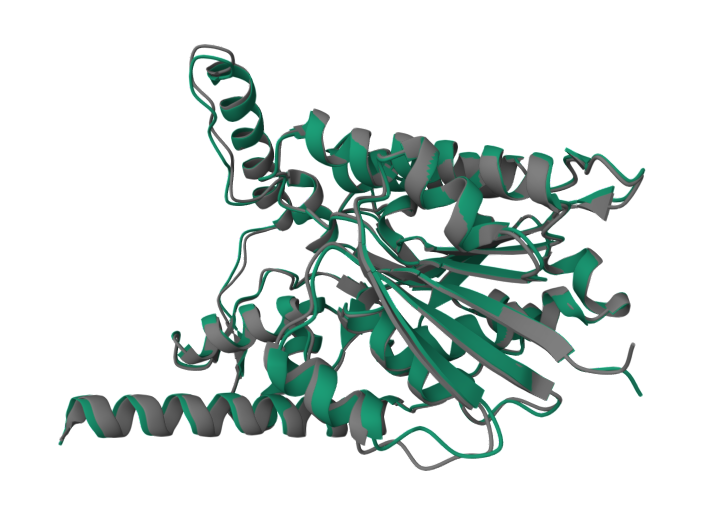}
\end{subfigure}\hfill
\begin{subfigure}{0.31\textwidth}
\includegraphics[width=\linewidth]{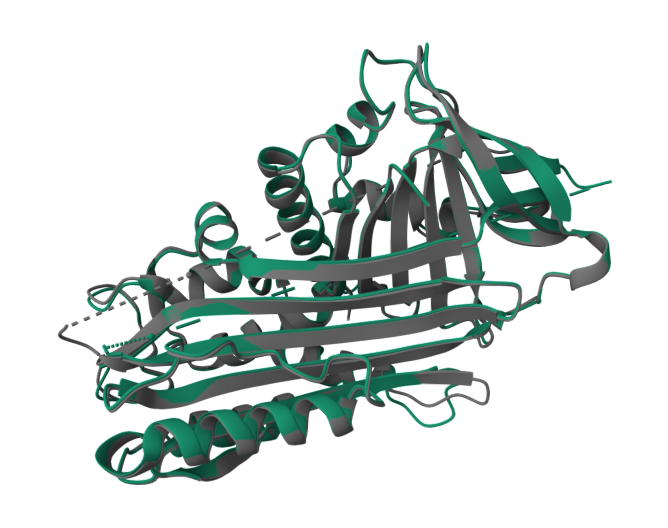}
\end{subfigure}\hfill
\begin{subfigure}{0.31\textwidth}
\includegraphics[width=\linewidth]{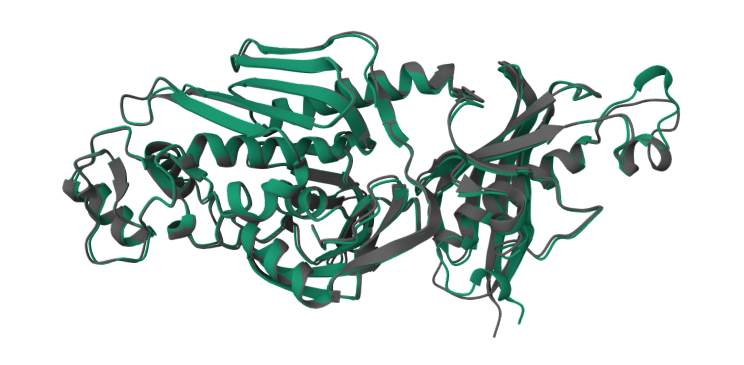}
\end{subfigure}\hfill
\caption{Additional Visualization for Protein Generation task (a).}
\label{fig:protein_translation_figure_b}
\end{figure}

\begin{figure}[h]
\centering
\begin{subfigure}{0.24\textwidth}
\includegraphics[width=\linewidth]{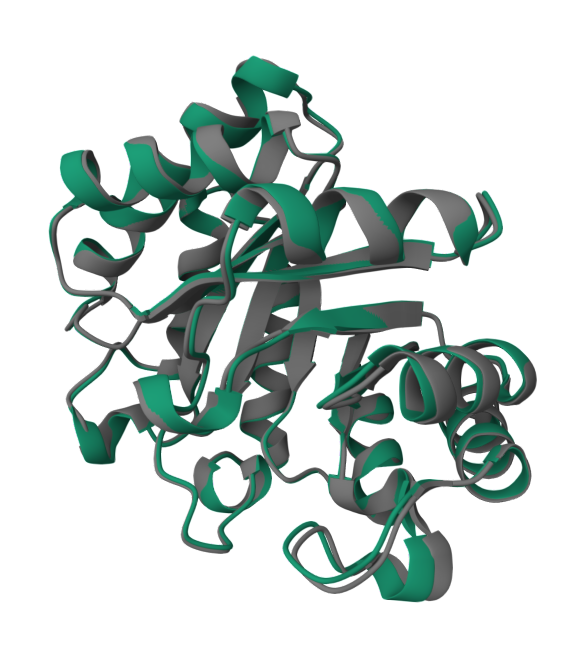}
\end{subfigure}\hfill
\begin{subfigure}{0.24\textwidth}
\includegraphics[width=\linewidth]{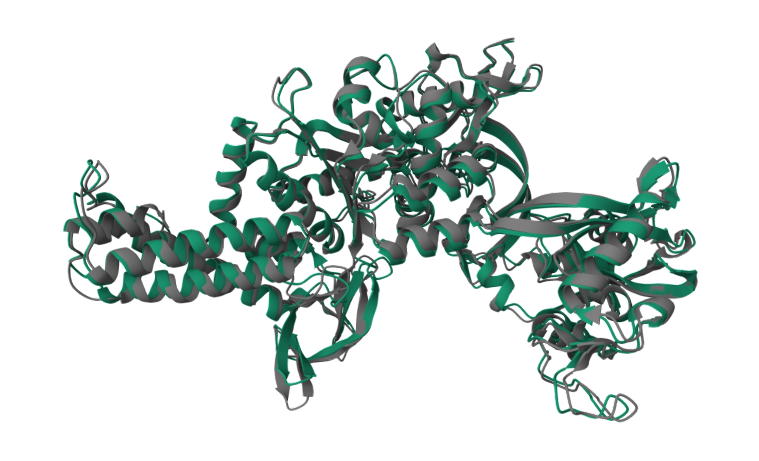}
\end{subfigure}\hfill
\begin{subfigure}{0.24\textwidth}
\includegraphics[width=\linewidth]{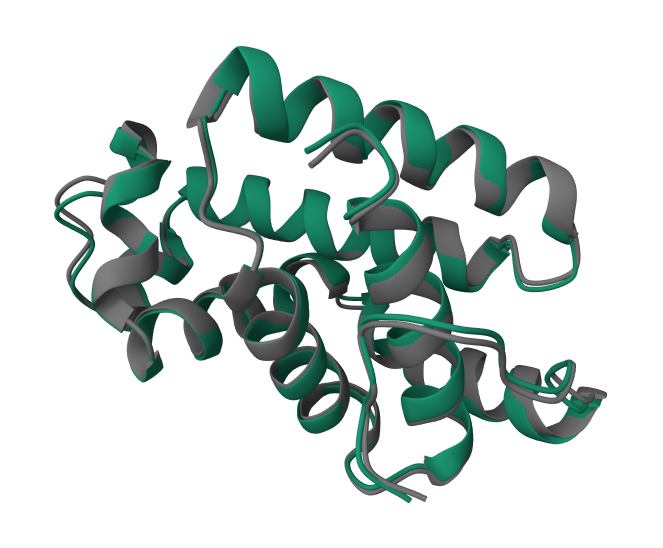}
\end{subfigure}\hfill
\begin{subfigure}{0.24\textwidth}
\includegraphics[width=\linewidth]{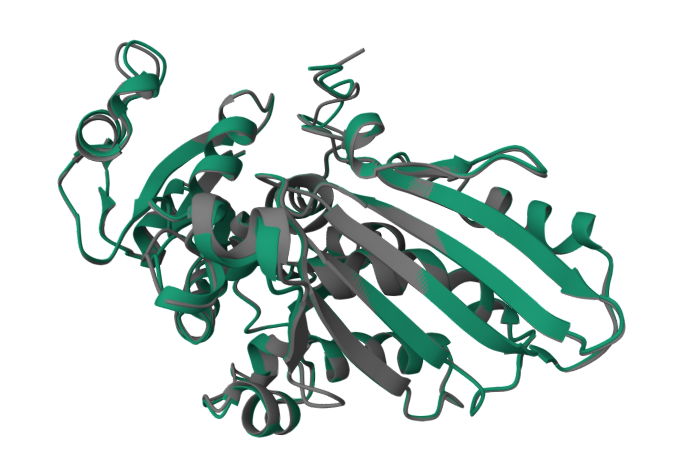}
\end{subfigure}\hfill
\caption{Additional Visualization for Protein Generation task (b).}
\label{fig:protein_translation_figure_c}
\end{figure}

\subsection*{Translation between SMILES and IUPAC}
\begin{figure}[h]
    \centering
    \includegraphics[width=1.0\linewidth]{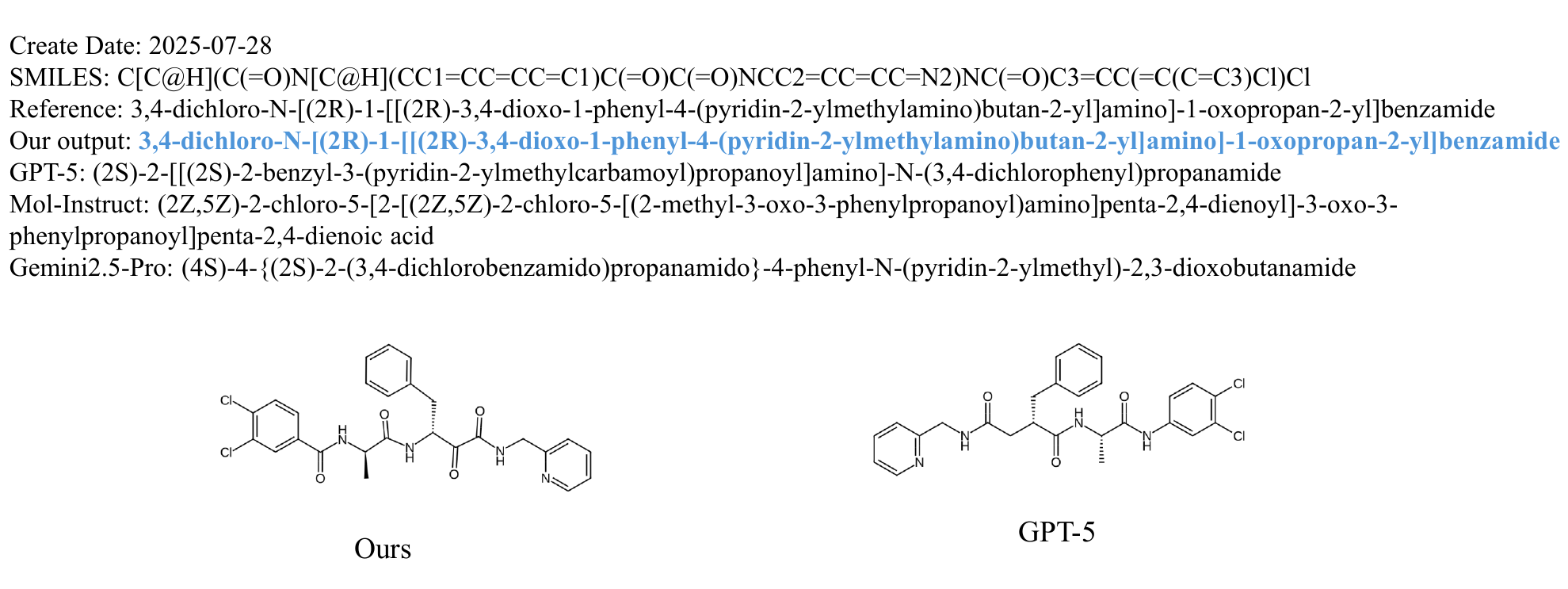}
    \caption{Additional Visualization for  SMILES-to-IUPAC Translation task}
    \label{fig:mol_translation_appendix}
\end{figure}

\clearpage

\subsection*{Task Examples}

\paragraph{ESOL}
\begin{framed}
    \small\textit{Example 1:}\\
    \texttt{\textbf{Prompt}: How soluble is <SMILES> CC(C)Cl </SMILES> ?}\\
    \texttt{\textbf{Response}: Its log solubility is -1.41 mol/L.}
\end{framed}
This task has 112 test samples.

\paragraph{LIPO}
\begin{framed}
    \small\textit{Example 1:}\\
    \texttt{\textbf{Prompt}: Predict the octanol/water distribution coefficient logD under the circumstance of pH 7.4 for <SMILES> NC(=O)C1=CC=CC=C1O </SMILES> .}\\
    \texttt{\textbf{Response}: 1.090}
\end{framed}
This task has 420 test samples.

\paragraph{BBBP}
\begin{framed}
    \small\textit{Example 1:}\\
    \texttt{\textbf{Prompt}: Is blood-brain barrier permeability (BBBP) a property of <SMILES> CCNC(=O)/C=C/C1=CC=CC(Br)=C1 </SMILES>?}\\
    \texttt{\textbf{Response}: Yes}
\end{framed}
This task has 197 test samples.

\paragraph{Clintox}
\begin{framed}
    \small\textit{Example 1:}\\
    \texttt{\textbf{Prompt}: Is <SMILES> COC[C@@H](NC(C)=O)C(=O)NCC1=CC=CC=C1 </SMILES> toxic?}\\
    \texttt{\textbf{Response}: No}
\end{framed}
This task has 144 test samples.

\paragraph{HIV}
\begin{framed}
    \small\textit{Example 1:}\\
    \texttt{\textbf{Prompt}: Can <SMILES> CC1=CN(C2C=CCCC2O)C(=O)NC1=O </SMILES> serve as an inhibitor of HIV replication?}\\
    \texttt{\textbf{Response}: No}
\end{framed}
This task has 4107 test samples.

\paragraph{SIDER}
\begin{framed}
    \small\textit{Example 1:}\\
    \texttt{\textbf{Prompt}: Are there any known side effects of <SMILES> CC1=CC(C)=C(NC(=O)CN(CC(=O\\)O)CC(=O)O)C(C)=C1Br </SMILES> affecting the heart?}\\
    \texttt{\textbf{Response}: No}
\end{framed}
This task has 2860 test samples.

\paragraph{MP$_\text{regression}$}
\begin{framed}
    \small\textit{Example 1:}\\
    \texttt{\textbf{Prompt}: You are a material scientist. Look at the chemical formula of the given crystalline material and predict its property. The output must be in a json format. For example: \{property\_name:predicted\_property\_value\}. Answer as precise as possible and in as few words as possible.chemical formula: KPrMnNbO6property name: volume}\\
    \texttt{\textbf{Response}: \{volume : 128.62405239223665\}}
\end{framed}
This task has 82241 test samples.

\paragraph{SNUMAT$_\text{regression}$}
\begin{framed}
    \small\textit{Example 1:}\\
    \texttt{\textbf{Prompt}: You are a material scientist. Look at the chemical formula of the given crystalline material and predict its property. The output must be in a json format. For example: \{property\_name:predicted\_property\_value\}. Answer as precise as possible and in as few words as possible.chemical formula: Na2Ti3Cl8property name: Band\_gap\_HSE}\\
    \texttt{\textbf{Response}: \{Band\_gap\_HSE : 1.583856\}}
\end{framed}
This task has 4152 test samples.

\paragraph{JARVISDFT}
\begin{framed}
    \small\textit{Example 1:}\\
    \texttt{\textbf{Prompt}: You are a material scientist. Look at the chemical formula of the given crystalline material and predict its property. The output must be in a json format. For example: \{property\_name:predicted\_property\_value\}. Answer as precise as possible and in as few words as possible.chemical formula: YReW2O8property name: formation\_energy\_peratom}\\
    \texttt{\textbf{Response}: \{formation\_energy\_peratom : -2.15925\}}
\end{framed}
This task has 54647 test samples.

\paragraph{JARVISQETB}
\begin{framed}
    \small\textit{Example 1:}\\
    \texttt{\textbf{Prompt}: You are a material scientist. Look at the chemical formula of the given crystalline material and predict its property. The output must be in a json format. For example: \{property\_name:predicted\_property\_value\}. Answer as precise as possible and in as few words as possible.chemical formula: BeMoRhproperty name: final\_energy}\\
    \texttt{\textbf{Response}: \{final\_energy : -169.1866374093\}}
\end{framed}
This task has 249596 test samples.

\paragraph{GNoME}
\begin{framed}
    \small\textit{Example 1:}\\
    \texttt{\textbf{Prompt}: You are a material scientist. Look at the chemical formula of the given crystalline material and predict its property. The output must be in a json format. For example: \{property\_name:predicted\_property\_value\}. Answer as precise as possible and in as few words as possible.chemical formula: Yb3Eu7(TlPd2)2property name: Density}\\
    \texttt{\textbf{Response}: \{Density : 8.2778\}}
\end{framed}
This task has 216455 test samples.

\paragraph{hMOF}
\begin{framed}
    \small\textit{Example 1:}\\
    \texttt{\textbf{Prompt}: You are a material scientist. Look at the chemical formula of the given crystalline material and predict its property. The output must be in a json format. For example: \{property\_name:predicted\_property\_value\}. Answer as precise as possible and in as few words as possible.chemical formula: Zn4C84N16H22O17property name: max\_co2\_adsp}\\
    \texttt{\textbf{Response}: \{max\_co2\_adsp : 8.95639\}}
\end{framed}
This task has 92925 test samples.

\paragraph{Cantor-HEA}
\begin{framed}
    \small\textit{Example 1:}\\
    \texttt{\textbf{Prompt}: You are a material scientist. Look at the chemical formula of the given crystalline material and predict its property. The output must be in a json format. For example: \{property\_name:predicted\_property\_value\}. Answer as precise as possible and in as few words as possible.chemical formula: Al2Co1Cr1Mn1Ni2Si1property name: e\_above\_hull}\\
    \texttt{\textbf{Response}: \{e\_above\_hull : 0.4014893125\}}
\end{framed}
This task has 33608 test samples.

\paragraph{QMOF}
\begin{framed}
    \small\textit{Example 1:}\\
    \texttt{\textbf{Prompt}: You are a material scientist. Look at the chemical formula of the given crystalline material and predict its property. The output must be in a json format. For example: \{property\_name:predicted\_property\_value\}. Answer as precise as possible and in as few words as possible.chemical formula: Zn12C48H88O80property name: lcd}\\
    \texttt{\textbf{Response}: \{lcd : 3.40559\}}
\end{framed}
This task has 3064 test samples.

\paragraph{OQMD}
\begin{framed}
    \small\textit{Example 1:}\\
    \texttt{\textbf{Prompt}: You are a material scientist. Look at the chemical formula of the given crystalline material and predict its property. The output must be in a json format. For example: \{property\_name:predicted\_property\_value\}. Answer as precise as possible and in as few words as possible.chemical formula: Sm2CuH6property name: e\_form}\\
    \texttt{\textbf{Response}: \{e\_form : -0.3328793676\}}
\end{framed}
This task has 192799 test samples.

\paragraph{OMDB}
\begin{framed}
    \small\textit{Example 1:}\\
    \texttt{\textbf{Prompt}: You are a material scientist. Look at the chemical formula of the given crystalline material and predict its property. The output must be in a json format. For example: \{property\_name:predicted\_property\_value\}. Answer as precise as possible and in as few words as possible.chemical formula: C4H5(CH)3(CH2)3C5H5NCH3C3NH4H3(CO)2C4NH4SO2property name: bandgap}\\
    \texttt{\textbf{Response}: \{bandgap : 2.886436\}}
\end{framed}
This task has 1213 test samples.

\paragraph{Catalytic Activity}
\begin{framed}
    \small\textit{Example 1:}\\
    \texttt{\textbf{Prompt}: Please evaluate the following protein sequence and provide an explanation of the enzyme's catalytic activity, including the chemical reaction it facilitates:  <protein>MVKKRIAVIGAGISGLGAIKCCLDEDLEPTCFERNDDIGGLWKF\\
    QKNASEKMPSIYRSVTINTSKEMMCFSDFPIPDHFPNYMHNSKLMDYFRMYAKRFSLLDYIRFKTTVRSVRKRPDFHIH\\
    GQWDVVVETDGKQESLVFDGVLVCSGHHTDPHLPLKSFPGIEKFEGCYFHSREYKSPEDYVGKRIIVVGIGNSGVDIAV\\
    ELGRVAKQVFLSTRRGSWILHRVWNNGYPMDSSFFTRFHSFLQKILTTEAVNKYLEKTLNSRFNHAHYGLQPQHRPLSQ\\
    HPTISDDLPNHIISGKVQVKPNVKEFTGTDVHFDDGTVEENIDVVIFATGYSISFPFLGDLIAVTDNEVSLYKLMFPPD\\
    LEKPTLAVIGLIQPLGIILPIAELQSRWAVRVFKGLSKLPSVKAMKADMDQRKKAMEKRYVKTARHTIQVDHIEYMDEI\\
    ASLAGVKPNLLLLFLSDPTLAMEVFFGPCTPYQYRLQGPGKWDGARRAILTQRERIIKPLKTRITSEKSRSAPGLFWIK\\
    MALFGLAFLVPSLTYFSYICQ</protein>}\\
    \texttt{\textbf{Response}: Upon reviewing the provided protein sequence, the corresponding enzyme's catalytic activity is identified as the following chemical reaction: H(+) + hexan-3-one + NADPH + O2 = ethyl butanoate + H2O +   NADP(+).}
\end{framed}
This task has 1601 test samples.

\paragraph{Physicochemical Prediction}
\begin{framed}
    \small\textit{Example 1:}\\
    \texttt{\textbf{Prompt}: Can you give me the energy difference between the HOMO and LUMO orbitals of this molecule? <SMILES> O=C1OCC2CC1C2 </SMILES>}\\
    \texttt{\textbf{Response}: 0.2756}    
\end{framed}
This task has 1999 test samples.

\paragraph{Fluorescence}
\begin{framed}
    \small\textit{Example 1:}\\
    \texttt{\textbf{Prompt}: <protein>SKGEELFTGVVPILVELDGDVNGHKFSVSGEGEGDATYGKLTLKFICTTGKLPVP\\
    WPTLVTTLSYGVQCFSRYPDHMKQHDFFKSAMPEGYVQERTIFFKDDGNYKTRAEVRFEGDTLVNRIELKGIDFKEDGNI\\
    LGHKLEYNYNSHNVYIMADKQKIGIKVNFRIRHNIEDGSVQLADHYQQNTPIGDGPVLLPDNHYLSTQSALSKDPNEKRD\\
    HMVLLEFVTAAGITHGMDELYK</protein> [Fluorescence Prediction Regression] What is the anticipated fitness of this green fluorescent protein mutant?}\\
    \texttt{\textbf{Response}: The fluorescence intensity for this GFP mutant is calculated as 3.67.}
\end{framed}
This task has 27217 test samples.

\paragraph{Stability}
\begin{framed}
    \small\textit{Example 1:}\\
    \texttt{\textbf{Prompt}: <protein>TTIKVNGQEYTVPLSPEQFAKAAKKRWPDYEVQIHGNTVKVTR</protein> What is the expected stability score for this sequence?}\\
    \texttt{\textbf{Response}: The stability score determined for this sequence is 0.53.}
\end{framed}
This task has nan test samples.

\paragraph{Thermostability}
\begin{framed}
    \small\textit{Example 1:}\\
    \texttt{\textbf{Prompt}: <protein>MNYARFITAASAARNPSPIRTMTDILSRGPKSMISLAGGLPNPNMFPFKTAVITV\\
    ENGKTIQFGEEMMKRALQYSPSAGIPELLSWLKQLQIKLHNPPTIHYPPSQGQMDLCVTSGSQQGLCKVFEMIINPGDNV\\
    LLDEPAYSGTLQSLHPLGCNIINVASDESGIVPDSLRDILSRWKPEDAKNPQKNTPKFLYTVPNGNNPTGNSLTSERKKE\\
    IYELARKYDFLIIEDDPYYFLQFNKFRVPTFLSMDVDGRVIRADSFSKIISSGLRIGFLTGPKPLIERVILHIQVSTLHP\\
    STFNQLMISQLLHEWGEEGFMAHVDRVIDFYSNQKDAILAAADKWLTGLAEWHVPAAGMFLWIKVKGINDVKELIEEKAV\\
    KMGVLMLPGNAFYVDSSAPSPYLRASFSSASPEQMDVAFQVLAQLIKESL</protein> The task evaluates protein thermostability using mass spectrometry to measure melting points in human cells. How is the thermostability of this protein assessed?}\\
    \texttt{\textbf{Response}: The task uses mass spectrometry to measure protein melting curves and predict thermostability in human cells. The thermostability measure recorded for this protein is 56.77}
\end{framed}
This task has 1336 test samples.

\paragraph{Enhancer Activity Prediction}
\begin{framed}
    \small\textit{Example 1:}\\
    \texttt{\textbf{Prompt}: <dna>TGTCGCTCCCATTTCGTCAAATGTTGCGTGCTAATTCGCTTGCCTTCTGGCGGCTATTT\\
    TTGTTTTGATATAATTTATATTTCTCGCTCTTTACTTATCTTTTTTGCTTTTTTTTTGTATTTTGTTTGCCGCTTACACT\\
    TCACATTTGCCTTCGGGGGCCGTCGCTTAAAAATAAACGCACACACACACTCGCGGCATTCACATATTTATTTTTGGTTT\\CTCCAACTTTGTACGCTCTGCGTTTTCTTC</dna> [Regression: Enhancer activity prediction] Please provide an enhancer activity prediction for this sequence.}\\
    \texttt{\textbf{Response}: The enhancer activity prediction yields: HK -0.68, Dev -1.26}
\end{framed}
This task has 41186 test samples.

\paragraph{APA Isoform Prediction}
\begin{framed}
    \small\textit{Example 1:}\\
    \texttt{\textbf{Prompt}: <rna>CTGCTTGGAGTTTACATTGAGGAAGCTGGTAACTGACCTTCAAAGCTAAAATATAAAAC\\
    TATTTGGGAAGTATGAAACGATGTCTCGTGATCTGGTGTACCCTTATCCCTGTGACGTTTGGCCTCTGACAATACTGGTA\\
    TAATTGTAAATAATGTCAAACTCCGTTTTCTAGCAAGTATTAAGGGA</rna> [Isoform prediction] What is the likelihood that this sequence ends in a proximal isoform?}\\
    \texttt{\textbf{Response}: The processing of 3' UTR regions in the sequence results in a proximal isoform proportion of 0.33.}
\end{framed}
This task has 49755 test samples.

\paragraph{Mean Ribosome Loading Prediction}
\begin{framed}
    \small\textit{Example 1:}\\
    \texttt{\textbf{Prompt}: <rna>TAATCGGTTCTAAATACGATTAGTG</rna> What ribosome loading rate is expected from the sequence?}\\
    \texttt{\textbf{Response}: The sequence impacts ribosome loading during translation initiation with an efficiency of 0.63}
\end{framed}
This task has 7600 test samples.

\paragraph{Programmable RNA Switches}
\begin{framed}
    \small\textit{Example 1:}\\
    \texttt{\textbf{Prompt}: <rna>GGGCTTGGGGAAAGTGGACAGAGATGATTATTAAACCAAACACACAAACGCACTAATAA\\
    TCATCTCTGTCCACTTTCCCCAAGAACAGAGGAGACTTGGGATGAGTGGACAGAACCTGGCGGC\\
    AGCGCAAAAGATGCGTAAAGGAGAA</rna> [Programmable RNA switches prediction] How does the sequence influence the ON and OFF states of the RNA switch?}\\
    \texttt{\textbf{Response}: The RNA switch with this sequence is predicted to have an ON state of 0.02, an OFF state of 0.47, and an ON/OFF ratio of 0.06}
\end{framed}
This task has 9154 test samples.

\paragraph{CRISPR On Target Prediction}
\begin{framed}
    \small\textit{Example 1:}\\
    \texttt{\textbf{Prompt}: <rna>TTCACTGCGCAGCCTGGCATTGG</rna> What gene disruption success rate is associated with the sequence?}\\
    \texttt{\textbf{Response}: The predicted gene disruption level for this sequence is 0.26.}
\end{framed}
This task has 416 test samples.

\paragraph{siRNA Efficiency Prediction}
\begin{framed}
    \small\textit{Example 1:}\\
    \texttt{\textbf{Prompt}: <sirna>a Af a c u Uf g a a u g a a Af c Gf a c u u c u c</sirna> <rna>ACGTACCTGCAGATCTACGGGGTCCTGGAGCACCCTGAAGACTCAAGACCAAAAGACTGTCAGGA\\
    AGGCAGAGTGCAGAGCAATCCACTGTCCAAGACCACACGACTTCGAGAACGGGGAATACTGGCCCCGGTCTCCCTACTAC\\
    AATGTGAGTGATGAGATCTCTTTCCACTGCTATGACGGTTACACTCTCCGGGGCTCTGCCAATCGCACCTGCCAAGTGAA\\
    TGGCCGATGGAGTGGGCAGACAGCGATCTGTGACAACGGAGCGGGGTACTGCTCCAACCCGGGCATCCCCATTGGCACAA\\
    GGAAGGTGGGCAGCCAGTACCGCCTTGAAGACAGCGTCACCTACCACTGCAGCCGGGGGCTTACCCTGCGTGGCTCCCAG\\
    CGGCGAACGTGTCAGGAAGGTGGCTCTTGGAGCGGGACGGAGCCTTCCTGCCAAGACTCCTTCATGTACGACACCCCTCA\\
    AGAGGTGGCCGAAGCTTTCCTGTCTTCCCTGACAGAGACCATAGAAGGAGTCGATGCTGAGGATGGGCACGGCCCAGGGG\\
    AACAACAGAAGCGGAAGATCGTCCTGGACCCTTCAGGCTCCATGAACATCTACCTGGTGCTAGATGGATCAGACAGCATT\\
    GGGGCCAGCAACTTCACAGGAGCCAAAAAGTGTCTAGTCAACTTAATTGAGAAGGTGGCAAGTTATGGTGTGAAGCCAAG\\
    ATATGGTCTAGTGACATATGCCACATACCCCAAAATTTGGGTCAAAGTGTCTGAAGCAGACAGCAGTAATGCAGACTGGG\\
    TCACGAAGCAGCTCA</rna> What is the likely mRNA remaining percentage after siRNA treatment?}\\
    \texttt{\textbf{Response}: The sequence leads to an mRNA remaining percentage of 28.00 after siRNA treatment.}
\end{framed}
This task has 6688 test samples.

\paragraph{MP$_\text{classification}$}
\begin{framed}
    \small\textit{Example 1:}\\
    \texttt{\textbf{Prompt}: You are a material scientist. Look at the chemical formula of the given crystalline material and predict its property. The output must be in a json format. For example: \{property\_name:predicted\_property\_value\}. Answer as precise as possible and in as few words as possible.chemical formula: KPrMnNbO6property name: is\_stable}\\
    \texttt{\textbf{Response}: \{is\_stable : False\}}
\end{framed}
This task has 20575 test samples.

\paragraph{SNUMAT$_\text{classification}$}
\begin{framed}
    \small\textit{Example 1:}\\
    \texttt{\textbf{Prompt}: You are a material scientist. Look at the chemical formula of the given crystalline material and predict its property. The output must be in a json format. For example: \{property\_name:predicted\_property\_value\}. Answer as precise as possible and in as few words as possible.chemical formula: PbSproperty name: SOC}\\
    \texttt{\textbf{Response}: \{SOC : True\}}
\end{framed}
This task has 4152 test samples.

\paragraph{Domain/Motif}
\begin{framed}
    \small\textit{Example 1:}\\
    \texttt{\textbf{Prompt}: Please conduct a domain/motif search on the given protein sequence and provide your findings. The sequence is:  <protein>MGTTTTHKFDRPLRLPPLPCPFPSEVNPYVEQVDKETLEWLIDSEMLDDAETVERYRQAKYGWLSARTYP\\
    YAEHHTLRLVSDWCVWLFAFDDAFCESDRRAAEIARALPQLYAVLEDLDVGSEVDDVFAKSLLEIKGRIAAYGDDEQLD\\
    RWRNVTKDYLFAQVWEAANREDEVVPSLEDYIFMRRRTGAMLTVFALIDVASGRSLSADEWRHPGMRAITESANDVVVW\\
    DNDLISYAKESNSGNSRNNLVNVLAEHRHYSRQEAMEEIGEMRNQAIADMVAVRPSLEALGSDAVLAYVRGLEFWISGSV\\
    DYSLTSSRYTDAWRTARQPSIR</protein>}\\
    \texttt{\textbf{Response}: Based on computational analysis, the provided sequence potentially contains the following protein domains or motifs: DDXXXE motif domains}
\end{framed}
This task has 1400 test samples.

\paragraph{Protein Function}
\begin{framed}
    \small\textit{Example 1:}\\
    \texttt{\textbf{Prompt}: Analyze the following amino acid sequence, and determine the function of the resulting protein, its subcellular localization, and any biological processes it may be part of:  <protein>MANHKSTQKSIRQDQKRNLINKSRKSNVKTFLKRVTLAINAGDKKVASEALSAAHSKLAKAANKG\\IYKLNTVSRKVSRLSRKIKQLEDKI</protein>}\\
    \texttt{\textbf{Response}: The protein with the amino acid sequence is expected to exhibit rRNA binding, structural constituent of ribosome, contributing to the translation. It can be typically found in the ribonucleoprotein complex, ribosome of the cell.}
\end{framed}
This task has 3494 test samples.

\paragraph{Solubility}
\begin{framed}
    \small\textit{Example 1:}\\
    \texttt{\textbf{Prompt}: <protein>MPLGLFSSGKAQVLCDEKIPGGKKKEPKQLSENKCKGVTLKLDHTRVTIEKQIAE\\
    GGFAIVYVASDRKNNKFALKRQFTKDNEKQLEACCREHSFLKQCIGHKNIVEFVDSYTNCLGNGIWECMLLTEYHQKNVL\\
    QLMNERISQNQYLTNDEILSIFTDLCEAVSFIHNRPQPIIHRDLKVENVLISSHKPPHYVLCDFGSATTQILSVEKYGVE\\
    YVKSEVERNTTMCYRSPEMIDFYSGLEIGLKSDIWALGVLLYRLCFFCVPFEESPLAIQSVNYQFPSVPNIPDEIKVLIY\\
    MLLDIDVNRRPSIYQTSVLAFEANHRKPLSEEIQNKKCTDAVPSLKSCIQLMRDGSNPRNKRDSSPRNPEAPPIQSSSKM\\
    ASLSQQVPSISNISMPSGSGTVETSVAPRLRPKATTVVPNVPSISPVPPVGLPHLRLPSKGSTDETDGSQVRKVPIDFHH\\
    RQSFSGEEQLKPAAEADSAGPLSCPLIKPTDLGFTDLDKPALPRDRAQTDGKRRLPHESDIIFQQQHRRNVSDTSQISRS\\
    AFKPYSSQQTTSKTSSQVVRSVEDMSQRQNGGSGEWNPFLVAPFSNNSISRKDGQESAFMMDDSHFGMVFDEIRRKEIPA\\
    ELDSETSSIDSRDPFGAAPFDQLTVSTSSSAQPVSLPPGSVHLHTSIAPSPLLRHVLTKAPPPVPRRTTSRGFQV</pro\\tein> Is this sequence associated with a soluble protein?}\\
    \texttt{\textbf{Response}: The protein predicted from this sequence is not soluble.}
\end{framed}
This task has 2001 test samples.

\paragraph{Function-EC}
\begin{framed}
    \small\textit{Example 1:}\\
    \texttt{\textbf{Prompt}: <protein>MGSSHHHHHHSQDLENLYFQGSMNIFEMLRIDEGLRLKIYKDTEGYYTIGIGHLL\\
    TKSPSLNAAKSELDKAIGRNTNGVITKDEAEKLFNQDVDAAVRGILRNAKLKPVYDSLDAVRRAALINMVFQMGETGVAG\\
    FTNSLRMLQQKRWDEAAVNLAKSRWYNQTPNRTKRVITTFRTGTWDAYKNLRKKLEQLYNRYKDPQDENKIGIDGIQQFC\\
    DDLALDPASISVLIIAWKFRAATQCEFSKQEFMDGMTELGCDSIEKLKAQIPKMEQELKEPGRFKDFYQFTFNFAKNPGQ\\
    KGLDLEMAIAYWNLVLNGRFKFLDLWNKFLLEHHKRSIPKDTWNLLLDFSTMIADDMSNYDEEGAWPVLIDDFVEFARPQ\\
    IAGTKSTTV</protein> [EC Number Prediction Multilabel Classification] What are the donor and acceptor molecules in this enzyme's reaction?}\\
    \texttt{\textbf{Response}: Donor molecules utilized by enzymes are identified by their EC numbers. Donor molecules it utilizes are identified by EC number EC3.2.1.17,EC3.2.1.-.}
\end{framed}
This task has 1919 test samples.

\paragraph{Antibody-Antigen}
\begin{framed}
    \small\textit{Example 1:}\\
    \texttt{\textbf{Prompt}: Ab-Ag interaction detection: Any binding between <protein>EVQLVESGANVVRPGGSLRLSCKASGFIFENFGFSWVRQAPGKGLQWVAGLNWNGGDTRYADSVKGRFRMS\\
    RDNSRNFVYLDMDKVGVDDTAFYYCARGTDYTIDDAGIHYQGSGTFWYFDLWGRGTLVSVSSXEIVLAQSPGTLSLSPGE\\
    RATLSCRASHNVHPKYFAWYQQKPGQSPRLLIYGGSTRAAGIPGKFSGSGSGTDFTLTISRVDPEDFAVYYCQQYGGSPY
    TFGQGTKVEIKX</protein> and <protein>MRVMGIQKNYPPLWRWGMIIFWIMMICNAKQMWVTVYYGVPVWR
    DASTTLFCASDAKAYDPEAHNAWATHACVPTDPNPQEMQLKNVTENFNMWKNDMVEQMHVDIISLWDQSLKPCVKLTPLC
    VTLNCTDVRNRTLNYTYNNSTSNISLVSSDMEGEIKNCSFNMTTELRDKNQKVYALFYRQDVIQNGNNNSSYYRLINCNT
    SAMTQACPKVSFNPIPIHYCAPAGFAILKCNDKEFNGTGPCKNVSTVQCTHGIKPVISTQLLLNGSIAEEEIVIRSENIT
    NNAKTIIVQLNEPVEINCTRPNNNTRKSVRIGPGQTFYATGDIIGNIRQAHCNVSRAKWNETLHKVAEQLRTYFKNKTII
    FTNHSGGDVELKTHMFNCGGEFFYCNTTALFNSTWDNNSTREDNNSTGNDTITLQCRIKQIVRMWQRSGQAMYAPPIQGV
    IKCKSNITGLLLTRDGGNNESTNETFRPGGGDMRDNWRSELYKYKVVKIEPLGVAPTHAKRRVVQREKRAVGLGAVFFGF
    LGAAGSTMGAASITLTVQARQLLSGIVQQQSNLLKAIEAQQHLLKLTVWGIKQLQARVLALERYLGDQQLLGIWGCSGKL
    ICTTNVPWNSSWSNKTYGDIWDNMTWLQWDKEISNYTNIIYALIEESQNQQEKNEQDLLALDKWASLWNWFDISKWLWYI
    KIFIMIVGGLIGLRIVFTIITIINRVRQGYSPLSFQTLTHHQREPDRPERIEEEGGEQDKDRSIRLVGGFLALAWDDLRN
    LCLFIYHRLRDFVLIAARAVSLLGHSSLKGLRLGWEALKYLWNLLAYWGKELKNSAVNLLDTVAIAVANWTDRVIEIGQR
    AGRAILNIPRRIRQGAERALV</protein>?}\\
    \texttt{\textbf{Response}: Yes. Based on their primary structures, a physical interaction appears plausible.}
\end{framed}
This task has 3301 test samples.

\paragraph{RNA-Protein}
\begin{framed}
    \small\textit{Example 1:}\\
    \texttt{\textbf{Prompt}: I need to know if <rna>TGCTTCGTGCTTGCTTGAAGTTAAACCCGGCGGGGGAACGA\\
    TTGTTCCCACTGAACGTTCCCACAGGAGAACCCTTGGAGACACTACCTTTCCCACAGAATTTCATTCAAGGAGCTCAAAG\\
    GCCAGGAAATCTTTTGAAAAGGTACCATCTACCTGGAGGGCATCATCCTGAGTGAGAGGACAAGCAAAAAGAGGAATGGA\\
    AGACTAGATACACAGCTAAATGGAAGGAGAACTTCTGGCCTAAATGCAATAAGGAAATCTATGTGGCTACCGAGAGCAAC\\
    AGGACACTCCTGACATTGTGTAACCCAAAGATAGAAACATGCAATTGTGCTACATGGCCCTCCAGGCTCTGGAAAAACCA\\
    CAATGGCAAAGCAGCTAATGTTAGAATGGTCAGAGCGTAAGCAGGCCCAGATTTTCTCATGTGCCTTCTACATCAGCTGC\\AGGGAAGTAAATAATACTAAACTTTGCACTTTTGCCCACCTCCTCTCCATGGACAATCC</rna> has any functional or structural interaction with <protein>HVLFEHAVGYALLALKEVEEISLLQPQVE\\
    ESVLNLGKFHSIVRLVAFCPFASSQVALENANAVSEGVVHEDLRLLLETHLPSKKKKVLLGVGDPKIGAAIQEELGYNCQ\\
    TGGVIAEILRGVRLHFHNLVKGLTDLSACKAQLGLGHSYSRAKVKFNVNRVDNMIIQSISLLDQLDKDINTFSMRVREWY\\
    GYHFPELVKIINDNATYCRLAQFIGNRRELNEDKLEKLEELTMDGAKAKAILDASRSSMGMDISAIDLINIESFSSRVVS\\
    LSEYRQSLHTYLRSKMSQVAPSLSALIGEAVGARLIAHAGSLTNLAKYPASTVQILGAEKALFRALKTRGNTPKYGLIFH\\
    STFIGRAAAKNKGRISRYLANKCSIASRIDCFSEVPTSVFGEKLREQVEERLSFYETGEIPRKNLDVMKEAMVQAEEAAA\\
    EITRKLEKQEKKRLKKKK</protein>.}\\
    \texttt{\textbf{Response}: The interaction between the RNA and protein is predicted to be positive.}
\end{framed}
This task has 4164 test samples.

\paragraph{Epigenetic Marks Predictions}
\begin{framed}
    \small\textit{Example 1:}\\
    \texttt{\textbf{Prompt}: <dna>CAACGGTACCCAATGGTTCCATCTCTAAGTCGTTAGGGCCAATCGTGGGCGGTATGCTA\\
    TTGTCACCTTCATCTTCGCTGTCTTCCTCACTCAACTTGAAAGTGGAAGAATCATTATTGTGCATTGTGAAACCTCCAGC\\
    ATTGGAAAATCTGTGTTTTAAGCTGGTGGACCTTGAAACCAAAAACAAGAAACATTGAAATGCAATGGCGGAGTGTAGTG\\
    AAGGAACAGCACCAAACACAATAGGAGACATGTGAAATCCTTTAGTGTTCAAATGAGTACCCAAATGGCTATCTACTCTT\\
    ATCAAACCGGCAGCAAACCCTTCCTGGGTGTAATTAACATGTTCGGTGTCATCAATACCGTACAGATGAGTAAACCATGG\\
    GGAGGCCATTGGAACAAGCAGATGGGTAAGGACCCCCGCTATATTTTGTAATCCTAGGGCAAAACTAAAGCATTTTAAAG\\
    TCCCCGGTGGTTGGAAGACGTATAAGTATACAGCTGTTAAA</dna> [Classification: epigenetic marks prediction] I need help with an Epigenetic Mark Prediction task. Here's the sequence:}\\
    \texttt{\textbf{Response}: EMP, or Epigenetic Marks Prediction, aims to identify epigenetic changes, successfully confirmed in yeast DNA sequences.}
\end{framed}
This task has 28741 test samples.

\paragraph{TF-Mouse}
\begin{framed}
    \small\textit{Example 1:}\\
    \texttt{\textbf{Prompt}: <dna>CTGCTGCTGCTGCTGGCTCTGGCTTTTGTTGTTGATGGTGACATCATCATTTAAAAAAA\\AACATATGCTGATCACTGTATGAAAGTTCAACAATGTTAGTG</dna>: I need a detailed review of transcription factor binding sites in this DNA sequence.}\\
    \texttt{\textbf{Response}: Upon examination, I can confirm that the mouse DNA sequence piece contains TF binding sites, which is crucial for gene expression.}
\end{framed}
This task has 10005 test samples.

\paragraph{Non-coding RNA Function Classification}
\begin{framed}
    \small\textit{Example 1:}\\
    \texttt{\textbf{Prompt}: <rna>AGCTACGGCCATATCACCCTGAACATGCCTGATCTCGTCTGATCTCAGAAGCTAAGCAC\\GGTAGGGCCTGGTTAGTACTTGGATGGAAGACCACCTGGGAATACCAGGTGCTGTAGGGG</rna> How would you classify this sequence within non-coding RNA classes?}\\
    \texttt{\textbf{Response}: The RNA sequence is categorized under the 5S\_rRNA family based on its structural characteristics.}
\end{framed}
This task has 2600 test samples.

\paragraph{Modification Prediction}
\begin{framed}
    \small\textit{Example 1:}\\
    \texttt{\textbf{Prompt}: <rna>TAAAGCACTTAACATACATAACTTACTGTGGCTCAGAGTAAGCGCTTAGTAAAAATTGG\\AGTTATTTTTCAATAGATTCATTATCTATAATAAAGGAGAAA</rna> What does the sequence indicate about its potential RNA modifications?Please choose from the following options:['Am', 'Cm', 'Gm', 'Um', 'm1A', 'm5C', 'm5U', 'm6A', 'm6Am', 'm7G', 'Psi', 'AtoI', 'none']}\\
    \texttt{\textbf{Response}: AtoI.The RNA modifications predicted for the sequence are AtoI.}
\end{framed}
This task has 1200 test samples.

\paragraph{Enhancer-Promoter Interaction Prediction}
\begin{framed}
    \small\textit{Example 1:}\\
    \texttt{\textbf{Prompt}: Analyze the interaction potential between <dna>CCTCCTGGGGCTGTTTACCAGAG\\
    CCTTAACTAGGTGCGACCTGTTGATAGCTGGTGTAGTCCCAGGAACCAGGATTTGGATGGAGGGTACAATGAGGCTCTGA
    GCTGCCTTCTCCCCTTCCAGGGCTGGGGGCAGCAGCCATGCCTACCTGGGGGGCCCGCCCCGCGTCCCCAGATCGCTTTG
    CGGTGTCTGCGGAGGCTGAGAACAAGGTTCGGGAACAGCAGCCCCATGTGGAGCGCATCTTCAGCGTGGGGGTGAGCGTC
    CTTCCGAAGGACTGTCCGGACAACCCCCACATCTGGCTGCAGCTGGAGGGCCCCAAGGAAAACGCCAGCAGAGCCAAGGT
    GAACGCCTTCTCTCCCCCATCCCTCCAGGCACCAAGGACGCTTTCCCCCAGGGCGGAGGAGAGCAGGGCCGAGGAGAGCA
    GGGAAGAGGAGGGGCCTGGGGAGGATCTCCACTTCTTGGTTGTCTCACTGGCAACCTCAGTGCCCCTGAGTTTTGAG</d
    na> and <dna>GGTCGTGCAGGTAAAACATCCCAAGCACCACTTCTTTCAGCAGGCGGCAAAGGAGCGGCCAGGGCCG
    AGGGCACTGGGACTGCAGCAGCCCCGACAAGGAGCCGTTCTCCATGAATTTAGTCACCAGAGCCGGCTTGGGATCTTGGT
    CCCAGTTCACCTTCTCGATAACCCCTTCTAGGCGCAGCACGAATTCGTTATCCAGACTTGCCATGGCCTTGACCTCCCTG
    GATATCGCCTTCCTACACTCCAGGAGAGAGCTGGAGTCGCACCGGGGTCGTGGGAAAATCCCTCCCTTCGCCATTCAGGC
    CCCAGAGCACAGTGGCTCCACCTTTTTGGCCAGATTGCGGCTGGGGTCAACCGGGGTCACTCACGAGTTTACGATCTTGA
    CCGCCACATCGTAGCCCCACTTCCTATGTTGCGCCCGGAACACTGTGCCGAACCCGCCTTTGCCGACGAGCTCCTGGTTC
    TCCAGTTCCTCGATGGACACCAAGGGGGCGGGG</dna>.}\\
    \texttt{\textbf{Response}: This DNA sequence suggests a high probability of promoter-enhancer interaction.}
\end{framed}
This task has 1762 test samples.

\paragraph{TF-Human}
\begin{framed}
    \small\textit{Example 1:}\\
    \texttt{\textbf{Prompt}: <dna>AGAAACCCTATGAATGCAGTCTTTGTAGGAAAGCTTTTTTTGAGAAGTCGGAGCTAATTAGACAT\\CTGAGAACTCATACAGGAGAAAAACCTTATGAATGC</dna>: Assist me in locating transcription factor binding sites in this DNA sequence.}\\
    \texttt{\textbf{Response}: My biological analysis indicates that TF binding sites are present in the given DNA fragment.}
\end{framed}
This task has 28741 test samples.

\paragraph{Promoter-300}
\begin{framed}
    \small\textit{Example 1:}\\
    \texttt{\textbf{Prompt}: <dna>TGTCGCGCCCTCCCGGGCTCGGCCACTCTGCCCTTCCCTGCCCCAGCGCAGGTGAGCTGGCGCTT\\
    GTCCTTAACCTTCTTCAACGACGAGGCGGGACGGAGGCTCGGACGCCTCCCCAGCCTCTGGGCTGGCGCAAGTCGACTGC
    GGTTCGTCCTCAGTCCACGATGAAGTCATAGGGCTGTGGGTCCGTGTGTCTGTGACGTCAGAAGATGGGTCCTGGGATGT
    CACTGCAGGCCCGGGCGGGGACCTGGGCTCACGGAAGGGCCGGGGGGACCCGCAGGGTAGAAAAAGGAAGTTGTC</dna
    >: Could you determine if this sequence contains a promoter?}\\
    \texttt{\textbf{Response}: The DNA fragment includes identifiable promoter characteristics.}
\end{framed}
This task has 5920 test samples.

\paragraph{Promoter-core}
\begin{framed}
    \small\textit{Example 1:}\\
    \texttt{\textbf{Prompt}: <dna>CTCTTGCCTTCCGGCCCGGCGCCCGATTTCCGCCTTCCGACCCAGCTGTGGGCTGCGCCCCACGC
    CAGCC</dna> Does this sequence function as a core promoter?}\\
    \texttt{\textbf{Response}: Yes, core promoter elements have been found in the given sequence.}
\end{framed}
This task has 5000 test samples.

\paragraph{Human PPI}
\begin{framed}
    \small\textit{Example 1:}\\
    \texttt{\textbf{Prompt}: Based on their primary structures, is an interaction between <protein>MSGEENPASKPTPVQDVQGDGRWMSLHHRFVADSKDKEPEVVFIGDSLVQLMHQCEIWRELFSPLHALNF\\
    GIGGDGTQHVLWRLENGELEHIRPKIVVVWVGTNNHGHTAEQVTGGIKAIVQLVNERQPQARVVVLGLLPRGQHPNPLR\\
    EKNRQVNELVRAALAGHPRAHFLDADPGFVHSDGTISHHDMYDYLHLSRLGYTPVCRALHSLLLRLLAQDQGQGAPLLE\\
    PAP</protein> and <protein>MWSPEREAEAPAGGDPAGLLPPEWEEDEERMSFLFSAFKRSREVNSTDWDSK\\
    MGFWAPLVLSHSRRQGVVRLRLRDLQEAFQRKGSVPLGLATVLQDLLRRGELQRESDFMASVDSSWISWGVGVFLLKPL\\
    KWTLSNMLGDNKVPAEEVLVAVELLKEKAEEVYRLYQNSPLSSHPVVALSELSTLCANSCPDERTFYLVLLQLQKEKRV\\
    TVLEQNGEKIVKFARGPRAKVSPVNDVDVGVYQLMQSEQLLSRKVESLSQEAERCKEEARRACRAGKKQLALRSL\\KAKQRTEKRIEALHAKLDTVQGILDRIYASQTDQMVFNAYQAGVGALKLSMKDVTVEKAESLVDQIQELCDTQDEVSQT\\LAGGVTNGLDFDSEELEKELDILLQDTTKEPLDLPDNPRNRHFTNSVPNPRISDAELEAELEKLSLSEGGLVPSSKSPK\\RQLEPTLKPL</protein> plausible in a human cell? Please answer 'Yes' if there is a stable physical association between two proteins and 'No' if not.}\\
    \texttt{\textbf{Response}: It is likely that these proteins interact. The answer is Yes.}
\end{framed}
This task has 237 test samples.

\paragraph{Yeast PPI}
\begin{framed}
    \small\textit{Example 1:}\\
    \texttt{\textbf{Prompt}: Based on the primary amino acid chains, is there evidence for a stable physical association between yeast protein A (<protein>MSNYPLHQACMENEFFKVQELLHSKPSLLLQKDQDGRIPLHWSVSFQAHEITSFLLSKMENVNLDDYPDD\\
    SGWTPFHIACSVGNLEVVKSLYDRPLKPDLNKITNQGVTCLHLAVGKKWFEVSQFLIENGASVRIKDKFNQIPLHRAASV\\
    GSLKLIELLCGLGKSAVNWQDKQGWTPLFHALAEGHGDAAVLLVEKYGAEYDLVDNKGAKAEDVALNEQVKKFFLNNV\\</protein>) and yeast protein B (<protein>METIDIQNRSFVVRWVKCGRGDVINYQIKPLKKSIEVG\\
    IYKKLKSSVDDHASAVHIAPDTKTLLDYTTKSLLHKGSSSNIEEHHRRSSQHSHSSSNGSDNKRKERSYSSLSISGIQQQ\\
    SQEIPLREKLSASGFTLVKRVGNVSGNTMVQGDLEVKDTDYYYAFILDNSSSKNAKKKILFNASVINGDNQSMISTRSTP\\
    PARPTALSRTSTQQDMLFRVGQGRYLQGYLLKKRRKRLQGFKKRFFTLDFRYGTLSYYLNDHNQTCRGEIVISLSSVSAN\\
    KKDKIIIIDSGMEVWVLKATTKENWQSWVDALQTCFDDQFEDKDTSTLEENPDILDDDKEVINKSSPQDHDHLTPTATTK\\
    SALSHRQHTQKDMDDIYVPLPSESYATFSMNLRLIQQRLEQCKKDSLSYKPTTLHQRSEGLNGTHSSSSVFTNNRVSSFN\\
    HSSSGMTSSDSLASEEVPSNKTYIEHALYNQLADLEVFVSRFVTQGEVLFKDHQILCKKAKDTRVSLTSYLSENDEFFDA\\
    EEEISRGVIILPDTEDDINNIVEETPLLGKSDQNEFTKEVQLSGSEQIASSSVESYTTNDENHSRKHLKNRHKNRRRGHP\\
    HHQKTKSAQSSTETFTSKDLFALSYPKSVTRRNDIPEAAASPPSLLSFLRKNVGKDLSSIAMPVTSNEPISILQLISETF\\
    EYAPLLTKATQRPDPITFVSAFAISFLSIYRDKTRTLRKPFNPLLAETFELIREDMGFRLISEKVSHRPPVFAFFAEHLD\\
    WECSYTVTPSQKFWGKSIELNNEGILRLKFKTTGELFEWTQPTTILKNLIAGERYMEPVNEFEVHSSKGDKSHILFDKAG\\
    MFSGRSEGFKVSIIPPPSSNRKKETLAGKWTQSLANETTHETIWEVGDLVSNPKKKYGFTKFTANLNEITEIEKGNLPPT\\
    DSRLRPDIRAYEEGNVDKAEEWKLKLEQLQRERRNKGQDVEPKYFEKVSKNEWKYITGPKSYWERRKKHDWSDISQLW\\</protein>)? Please answer 'Yes' if there is a stable physical association between two proteins and 'No' if not.}\\
    \texttt{\textbf{Response}: Interaction: Yes. The answer is Yes.}
\end{framed}
This task has 394 test samples.

\paragraph{Numerical EC (New-392)}
\begin{framed}
    \small\textit{Example 1:}\\
    \texttt{\textbf{Prompt}: : Return the EC number of the protein sequence. <protein>MSILEWFWNILCGMAQYLTFSKNLTNDNLVNIYVKSNVGGTISVNLDPKSDIKNVKELVAPKLGLEPDDVK
    IIFAGKELLDSTVIEVLDFFSDILHAVKVNKKIKNVIPDKPLCETLEELHQLNDQKNVESIEESNLKNEGKNKAHFFIYC
    ANPCKKINTGKLRVCCSECKHGAFTVDTDPQSWADVLDKNKITGVCNNVGCEGLYAKFYFKCASHPSQGENDTAVPLNLI
    KRNHKKIPCLACTDICDPVLVFSCDNRHVTCLECFKNYCGSRLKDRQFLSHPDFGYTLPCPAGCSNSFIEEVHHFRLLTD
    AQYEQYHRFATEEFILQAGGVLCPQPGCGQGILIDQNCNRVQCSCGYVFCGKCLEGFHLGECLNPTDVPFLSQNCDYPLD
    PEKLEKARWDEASSTVIKVLTKPCPKCRTSTERAGGCMHMICTRANCGFHWCWVCQGPWERDCMASHWFG</protein>}\\
    \texttt{\textbf{Response}: : 2 . 3 . 2 . 31}
\end{framed}
This task has 382 test samples.

\paragraph{NumericalEC (Price-149)}
\begin{framed}
    \small\textit{Example 1:}\\
    \texttt{\textbf{Prompt}: Return the EC number of the protein sequence. <protein>MAIPPYPDFRSAAF\\
    LRQHLRATMAFYDPVATDASGGQFHFFLDDGTVYNTHTRHLVSATRFVVTHAMLYRTTGEARYQVGMRHALEFLRTAFL\\
    DPATGGYAWLIDWQDGRATVQDTTRHCYGMAFVMLAYARAYEAGVPEARVWLAEAFDTAEQHFWQPAAGLYADEASPDW
    QLTSYRGQNANMHACEAMISAFRATGERRYIERAEQLAQGICQRQAALSDRTHAPAAEGWVWEHFHADWSVDWDYNRHD
    RSNIFRPWGYQVGHQTEWAKLLLQLDALLPADWHLPCAQRLFDTAVERGWDAEHGGLYYGMAPDGSICDDGKYHWVQA
    ESMAAAAVLAVRTGDARYWQWYDRIWAYCWAHFVDHEHGAWFRILHRDNRNTTREKSNAGKVDYHNMGACYDVLLWAL
    DAPGFSKESRSAALGRP</protein>}\\
    \texttt{\textbf{Response}: 5 . 3 . 1 . 7}
\end{framed}
This task has 149 test samples.

\paragraph{Subcellular Localization}
\begin{framed}
    \small\textit{Example 1:}\\
    \texttt{\textbf{Prompt}: Through sequence analysis, forecast the protein's subcellular positioning in eukaryotic cells. <protein>AGFPEQEPEPKFWNDWAQKTLDKALSLQTLNKNKAQN\\
    LILFLGDGMGVPTVTAARILKGQLRGQPGEEGQLEMDKFPFVALSKTYNTNAQVADSAGTATAYLCGVKANEGTVGVSAA
AVRSQANTTQGNEVTSILRWAKDAGKSIGIVTTTRVNHATPSAAYAHCVDRDWYSDNEMPADAVEAGCKDIARQLFENIP
DIDVIMGGGRKYMYPKNTTDVEYPGQPKHSGTRKDGRNLVKEWVDRNTEKKGHYVWNKKDLLSLNPTKVDYLLGLFEPAD
LPYDLERNKETDPSLSEMVEVAIKILRRNPNGFYLLVEGGRIDHGHHEGKDKQAIHEAVEMDRAIGRADLMTSTSDTLTV
VTADHSHLFSFGGYTPRGNEIFGLAAFISDVDQKPFTAILYGNGPGYKLVNGARENVSTVDYQDNSYLAQAAVPLSSETH
GGEDVAVFAKGPMAHLLHGVHEQNYIPHAMAYAACIGQNR</protein>}\\
    \texttt{\textbf{Response}: membrane}
\end{framed}
This task has 2772 test samples.

\paragraph{Keywords-CASPSimilarSeq}
\begin{framed}
    \small\textit{Example 1:}\\
    \texttt{\textbf{Prompt}: Analyze the provided protein sequence and predict the UniProtKB keywords associated with it. <protein>MAAADGGGPGGASVGTEEDGGGVGHRTVYLFDRREKESELG\\
    DRPLQVGERSDYAGFRACVCQTLGISPEEKFVITTTSRKEITCDNFDETVKDGVTLYLLQSVNQLLLTATKERIDFLPHY\\
    DTLVKSGMYEYYASEGQNPLPFALAELIDNSLSATSRNIGVRRIQIKLLFDETQGKPAVAVIDNGRGMTSKQLNNWAVYR\\
    LSKFTRQGDFESDHSGYVRPVPVPRSLNSDISYFGVGGKQAVFFVGQSARMISKPADSQDVHELVLSKEDFEKKEKNKEA
    IYSGYIRNRKPSDSVHITNDDERFLHHLIIEEKEKDSFTAVVITGVQPEHIQYLKNYFHLWTRQLAHIYHYYIHGPKGNE
    IRTSKEVEPFNNIDIEISMFEKGKVPKIVNLREIQDDMQTLYVNTAADSFEFKAHVEGDGVVEGIIRYHPFLYDRETYPD
    DPCFPSKLKDEDDEDDCFILEKAARGKRPIFECFWNGRLIPYTSVEDFDWCTPPKKRGLAPIECYNRISGALFTNDKFQV
    STNKLTFMDLELKLKDKNTLFTRILNGQEQRMKIDREFALWLKDCHEKYDKQIKFTLFKGVITRPDLPSKKQGPWATYAA
    IEWDGKIYKAGQLVKTIKTLPLFYGSIVRFFLYGDHDGEVYATGGEVQIAMEPQALYDEVRTVPIAKLDRTVAEKAVKKY
    VEDEMARLPDRLSVTWPEGDELLPNEVRPAGTPIGALRIEILNKKGEAMQKLPGTSHGGSKKLLVELKVILHSSSGNKEI
    ISHISQHGGKWPYWFKKMENIQKLGNYTLKLQVVLNESNADTYAGRPLPSKAIKFSVKEGKPEKFSFGLLDLPFRVGVPF
    NIPLEFQDEFGHTSQLVTDIQPVLEASGLSLHYEEITKGPNCVIRGVTAKGPVNSCQGKNYNLKVTLPGLKEDSQILKIR
    LLPGHPRRLKVKPDSEILVIENGTAFPFQVEVLDESDNITAQPKLIVHCKFSGAPNLPVYVVDCSSSGTSILTGSAIQVQ
    NIKKDQTLKARIEIPSCKDVAPVEKTIKLLPSSHVARLQIFSVEGQKAIQIKHQDEVNWIAGDIMHNLIFQMYDEGEREI
    NITSALAEKIKVNWTPEINKEHLLQGLLPDVQVPTSVKDMRYCQVSFQDDHVSLESAFTVRPLPDEPKHLKCEMKGGKTV
    QMGQELQGEVVIIITDQYGNQIQAFSPSSLSSLSIAGVGLDSSNLKTTFQENTQSISVRGIKFIPGPPGNKDLCFTWREF
    SDFIRVQLISGPPAKLLLIDWPELKESIPVINGRDLQNPIIVQLCDQWDNPAPVQHVKISLTKASNLKLMPSNQQHKTDE
    KGRANLGVFSVFAPRGEHTLQVKAIYNKSIIEGPIIKLMILPDPEKPVRLNVKYDKDASFLAGGLFTDFMISVISEDDSI
    IKNINPARISMKMWKLSTSGNRPPANAETFSCNKIKDNDKEDGCFYFRDKVIPNKVGTYCIQFGFMMDKTNILNSEQVIV
    EVLPNQPVKLVPKIKPPTPAVSNVRSVASRTLVRDLHLSITDDYDNHTGIDLVGTIIATIKGSNEEDTDTPLFIGKVRTL
    EFPFVNGSAEIMSLVLAESSPGRDSTEYFIVFEPRLPLLSRTLEPYILPFMFYNDVKKQQQMAALTKEKDQLSQSIVMYK
    SLFEASQQLLNEMKCQVEEARLKEAQLRNELKIHNIDIPTTQQVPHIEALLKRKLSEQEELKKKPRRSCTLPNYTKGSGD
    VLGKIAHLAQIEDDRAAMVISWHLASDMDCVVTLTTDAARRIYDETQGRQQVLPLDSIYKKTLPDWKRSLPHFRNGKLYF
    KPIGDPVFARDLLTFPDNVEHCETVFGMLLGDTIILDNLDAANHYRKEVVKITHCPTLLTRDGDRIRSNGKFGGLQNKAP
    PMDKLRGMVFGAPVPKQCLILGEQIDLLQQYRSAVCKLDSVNKDLNSQLEYLRTPDMRKKKQELDEHEKNLKLIEEKLGM
    TPIRKCNDSLRHSPKVETTDCPVPPKRMRREATRQNRIITKTDV</protein>}\\
    \texttt{\textbf{Response}: 3D-structure ; Acetylation ; Alternative splicing ; Chromatin regulator ; Chromosome ; Direct protein sequencing ; Disease variant ; DNA damage ; DNA repair ; DNA-binding ; Hydrolase ; Hypogonadotropic hypogonadism ; Isopeptide bond ; Kallmann syndrome ; Microphthalmia ; Phosphoprotein ; Reference proteome ; Ubl conjugation}
\end{framed}
This task has 184 test samples.

\paragraph{Keywords-IDFilterSeq}
\begin{framed}
    \small\textit{Example 1:}\\
    \texttt{\textbf{Prompt}: Please list the UniProtKB keywords associated with the provided protein sequence. <protein>MTSVAKVYYSQTTQTESRPLMGPGIRRRRVLTKDGRSNVRMEHIADKRFLYL
    KDLWTTFIDMQWRYKLLLFSATFAGTWFLFGVVWYLVAVAHGDLLELDPPANHTPCVVQVHTLTGAFLFSLESQTTIGYG
    FRYISEECPLAIVLLIAQLVLTTILEIFITGTFLAKIARPKKRAETIRFSQHAVVASHNGKPCLMIRVANMRKSLLIGCQ
    VTGKLLQTHQTKEGENIRLNQVNVTFQVDTASDSPFLILPLTFYHVVDETSPLKDLPLRSGEGDFELVLILSGTVESTSA
    TCQVRTSYLPEEILWGYEFTPAISLSASGKYIADFSLFDQVVKVASPSGLRDSTVRYGDPEKLKLEESLREQAEKEGSAL
    SVRISNV</protein>}\\
    \texttt{\textbf{Response}: ATP-binding ; Cell membrane ; Deafness ; Disease variant ; Epilepsy ; Intellectual disability ; Ion channel ; Ion transport ; Membrane ; Nucleotide-binding ; Potassium ; Potassium transport ; Reference proteome ; Transmembrane ; Transmembrane helix ; Transport ; Voltage-gated channel}
\end{framed}
This task has 1112 test samples.

\paragraph{Keywords-UniProtSeq}
\begin{framed}
    \small\textit{Example 1:}\\
    \texttt{\textbf{Prompt}: Analyze the given protein sequence and generate the relevant UniProtKB keywords. <protein>MWLKLFFLLLYFLVLFVLARFFEAIVWYETGIFATQLVDPVALSFKKLKT
    ILECRGLGYSGLPEKKDVRELVEKSGDLMEGELYSALKEEEASESVSSTNFSGEMHFYELVEDTKDGIWLVQVIANDRSP
    LVGKIHWEKMVKKVSRFGIRTGTFNCSSDPRYCRRRGWVRSTLIMSVPQTSTSKGKVMLKEYSGRKIEVEHIFKWITAHA
    ASRIKTIYNAEHLKEEWNKSDQYWLKIYLFANLDQPPAFFSALSIKFTGRVEFIFVNVENWDNKSYMTDIGIYNMPSYIL
    RTPEGIYRYGNHTGEFISLQAMDSFLRSLQPEVNDLFVLSLVLVNLMAWMDLFITQGATIKRFVVLISTLGTYNSLLIIS
    WLPVLGFLQLPYLDSFYEYSLKLLRYSNTTTLASWVRADWMFYSSHPALFLSTYLGHGLLIDYFEKKRRRNNNNDEVNAN
    NLEWLSSLWDWYTSYLFHPIASFQNFPVESDWDEDPDLFLERLAFPDLWLHPLIPTDYIKNLPMWRFKCLGVQSEEEMSE
    GSQDTENDSESENTDTLSSEKEVFEDKQSVLHNSPGTASHCDAEACSCANKYCQTSPCERKGRSYGSYNTNEDMEPDWLT
    WPADMLHCTECVVCLENFENGCLLMGLPCGHVFHQNCIVMWLAGGRHCCPVCRWPSYKKKQPYAQHQPLSNDVPS</pro
    tein>}\\
    \texttt{\textbf{Response}: Endoplasmic reticulum ; Membrane ; Metal-binding ; Reference proteome ; Transferase ; Transmembrane ; Transmembrane helix ; Ubl conjugation pathway ; Zinc ; Zinc-finger}
\end{framed}
This task has 4562 test samples.

\paragraph{Fold Type}
\begin{framed}
    \small\textit{Example 1:}\\
    \texttt{\textbf{Prompt}: Please predict the folding type of the protein sequence, the prediction range is from 0 to 1194. <protein>MLTAEEKAAVTGFWGKVDVDVVGAQALGRLLVVY
    PWTQRFFQHFGNLSSAGAVMNNPKVKAHGKRVLDAFTQGLKHLDDLKGAFAQLSGLHCNKLHVNPQNFRLLGNVLALVV
    ARNFGGQFTPNVQALFQKVVAGVANALAHKYH</protein>}\\
    \texttt{\textbf{Response}: 0}
\end{framed}
This task has 3243 test samples.

\paragraph{GO-CASPSimilarSeq}
\begin{framed}
    \small\textit{Example 1:}\\
    \texttt{\textbf{Prompt}: Predict the GO term based on the protein sequence. <protein>MAAADGGGPGGASVGTEEDGGGVGHRTVYLFDRREKESELGDRPLQVGERSDYAGFRACVCQTLGISPEEK\\
    FVITTTSRKEITCDNFDETVKDGVTLYLLQSVNQLLLTATKERIDFLPHYDTLVKSGMYEYYASEGQNPLPFALAELIDN\\
    SLSATSRNIGVRRIQIKLLFDETQGKPAVAVIDNGRGMTSKQLNNWAVYRLSKFTRQGDFESDHSGYVRPVPVPRSLNSD
    ISYFGVGGKQAVFFVGQSARMISKPADSQDVHELVLSKEDFEKKEKNKEAIYSGYIRNRKPSDSVHITNDDERFLHHLII
    EEKEKDSFTAVVITGVQPEHIQYLKNYFHLWTRQLAHIYHYYIHGPKGNEIRTSKEVEPFNNIDIEISMFEKGKVPKIVN
    LREIQDDMQTLYVNTAADSFEFKAHVEGDGVVEGIIRYHPFLYDRETYPDDPCFPSKLKDEDDEDDCFILEKAARGKRPI
    FECFWNGRLIPYTSVEDFDWCTPPKKRGLAPIECYNRISGALFTNDKFQVSTNKLTFMDLELKLKDKNTLFTRILNGQEQ
    RMKIDREFALWLKDCHEKYDKQIKFTLFKGVITRPDLPSKKQGPWATYAAIEWDGKIYKAGQLVKTIKTLPLFYGSIVRF
    FLYGDHDGEVYATGGEVQIAMEPQALYDEVRTVPIAKLDRTVAEKAVKKYVEDEMARLPDRLSVTWPEGDELLPNEVRPA
    GTPIGALRIEILNKKGEAMQKLPGTSHGGSKKLLVELKVILHSSSGNKEIISHISQHGGKWPYWFKKMENIQKLGNYTLK
    LQVVLNESNADTYAGRPLPSKAIKFSVKEGKPEKFSFGLLDLPFRVGVPF... (We hide the remaining seq
    uences for readability)</protein>}\\
    \texttt{\textbf{Response}: Barr body ; site of double-strand break ; ATP binding ; ATP hydrolysis activity ; DNA binding ; protein homodimerization activity ; chromosome organization ; dosage compensation by inactivation of X chromosome ; double-strand break repair ; inactivation of X chromosome by heterochromatin formation ; negative regulation of double-strand break repair via homologous recombination ; nose development ; positive regulation of DNA repair ; positive regulation of double-strand break repair via nonhomologous end joining}
\end{framed}
This task has 184 test samples.

\paragraph{GO-IDFilterSeq}
\begin{framed}
    \small\textit{Example 1:}\\
    \texttt{\textbf{Prompt}: Given a protein sequence, predict the corresponding Gene Ontology term that describes its molecular function, biological process, and cellular component. <protein>MTSVAKVYYSQTTQTESRPLMGPGIRRRRVLTKDGRSNVRMEHIADKRFLYLKDLWTTF
    IDMQWRYKLLLFSATFAGTWFLFGVVWYLVAVAHGDLLELDPPANHTPCVVQVHTLTGAFLFSLESQTTIGYGFRYISEE
    CPLAIVLLIAQLVLTTILEIFITGTFLAKIARPKKRAETIRFSQHAVVASHNGKPCLMIRVANMRKSLLIGCQVTGKLLQ
    THQTKEGENIRLNQVNVTFQVDTASDSPFLILPLTFYHVVDETSPLKDLPLRSGEGDFELVLILSGTVESTSATCQVRTS
    YLPEEILWGYEFTPAISLSASGKYIADFSLFDQVVKVASPSGLRDSTVRYGDPEKLKLEESLREQAEKEGSALSVRISNV
    </protein>}\\
    \texttt{\textbf{Response}: astrocyte projection ; basolateral plasma membrane ; cell body ; ciliary base ; plasma membrane ; presynapse ; ATP binding ; ATP-activated inward rectifier potassium channel activity ; inward rectifier potassium channel activity ; adult walking behavior ; cellular response to potassium ion ; central nervous system myelination ; glutamate reuptake ; non-motile cilium assembly ; potassium ion homeostasis ; potassium ion import across plasma membrane ; potassium ion transport ; regulation of long-term neuronal synaptic plasticity ; regulation of monoatomic ion transmembrane transport ; regulation of resting membrane potential ; visual perception}
\end{framed}
This task has 1112 test samples.

\paragraph{GO-UniProtSeq}
\begin{framed}
    \small\textit{Example 1:}\\
    \texttt{\textbf{Prompt}: For a given protein sequence, identify the appropriate Gene Ontology term that corresponds to its molecular function, biological process, and cellular component. <protein>MWLKLFFLLLYFLVLFVLARFFEAIVWYETGIFATQLVDPVALSFKKLKT
    ILECRGLGYSGLPEKKDVRELVEKSGDLMEGELYSALKEEEASESVSSTNFSGEMHFYELVEDTKDGIWLVQVIANDRSP
    LVGKIHWEKMVKKVSRFGIRTGTFNCSSDPRYCRRRGWVRSTLIMSVPQTSTSKGKVMLKEYSGRKIEVEHIFKWITAHA
    ASRIKTIYNAEHLKEEWNKSDQYWLKIYLFANLDQPPAFFSALSIKFTGRVEFIFVNVENWDNKSYMTDIGIYNMPSYIL
    RTPEGIYRYGNHTGEFISLQAMDSFLRSLQPEVNDLFVLSLVLVNLMAWMDLFITQGATIKRFVVLISTLGTYNSLLIIS
    WLPVLGFLQLPYLDSFYEYSLKLLRYSNTTTLASWVRADWMFYSSHPALFLSTYLGHGLLIDYFEKKRRRNNNNDEVNAN
    NLEWLSSLWDWYTSYLFHPIASFQNFPVESDWDEDPDLFLERLAFPDLWLHPLIPTDYIKNLPMWRFKCLGVQSEEEMSE
    GSQDTENDSESENTDTLSSEKEVFEDKQSVLHNSPGTASHCDAEACSCANKYCQTSPCERKGRSYGSYNTNEDMEPDWLT
    WPADMLHCTECVVCLENFENGCLLMGLPCGHVFHQNCIVMWLAGGRHCCPVCRWPSYKKKQPYAQHQPLSNDVPS</pro
    tein>}\\
    \texttt{\textbf{Response}: endoplasmic reticulum ; endoplasmic reticulum membrane ; endoplasmic reticulum quality control compartment ; metal ion binding ; ubiquitin protein ligase activity ; ubiquitin-protein transferase activity ; central nervous system development ; endoplasmic reticulum mannose trimming ; protein ubiquitination ; ubiquitin-dependent ERAD pathway}
\end{framed}
This task has 4562 test samples.

\paragraph{gSymbol2Tissue}
\begin{framed}
    \small\textit{Example 1:}\\
    \texttt{\textbf{Prompt}: According to the relevant data reported in the scientific literature, the Gene symbol of a given protein can be used to predict which tissues the protein is expressed in. TMEM198}\\
    \texttt{\textbf{Response}: adrenal gland , appendix , breast , bronchus , cerebral cortex , colon , fallopian tube , gallbladder , kidney , nasopharynx , prostate , rectum , salivary gland , seminal vesicle , skeletal muscle , small intestine , stomach , urinary bladder}
\end{framed}
This task has 2181 test samples.

\paragraph{gSymbol2Cancer}
\begin{framed}
    \small\textit{Example 1:}\\
    \texttt{\textbf{Prompt}: Get the cancer name that is associated with the given gene symbol. MLLT1}\\
    \texttt{\textbf{Response}: AL}
\end{framed}
This task has 148 test samples.

\paragraph{gName2Cancer}
\begin{framed}
    \small\textit{Example 1:}\\
    \texttt{\textbf{Prompt}: Return the cancer name associated with the given gene name. myeloid/lymphoid or mixed-lineage leukemia (trithorax homolog, Drosophila); translocated to, 1 (ENL)}\\
    \texttt{\textbf{Response}: AL}
\end{framed}
This task has 148 test samples.

\paragraph{S2I}
\begin{framed}
    \small\textit{Example 1:}\\
    \texttt{\textbf{Prompt}: Translate the given SMILES formula of a molecule <SMILES> CCC(C)C1CNCCCNC1 </SMILES> into its IUPAC name.}\\
    \texttt{\textbf{Response}: <IUPAC> 3-butan-2-yl-1,5-diazocane </IUPAC>}
\end{framed}
This task has 2993 test samples.

\paragraph{I2S}
\begin{framed}
    \small\textit{Example 1:}\\
    \texttt{\textbf{Prompt}: Could you provide the SMILES for <IUPAC> 4-ethyl-4-methyloxolan-2-one </IUPAC> ?}\\
    \texttt{\textbf{Response}: Of course. It's <SMILES> CCC1(C)COC(=O)C1 </SMILES> .}
\end{framed}
This task has 2993 test samples.

\paragraph{S2F}
\begin{framed}
    \small\textit{Example 1:}\\
    \texttt{\textbf{Prompt}: Given the SMILES representation <SMILES> S=P1(N(CCCl)CCCl)NCCCO1 </SMILES>, what would be its molecular formula?}\\
    \texttt{\textbf{Response}: It is C7H15Cl2N2OPS .}
\end{framed}
This task has 2993 test samples.

\paragraph{I2F}
\begin{framed}
    \small\textit{Example 1:}\\
    \texttt{\textbf{Prompt}: What is the molecular formula of the compound with this IUPAC name <IUPAC> 2,5-diphenyl-1,3-oxazole </IUPAC> ?}\\
    \texttt{\textbf{Response}: C15H11NO}
\end{framed}
This task has 2993 test samples.

\paragraph{Molecular Description}
\begin{framed}
    \small\textit{Example 1:}\\
    \texttt{\textbf{Prompt}: Could you provide a description of this molecule? <SMILES> C/C=C/C1=CC2=C(C=C1)OC(=C2C)C3=CC=C(C=C3)OC4=C(C=CC(=C4)C5=C(C6=C(O5)C=CC(=C6)\\/C=C/C)C)O </SMILES>}\\
    \texttt{\textbf{Response}: The molecule is a natural product found in Piper aequale with data available.}
\end{framed}
This task has 1000 test samples.

\paragraph{Molecule Captioning}
\begin{framed}
    \small\textit{Example 1:}\\
    \texttt{\textbf{Prompt}: Describe this molecule: <SMILES> CCOC(=O)C1=CN=CN1[C@H](C)C1=CC=CC=C1 </SMILES>}\\
    \texttt{\textbf{Response}: The molecule is an imidazole derivative with short-acting sedative, hypnotic, and general anesthetic properties. Etomidate appears to have gamma-aminobutyric acid (GABA) like effects, mediated through GABA-A receptor. The action enhances the inhibitory effect of GABA on the central nervous system by causing chloride channel opening events which leads to membrane hyperpolarization.}
\end{framed}
This task has 2538 test samples.

\paragraph{Protein Functional Translation-Molins}
\begin{framed}
    \small\textit{Example 1:}\\
    \texttt{\textbf{Prompt}: Examine the given protein sequence and share a brief overview of its attributes:  <protein>MKIVLASNNQGKLAELKAMLAPLGVQLLRQAELGIPEAAEPFRTFVENALAKARHA\\
    SALSGLPALADDAGLCVEAFGGLPGVDTAFYAVQFGYAKGDANNVKALLAQMAGITDRRAALVSTLVAVRSA\\
    EDPEPLIACGRVAGEVALEPMGSNGFGFDPVMFIPEFGQTFAQLPVEVKNANSHRGKATRQMMALMRER
    WIT\\</protein>}\\
    \texttt{\textbf{Response}: A concise description of the protein with the specified amino acid sequence includes: Pyrophosphatase that catalyzes the hydrolysis of nucleoside triphosphates to their monophosphate derivatives, with a high preference for the non-canonical purine nucleotides XTP (xanthosine triphosphate), dITP (deoxyinosine triphosphate) and ITP. Seems to function as a house-cleaning enzyme that removes non-canonical purine nucleotides from the nucleotide pool, thus preventing their incorporation into DNA/RNA and avoiding chromosomal lesions.}
\end{framed}
This task has 2633 test samples.

\paragraph{Protein Functional Translation-CASPSimilarSeq}
\begin{framed}
    \small\textit{Example 1:}\\
    \texttt{\textbf{Prompt}: Predict the general functional description of the protein sequence. 
    <protein>MAAADGGGPGGASVGTEEDGGGVGHRTVYLFDRREKESELGDRPLQVGERSDYAGFRACVCQTLGISPEEK
    FVITTTSRKEITCDNFDETVKDGVTLYLLQSVNQLLLTATKERIDFLPHYDTLVKSGMYEYYASEGQNPLPFALAELIDN
    SLSATSRNIGVRRIQIKLLFDETQGKPAVAVIDNGRGMTSKQLNNWAVYRLSKFTRQGDFESDHSGYVRPVPVPRSLNSD
    ISYFGVGGKQAVFFVGQSARMISKPADSQDVHELVLSKEDFEKKEKNKEAIYSGYIRNRKPSDSVHITNDDERFLHHLII
    EEKEKDSFTAVVITGVQPEHIQYLKNYFHLWTRQLAHIYHYYIHGPKGNEIRTSKEVEPFNNIDIEISMFEKGKVPKIVN
    LREIQDDMQTLYVNTAADSFEFKAHVEGDGVVEGIIRYHPFLYDRETYPDDPCFPSKLKDEDDEDDCFILEKAARGKRPI
    FECFWNGRLIPYTSVEDFDWCTPPKKRGLAPIECYNRISGALFTNDKFQVSTNKLTFMDLELKLKDKNTLFTRILNGQEQ
    RMKIDREFALWLKDCHEKYDKQIKFTLFKGVITRPDLPSKKQGPWATYAAIEWDGKIYKAGQLVKTIKTLPLFYGSIVRF
    FLYGDHDGEVYATGGEVQIAMEPQALYDEVRTVPIAKLDRTVAEKAVKKYVEDEMARLPDRLSVTWPEGDELLPNEVRPA
    GTPIGALRIEILNKKGEAMQKLPGTSHGGSKKLLVELKVILHSSSGNKEIISHISQHGGKWPYWFKKMENIQKLGNYTLK
    LQVVLNESNADTYAGRPLPSKAIKFSVKEGKPEKFSFGLLDLPFRVGVPFNIPLEFQDEFGHTSQLVTDIQPVLEASGLS
    LHYEEITKGPNCVIRGVTAKGPVNSCQGKNYNLKVTLPGLKEDSQILKIRLLPGHPRRLKVKPDSEILVIENGTAFPFQV
    EVLDESDNITAQPKLIVHCKFSGAPNLPVYVVDCSSSGTSILTGSAIQVQNIKKDQTLKARIEIPSCKDVAPVEKTIKLL
    PSSHVARLQIFSVEGQKAIQIKHQDEVNWIAGDIMHNLIFQMYDEGEREINITSALAEKIKVNWTPEINKEHLLQGLLPD
    VQVPTSVKDMRYCQVSFQDDHVSLESAFTVRPLPDEPKHLKCEMKGGKTVQMGQELQGEVVIIITDQYGNQIQAFSPSSL
    SSLSIAGVGLDSSNLKTTFQENTQSISVRGIKFIPGPPGNKDLCFTWREFSDFIRVQLISGPPAKLLLIDWPELKESIPV
    INGRDLQNPIIVQLCDQWDNPAPVQHVKISLTKASNLKLMPSNQQHKTDEKGRANLGVFSVFAPRGEHTLQVKAIYNKSI
    IEGPIIKLMILPDPEKPVRLNVKYDKDASFLAGGLFTDFMISVISEDDSIIKNINPARISMKMWKLSTSGNRPPANAETF
    SCNKIKDNDKEDGCFYFRDKVIPNKVGTYCIQFGFMMDKTNILNSEQVIVEVLPNQPVKLVPKIKPPTPAVSNVRSVASR
    TLVRDLHLSITDDYDNHTGIDLVGTIIATIKGSNEEDTDTPLFIGKVRTLEFPFVNGSAEIMSLVLAESSPGRDSTEYFI
    VFEPRLPLLSRTLEPYILPFMFYNDVKKQQQMAALTKEKDQLSQSIVMYKSLFEASQQLLNEMKCQVEEARLKEAQLRNE
    LKIHNIDIPTTQQVPHIEALLKRKLSEQEELKKKPRRSCTLPNYTKGSGDVLGKIAHLAQIEDDRAAMVISWHLASDMDC
    VVTLTTDAARRIYDETQGRQQVLPLDSIYKKTLPDWKRSLPHFRNGKLYFKPIGDPVFARDLLTFPDNVEHCETVFGMLL
    GDTIILDNLDAANHYRKEVVKITHCPTLLTRDGDRIRSNGKFGGLQNKAPPMDKLRGMVFGAPVPKQCLILGEQIDLLQQ
    YRSAVCKLDSVNKDLNSQLEYLRTPDMRKKKQELDEHEKNLKLIEEKLGMTPIRKCNDSLRHSPKVETTDCPVPPKRMRR
    EATRQNRIITKTDV</protein>}\\
    \texttt{\textbf{Response}: Non-canonical member of the structural maintenance of chromosomes (SMC) protein family that plays a key role in epigenetic silencing by regulating chromatin architecture (By similarity). Promotes heterochromatin formation in both autosomes and chromosome X, probably by mediating the merge of chromatin compartments (By similarity). Plays a key role in chromosome X inactivation in females by promoting the spreading of heterochromatin . Recruited to inactivated chromosome X by Xist RNA and acts by mediating the merge of chromatin compartments: promotes random chromatin interactions that span the boundaries of existing structures, leading to create a compartment-less architecture typical of inactivated chromosome X (By similarity). Required to facilitate Xist RNA spreading (By similarity). Also required for silencing of a subset of clustered autosomal loci in somatic cells, such as the DUX4 locus . Has ATPase activity; may participate in structural manipulation of chromatin in an ATP-dependent manner as part of its role in gene expression regulation . Also plays a role in DNA repair: localizes to sites of DNA double-strand breaks in response to DNA damage to promote the repair of DNA double-strand breaks . Acts by promoting non-homologous end joining (NHEJ) and inhibiting homologous recombination (HR) repair .}
\end{framed}
This task has 184 test samples.

\paragraph{Protein Functional Translation-IDFilterSeq}
\begin{framed}
    \small\textit{Example 1:}\\
    \texttt{\textbf{Prompt}: Predict the functional description of the protein sequence below. <protein>MTSVAKVYYSQTTQTESRPLMGPGIRRRRVLTKDGRSNVRMEHIADKRFLYLKDLWTTFIDMQWRYKLLLF
    SATFAGTWFLFGVVWYLVAVAHGDLLELDPPANHTPCVVQVHTLTGAFLFSLESQTTIGYGFRYISEECPLAIVLLIAQL
    VLTTILEIFITGTFLAKIARPKKRAETIRFSQHAVVASHNGKPCLMIRVANMRKSLLIGCQVTGKLLQTHQTKEGENIRL
    NQVNVTFQVDTASDSPFLILPLTFYHVVDETSPLKDLPLRSGEGDFELVLILSGTVESTSATCQVRTSYLPEEILWGYEF
    TPAISLSASGKYIADFSLFDQVVKVASPSGLRDSTVRYGDPEKLKLEESLREQAEKEGSALSVRISNV</protein>}\\
    \texttt{\textbf{Response}: May be responsible for potassium buffering action of glial cells in the brain. Inward rectifier potassium channels are characterized by a greater tendency to allow potassium to flow into the cell rather than out of it. Their voltage dependence is regulated by the concentration of extracellular potassium; as external potassium is raised, the voltage range of the channel opening shifts to more positive voltages. The inward rectification is mainly due to the blockage of outward current by internal magnesium. Can be blocked by extracellular barium and cesium (By similarity). In the kidney, together with KCNJ16, mediates basolateral K(+) recycling in distal tubules; this process is critical for Na(+) reabsorption at the tubules.}
\end{framed}
This task has 1112 test samples.

\paragraph{Protein Functional Translation-UniProtSeq}
\begin{framed}
    \small\textit{Example 1:}\\
    \texttt{\textbf{Prompt}: Predict the general functional description of the protein sequence. 
    <protein>MWLKLFFLLLYFLVLFVLARFFEAIVWYETGIFATQLVDPVALSFKKLKTILECRGLGYSGLPEKKDVREL
    VEKSGDLMEGELYSALKEEEASESVSSTNFSGEMHFYELVEDTKDGIWLVQVIANDRSPLVGKIHWEKMVKKVSRFGIRT
    GTFNCSSDPRYCRRRGWVRSTLIMSVPQTSTSKGKVMLKEYSGRKIEVEHIFKWITAHAASRIKTIYNAEHLKEEWNKSD
    QYWLKIYLFANLDQPPAFFSALSIKFTGRVEFIFVNVENWDNKSYMTDIGIYNMPSYILRTPEGIYRYGNHTGEFISLQA
    MDSFLRSLQPEVNDLFVLSLVLVNLMAWMDLFITQGATIKRFVVLISTLGTYNSLLIISWLPVLGFLQLPYLDSFYEYSL
    KLLRYSNTTTLASWVRADWMFYSSHPALFLSTYLGHGLLIDYFEKKRRRNNNNDEVNANNLEWLSSLWDWYTSYLFHPIA
    SFQNFPVESDWDEDPDLFLERLAFPDLWLHPLIPTDYIKNLPMWRFKCLGVQSEEEMSEGSQDTENDSESENTDTLSSEK
    EVFEDKQSVLHNSPGTASHCDAEACSCANKYCQTSPCERKGRSYGSYNTNEDMEPDWLTWPADMLHCTECVVCLENFENG
    CLLMGLPCGHVFHQNCIVMWLAGGRHCCPVCRWPSYKKKQPYAQHQPLSNDVPS</protein>}\\
    \texttt{\textbf{Response}: Acts as an E2-dependent E3 ubiquitin-protein ligase, probably involved in the ER-associated protein degradation pathway.}
\end{framed}
This task has 4562 test samples.

\paragraph{Unconditional Molecular Generation}
\begin{framed}
    \small\textit{Example 1:}\\
    \texttt{\textbf{Prompt}: Generate a molecule with <SMILES>}\\
    \texttt{\textbf{Response}: <SMILES> CCCOC1=CC=C(C=C1)C(=O)OCCN(CC)CC </SMILES>}
\end{framed}
This task has 800 test samples.

\paragraph{Unconditional Protein Generation}
\begin{framed}
    \small\textit{Example 1:}\\
    \texttt{\textbf{Prompt}: Generate a protein sequence with <protein> </protein>.}\\
    \texttt{\textbf{Response}: <protein>MASSNVSSLNNTFLGKAARILKQEGEKGRIPATAVTAYAMAVFYTLVTDIPERDTDMLI\\SNGVLNVVPTIISKVARKSATIRLDVQARKEPTLPTKRASANYEPEE</protein>}
\end{framed}
This task has 1000 test samples.

\paragraph{Unconditional RNA Generation}
\begin{framed}
    \small\textit{Example 1:}\\
    \texttt{\textbf{Prompt}: Please generate a novel RNA sequence. <rna>}\\
    \texttt{\textbf{Response}: <rna>GGCTGGTCTGATGGTAGTGGGTTATCAGAACTTATTAACATTAGTGTCACTAAAGTTGGTATAC\\CCCCACCCCCACTGCTAAATTTGACTGGCTTTAA</rna>}
\end{framed}
This task has 5000 test samples.

\paragraph{Unconditional Material Generation}
\begin{framed}
    \small\textit{Example 1:}\\
    \texttt{\textbf{Prompt}: Produce a material that has any bulk modulus or composition}\\
    \texttt{\textbf{Response}: <material>Zr Zr Zr Zr Zr Zr Zr Zr O O O O O O O O O O O O O O O O <sg> <sg61></material>}
\end{framed}
This task has 5000 test samples.

\paragraph{Composition to Gaterial Generation}
\begin{framed}
    \small\textit{Example 1:}\\
    \texttt{\textbf{Prompt}: Given F, In, Tc, Zr, design a material composed of them.}\\
    \texttt{\textbf{Response}: ⟨material⟩Zr Zr Zr Zr Zr Zr In In In In In In Tc Tc F F F F F F F F F F F F F F F F F F F F F F F F F F F F F F F F F F F F F F F F F F ⟨sg⟩ ⟨sg176⟩⟨/material⟩}
\end{framed}
This task has 10000 test samples.

\paragraph{Bulk to Material Generation}
\begin{framed}
    \small\textit{Example 1:}\\
    \texttt{\textbf{Prompt}: Given a bulk modulus of 75.321 GPa, formulate the composition needed to achieve it.}\\
    \texttt{\textbf{Response}: ⟨material⟩Li Li Li Li Li Li Li Li Li Li Li Li Rh Rh Rh Rh N N N N N N N N ⟨sg⟩ ⟨sg141⟩⟨/material⟩}
\end{framed}
This task has 3000 test samples.

\paragraph{Description-Guided Molecule Design}
\begin{framed}
    \small\textit{Example 1:}\\
        \texttt{\textbf{Prompt}: Design a molecule that meets the criteria outlined in the description. The molecule is a natural product found in Piper aequale with data available.}\\
    \texttt{\textbf{Response}: <SMILES> C/C=C/C1=CC2=C(C=C1)OC(=C2C)C3=CC=C(C=C3)OC4=C(C=CC(=C4)C5\\
    =C(C6=C(O5)C=CC(=C6)/C=C/C)C)O </SMILES>}
\end{framed}
This task has 1000 test samples.

\paragraph{Protein Design}
\begin{framed}
    \small\textit{Example 1:}\\
            \texttt{\textbf{Prompt}: Synthesize a protein sequence with the appropriate folding and stability properties for the desired function. 1. The protein should be able to modulate glycine decarboxylation via glycine cleavage system in a way that leads to a desirable outcome.}\\
    \texttt{\textbf{Response}: The designed protein's amino acid sequence is as follows:
    <protein>MNIIKTGIKQSLVSGIYSKVGIRAFCTHYSAELEWVKLSDDNKVATVGLSSFGAQRLGKINYVELP\\
    KEHRKCRREEKFGVLESSNATAFGLYAPVSGEVLEVNEKLKKSPSLLNEDPANNWMVKFKVSKPDEFKKLMDSNK\\
    YKKFVQWYR</protein>}
\end{framed}
This task has 1176 test samples.

\paragraph{Forward Reaction Prediction}
\begin{framed}
    \small\textit{Example 1:}\\
            \texttt{\textbf{Prompt}: With the provided reactants and reagents, propose a potential product. <SMILES> CCI.O=C(O)C1=CC=CC=C1Br.CN(C)C=O.O.O=C([O-1])O.[Na+1] </SMILES>}\\
    \texttt{\textbf{Response}: <SMILES> CCOC(=O)C1=CC=CC=C1Br </SMILES>}
\end{framed}
This task has 1000 test samples.

\paragraph{Reagent Prediction}
\begin{framed}
    \small\textit{Example 1:}\\
                \texttt{\textbf{Prompt}: Based on the given chemical reaction, can you propose some likely reagents that might have been utilized? <SMILES> CCCBr.CCOCCN1C(CN2CCNCC2)=NC3=CC=CC=C31 </SMILES>}\\
    \texttt{\textbf{Response}: <SMILES> CCO.O=C([O-1])[O-1].[K+1].[K+1] </SMILES>}
\end{framed}
This task has 1000 test samples.

\paragraph{Retrosynthesis Prediction}
\begin{framed}
    \small\textit{Example 1:}\\
    \texttt{\textbf{Prompt}: Please suggest potential reactants for the given product. <SMILES> COC1=C(N)C=CC2=C1CCCC(N3CCOCC3)C2 </SMILES>}\\
    \texttt{\textbf{Response}: <SMILES> COC1=C([N+1](=O)[O-1])C=CC2=C1CCCC(N3CCOCC3)C2 </SMILES>}
\end{framed}
\begin{framed}
    \small\textit{Example 2:}\\
    \texttt{\textbf{Prompt}: With the provided product, recommend some probable reactants that were likely used in its production.<SMILES> [CH3:1][C:2]([CH3:3])([CH3:4])[O:5][C:6](=[O:7])[n:15]1[c:14]2[cH:13][cH:12][c:1\\1]([C:9]([CH3:8])=[O:10])[cH:19][c:18]2[cH:17][cH:16]1 </SMILES>}\\
    \texttt{\textbf{Response}: <SMILES> CC(C)(C)OC(=O)O[C:6]([O:5][C:2]([CH3:1])([CH3:3])[CH3:4])=[\\O:7].[CH3:8][c:9]1[cH:10][cH:11][c:12]([S:13](=[O:14])(=[O:15])[O:16][C@@H:17]2[\\CH2:18][NH:19][C@H:20]3[C@@H:21]2[O:22][CH2:23][C@@H:24]3[OH:25])[cH:26][cH:27]1 </SMILES>}
\end{framed}
\begin{framed}
    \small\textit{Example 3:}\\
    \texttt{\textbf{Prompt}: Identify possible reactants that could have been used to create the specified product. <SMILES> CC1=CC=C(N)N=C1N </SMILES>}\\
    \texttt{\textbf{Response}: <SMILES> CC(C\#N)CCC\#N.N </SMILES>}
\end{framed}
This task has 4156 test samples.

\paragraph{Forward Synthesis}
\begin{framed}
    \small\textit{Example 1:}\\
    \texttt{\textbf{Prompt}: <SMILES> NC1=CC=C2OCOC2=C1.O=CO </SMILES> Based on the reactants and reagents given above, suggest a possible product.}\\
    \texttt{\textbf{Response}: A possible product can be <SMILES> O=CNC1=CC=C2OCOC2=C1 </SMILES> .}
\end{framed}
This task has 4062 test samples.

\paragraph{Molecule Generation}
\begin{framed}
    \small\textit{Example 1:}\\
    \texttt{\textbf{Prompt}: Give me a molecule that satisfies the conditions outlined in the description: The molecule is a member of the class of tripyrroles that is a red-coloured pigment with antibiotic properties produced by Serratia marcescens. It has a role as an antimicrobial agent, a biological pigment, a bacterial metabolite, an apoptosis inducer and an antineoplastic agent. It is a tripyrrole, an aromatic ether and a ring assembly.}\\
    \texttt{\textbf{Response}: Here is a potential molecule: <SMILES> CCCCCC1=C(C)NC(/C=C2=C(C3=CC\\=CN3)C=C2OC)=C1 </SMILES>}
\end{framed}
This task has 2493 test samples.

\paragraph{Chemical Disease Interaction Extraction}
\begin{framed}
    \small\textit{Example 1:}\\
    \texttt{\textbf{Prompt}: You are provided with a set of clinical trial summaries. Extract the chemical-disease relations from the summaries and present your findings in the format of (Subject, Object). Eating disorders and the associated behavioural problems and drug abuse are uncommon in pregnancy. When they do occur they are often unrecognized because of denial but when significant may pose a risk to both the mother and her fetus. This case illustrates a number of problems that may be encountered in women with eating disorders in pregnancy, including prolonged and recurrent metabolic disturbances and diuretic abuse. In particular it illustrates the derangements of thyroid function seen in pregnant women with eating disorders and reminds us that when a cause for thyrotoxicosis remains obscure, thyroxine abuse should be considered and explored.}\\
    \texttt{\textbf{Response}: (thyroxine, thyrotoxicosis)}
\end{framed}
This task has 50 test samples.

\paragraph{Chemical Entity Recognition}
\begin{framed}
    \small\textit{Example 1:}\\
    \texttt{\textbf{Prompt}: Identify the chemical entities mentioned in the given text. Hence , to elucidate the cellular mechanisms that can address these divergent reports , we studied intestinal absorption of the dual substrate loperamide in portal vein - cannulated P - gp - competent and P - gp - deficient mice .}\\
    \texttt{\textbf{Response}: loperamide}
\end{framed}
This task has 100 test samples.

\paragraph{Chemical Protein Interaction Extraction}
\begin{framed}
    \small\textit{Example 1:}\\
        \texttt{\textbf{Prompt}: Your task is to extract the subject and object entities from the provided text and establish the corresponding relation triples. Please format your answers as (Subject, Relation, Object). ORF 17583, a histamine H2-receptor antagonist, inhibited gastric acid secretion in pylorus-ligated rats (ED50 = 4.9 mg/kg intraduodenal; 3.4 mg/kg p.o.; and 0.21 mg/kg i.p.) and in total gastric fistula or Heidenhain pouch dogs stimulated by betazole (ED50 = 0.12 mg/kg p.o. and 0.08 mg/kg i.v.), histamine, tetragastrin, bethanechol, 2-deoxy-D-glucose or a meal (ED50 values ranged from 0.11-0.26 mg/kg p.o.). The nonspecific inhibition of gastric acid by ORF 17583 supports the existence of interdependence between histamine and the gastrin and cholinergic receptors on the parietal cell surface. Antisecretory potency of ORF 17583 after intraduodenal administration in pylorus-ligated rats was 6.4 times greater than cimetidine, 1.8 times greater than ranitidine, equal to that of omeprazole and 8 times less than that of famotidine. Oral antisecretory potency of ORF 17583 in gastric fistula dogs was 31 times greater than cimetidine, 3.7 times greater than ranitidine and equal to that of omeprazole and famotidine. Studies using equieffective antisecretory doses of ORF 17583 and ranitidine in dogs suggested that ORF 17583 has a short duration of antisecretory activity similar to that of ranitidine.}\\
    \texttt{\textbf{Response}: (ORF 17583, antagonist, histamine H2-receptor), (ORF 17583, antagonist, histamine H2-receptor)}
\end{framed}
This task has 70 test samples.

\paragraph{General Multiple Choice Question}
\begin{framed}
    \small\textit{Example 1:}\\
            \texttt{\textbf{Prompt}: The following is a marker of acute hepatitis B infection What of the following is the right choice?\\
            (A) DNA polymerase (B) Hepatitis core antigen (C) Anti HBs (D) IgG to core antigen}\\
    \texttt{\textbf{Response}: The final answer is (A).}
\end{framed}
This task has 1001 test samples.

\paragraph{Open-Ended Question}
\begin{framed}
    \small\textit{Example 1:}\\
            \texttt{\textbf{Prompt}: 'C' wave in JVP is due to }\\
    \texttt{\textbf{Response}: Tricuspid valve bulging into right atrium}
\end{framed}
This task has 1004 test samples.

\paragraph{True/False Question}
\begin{framed}
    \small\textit{Example 1:}\\
            \texttt{\textbf{Prompt}: Is routine intraoperative frozen-section examination of sentinel lymph nodes in breast cancer worthwhile? }\\
    \texttt{\textbf{Response}: Maybe, in breast cancer patients having SLN biopsy, the failure of routine intraoperative FS is largely the failure to detect micrometastatic disease. The benefit of routine intraoperative FS increases with tumor size. Routine FS may not be indicated in patients with the smallest invasive cancers.}
\end{framed}
This task has 100 test samples.

\clearpage

\subsection*{TMScore for Conditional Protein Generation}
% In Figwe calculate TMScore for the last 1000 samples

\begin{figure}[h]
    \centering
    \includegraphics[width=1.0\linewidth]{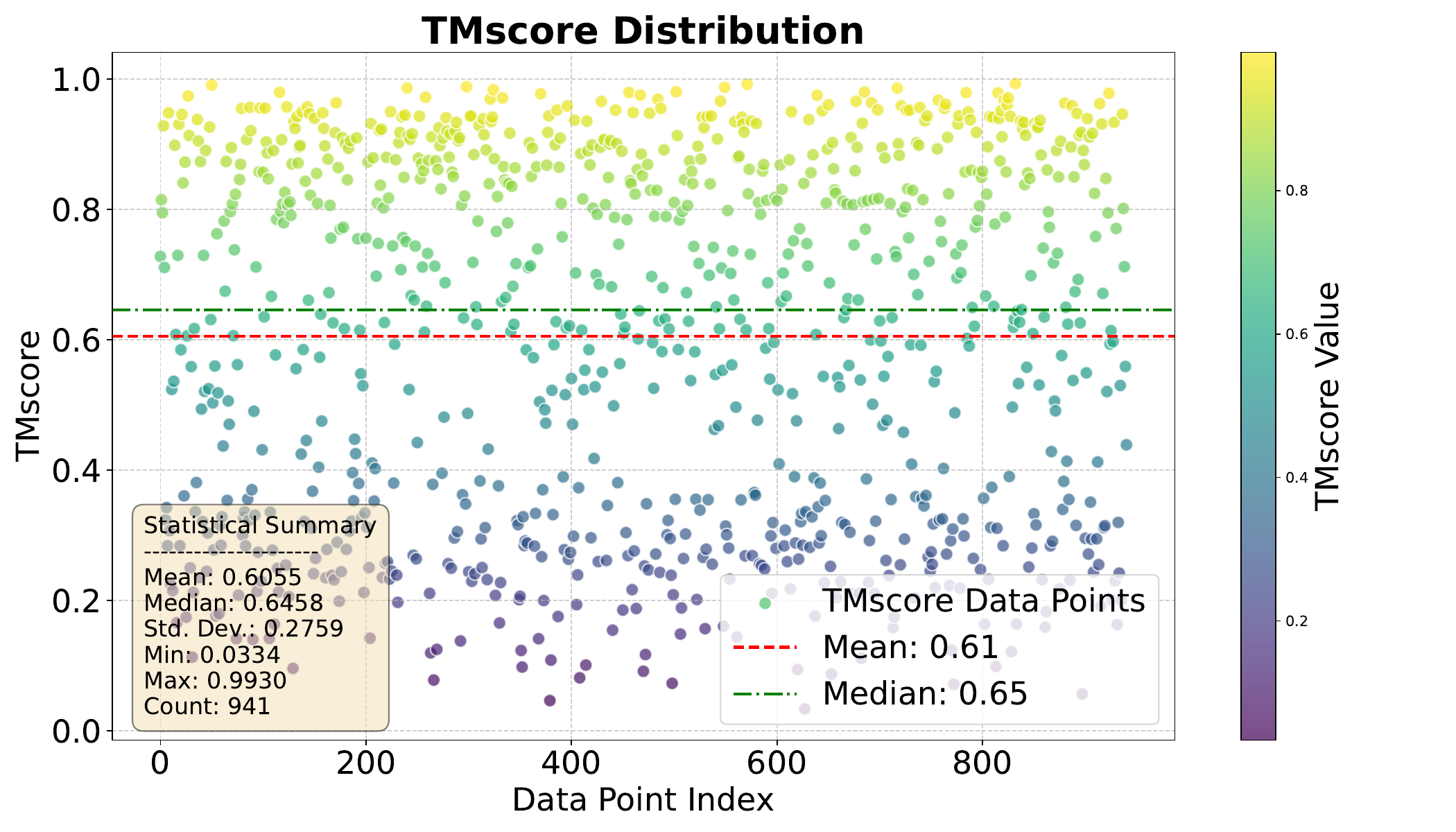}
    \caption{TMScore for last 1000 samples}
    \label{fig:tmscore_distribution}
\end{figure}

\clearpage

\end{document}